%% file: main.tex
\documentclass[10pt,twocolumn,letterpaper]{article}

\usepackage{cvpr}
\usepackage{times}
\usepackage{epsfig}
\usepackage{graphicx}
\usepackage{amsmath}
\usepackage{amssymb}
\usepackage{multirow}
\usepackage{graphbox}
\usepackage{caption}
\usepackage{array}
\newcolumntype{M}[1]{>{\centering\arraybackslash}m{#1}}

\usepackage{placeins}

\newenvironment{figurefont}{\fontfamily{qag}\selectfont}{\par}

\include{math_commands}

\usepackage[pagebackref=true,breaklinks=true,letterpaper=true,colorlinks,bookmarks=false]{hyperref}

\cvprfinalcopy 

\def\cvprPaperID{4729} 

\ifcvprfinal\fi
\begin{document}

	\title{Editing in Style: Uncovering the Local Semantics of GANs}
	
	\author{Edo Collins$^{1}$~~~~Raja Bala$^{2}$~~~~Bob Price$^{2}$~~~~Sabine Süsstrunk$^{1}$\\[1mm]
		\normalsize $^{1}$School of Computer and Communication Sciences, EPFL, Switzerland\\ \normalsize$^{2}$Interactive and Analytics Lab, Palo Alto Research Center, Palo Alto, CA\\
		{\tt\normalsize \{edo.collins,sabine.susstrunk\}@epfl.ch~~\{rbala,bprice\}@parc.com}
	}
	
	\twocolumn[{%
		\renewcommand\twocolumn[1][]{#1}%
		\maketitle
		\begin{center}
			\centering
			\includegraphics[width=\textwidth]{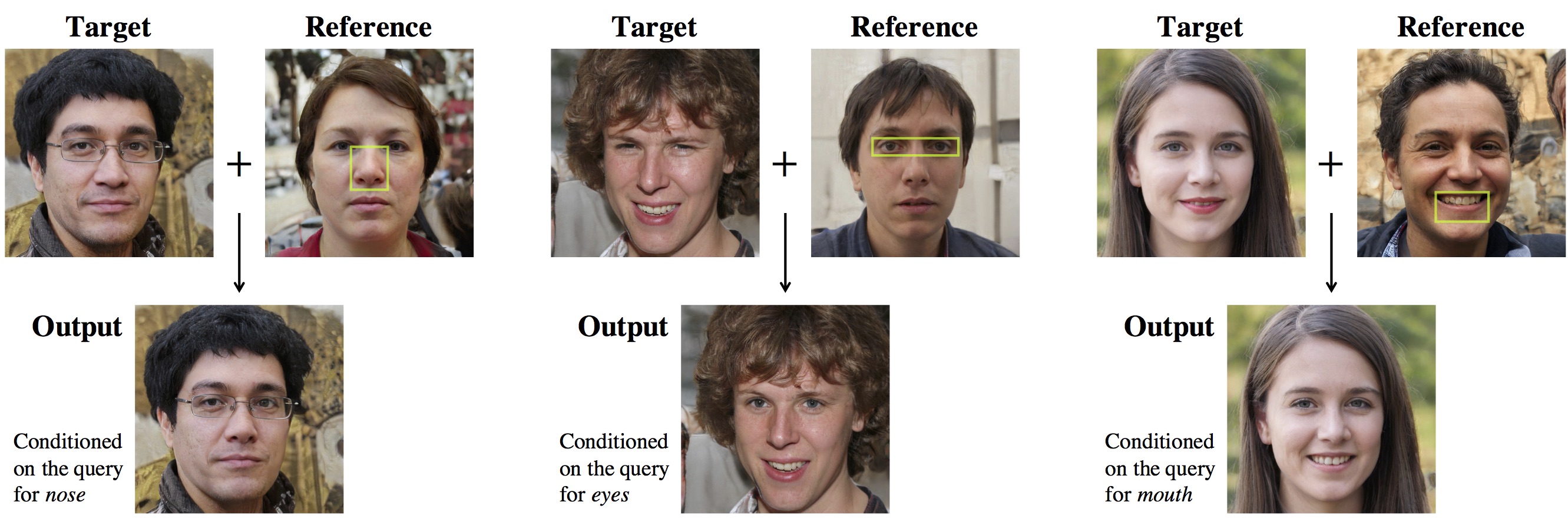}%
			\captionsetup{type=figure}
			\caption{Our method performs local semantic editing on GAN output images, transferring the appearance of a specific object part from a reference image to a target image.} \label{fig:splash-faces}

		\end{center}
	}]
	
	\begin{abstract}
		While the quality of GAN image synthesis has improved tremendously in recent years, our ability to control and condition the output is still limited. Focusing on StyleGAN, we introduce a simple and effective method for making local, semantically-aware edits to a target output image. This is accomplished by borrowing elements from a source image, also a GAN output, via a novel manipulation of style vectors. Our method requires neither supervision from an external model, nor involves complex spatial morphing operations. Instead, it relies on the emergent disentanglement of semantic objects that is learned by StyleGAN during its training. Semantic editing is demonstrated on GANs producing human faces, indoor scenes, cats, and cars. We measure the locality and photorealism of the edits produced by our method, and find that it accomplishes both.
	\end{abstract}
	
	\section{Introduction} \label{sec:Introduction}
	In the short span of five years, generative adversarial neural networks (GANs) have come to dominate the field of data-driven image synthesis. Like most other neural network models, however, the exact model they learn for the data is not straightforwardly interpretable.
	
	There have been significant steps towards alleviating this issue. For instance, state-of-the-art image GANs such as PG-GAN \cite{karras2018progressive} and StyleGAN \cite{karras2019style}, by virtue of their progressive training, encourage each layer to model the variation exhibited at given image resolutions (e.g., 8$\times$8 images capture coarse structure, $32\times32$ add finer details, etc.).
	
	The notion of a \emph{disentangled} representation has been used to describe such phenomena. While definitions of disentanglement are many and varied \cite{higgins2018towards}, the common idea is that an attribute of interest, which we often consider \emph{semantic}, can be manipulated independently of other attributes.
	
	In this paper we show that deep generative models like PG-GAN, StyleGAN and the recent StyleGAN2 \cite{karras2019analyzing} learn a representation of objects and object-parts that is disentangled in the sense that various semantic parts (e.g., the \emph{mouth} of a person or the \emph{pillows} in a bedroom) have a significant ability to vary independently of the rest of the scene.
	
	Based on this observation we propose an algorithm that performs \emph{spatially-localized semantic editing} on the outputs of GANs - primarily StyleGAN. Editing is performed by transferring semantically localized \emph{style} from a reference image to a target image, both outputs of a GAN. Our method is simple and effective, requiring no more than an off-the-shelf pre-trained GAN. Our method is unique in that it enacts a localized change through a global operation, akin to style transfer. As a result, unlike other GAN editing methods that make use of additional datasets and trained networks, or traditional image morphing methods requiring complex spatial operations, our method relies upon and benefits solely from the rich semantic representation learned by the GAN itself. Applications include forensic art where a human face is composited from various sources; and interior design where various combinations of design elements such as furniture, upholstery, etc., can be visualized. Extension to semantic editing of real images can be envisioned by combining our approach with the recent work that embeds natural images into the latent space of StyleGAN \cite{abdal2019,karras2019analyzing}.

	We make the following contributions:
	\begin{itemize}
		\item We provide insight into the structure of hidden activations of the StyleGAN generator, showing that the learned representations are largely disentangled with respect to semantic objects in the synthesized image.
		\item We exploit this structure to develop a novel image editor that performs semantic part transfer from a reference to a target synthesized image. The underlying formulation is simple and elegant and achieves naturalistic part transfer without the need for complex spatial processing, or supervision from additional training data and models.
	\end{itemize}
	
	The paper is structured as follows. In Section \ref{sec:Related Work} we review work related to GAN editing and interpretability. In Section \ref{sec:Local Semantics} we detail our observations regarding spatial disentanglement in GAN latent space and introduce our local editing method. In Section \ref{sec:Experiments} we show experimental results that validate our claims, and in Section \ref{sec:Conclusion} we conclude with a discussion of the results and future work.

	\section{Related Work} \label{sec:Related Work}
	The literature on the use of GANs for image synthesis has exploded since the seminal work by Goodfellow et al. \cite{goodfellow2014}, with today's state of art methods such as StyleGAN \cite{karras2019style}, StyleGAN2\cite{karras2019analyzing}, and BigGAN \cite{brock2019} producing extremely realistic outputs. For a thorough review of the GAN literature we refer the reader to recent surveys in \cite{creswell2018,Huang2018,WangFaceGan2019}. Our goal here is \emph{not} to propose another GAN, but to offer a local editing method for its output, by changing the style of specific objects or object parts to the style given in a reference image. We next review past work germane to semantic image editing, paying particular attention to recent GAN-based methods. 
	
	\subsection{GAN-based Image Editing} \label{sec:GAN-based Image Editing}
	Several works have explored the use of deep generative models for semantic image editing. We distinguish between two flavors: latent code-based methods for global attribute editing and activation-based methods for local editing.
	
	Latent code-based techniques learn a manifold for natural images in the latent code space facilitated by a GAN and perform semantic edits by traversing paths along this manifold \cite{Perarnau2016,zhu2016}. A variant of this framework employs auto-encoders to disentangle the image into semantic subspaces and reconstruct the image, thus facilitating semantic edits along the individual subspaces \cite{biofacenet,shu2017cvpr}. Examples of edits accomplished by these techniques include global changes in color, lighting, pose, facial expression, gender, age, hair appearance, eyewear and headwear  \cite{biofacenet,lample2017,shu2017cvpr,shen2019interpreting,zhang2018}. AttGAN \cite{he2019attgan} uses supervised learning with external attribute classifiers to accomplish attribute editing.
	
	Activation-based techniques for local editing directly manipulate specific spatial positions on the activation tensor at certain convolutional layers of the generator. In this way, GAN Dissection \cite{bau2018} controls the presence or absence of objects at given positions, guided by supervision from an independent semantic segmentation model.
	Similarly, feature blending \cite{suzuki2018spatially} transfers objects between a target GAN output and a reference by ``copy-pasting'' activation values from the reference onto the target. We compare that technique, together with traditional Poisson blending \cite{perez2003poisson}, to our approach in Fig. \ref{fig:copy-paste-comapre}.
	
	Distinct from all these works, our approach is a latent code-based approach for local editing. Crucially, it neither relies on external supervision by image segmentation models nor involves complex spatial blending operations. Instead, we uncover and exploit the disentangled structure in the embedding space of the generator that naturally permits spatially localized part editing. 
	
	\subsection{Face Swapping}
	Our technique for object-specific editing, when applied to face images, is akin to the problems of face swapping and transfer. Previous efforts \cite{korshunova2017,mosaddegh2014,natsume2018} describe methods for exchanging global properties between a pair of facial images. Our method stands out from these approaches by offering editing that is localized to semantic object parts. Furthermore, a primary motivation for face swapping is de-identification for privacy preservation, which is not relevant for our goal of editing synthetic images. Yang et al. \cite{yang_acm2011} present a method for transferring expression from one face to another. Certain specific cases of expression transfer (e.g., smile) involve localized part (e.g., mouth) transfer, and are thus similar to our setting. However, even in these common scenarios, our editing framework is unique in that it requires no explicit spatial processing such as warping and compositing.

	\section{Local Semantics in Generative Models} \label{sec:Local Semantics}
	
	\subsection{Feature factorization} \label{sec:Feature factorization}
	Deep feature factorization (DFF) \cite{collins2018} is a recent method that explains a convolutional neural network's (CNN) learned representation through a set of saliency maps, extracted by factorizing a matrix of hidden layer activations. With such a factorization, it has been shown that CNNs trained for ImageNet classification learn features that act as semantic object and object-part detectors.
	
	Inspired by this finding, we conducted a similar analysis of the activations of generative models such as PG-GAN, StyleGAN, and StyleGAN2. Specifically, we applied spherical {\tt k}-means clustering \cite{buchta2012spherical} to the $C$-dimensional activation vectors that make up the activation tensor $\tA\in\sR^{N\times C\times H\times W}$ at a given layer of the generator, where $N$ is the number of images, $C$ is the number of channels, and $H,W$ are spatial dimensions. The clustering generates a tensor of cluster memberships, $\tU\in\{0,1\}^{N\times K\times H\times W}$, where $K$ is user-defined and each $K$-dimensional vector is a one-hot vector which indicates to which of $K$ clusters a certain spatial location in the activation tensor belongs.
	
\begin{figure}
		\centering
		\includegraphics[width=0.47\textwidth]{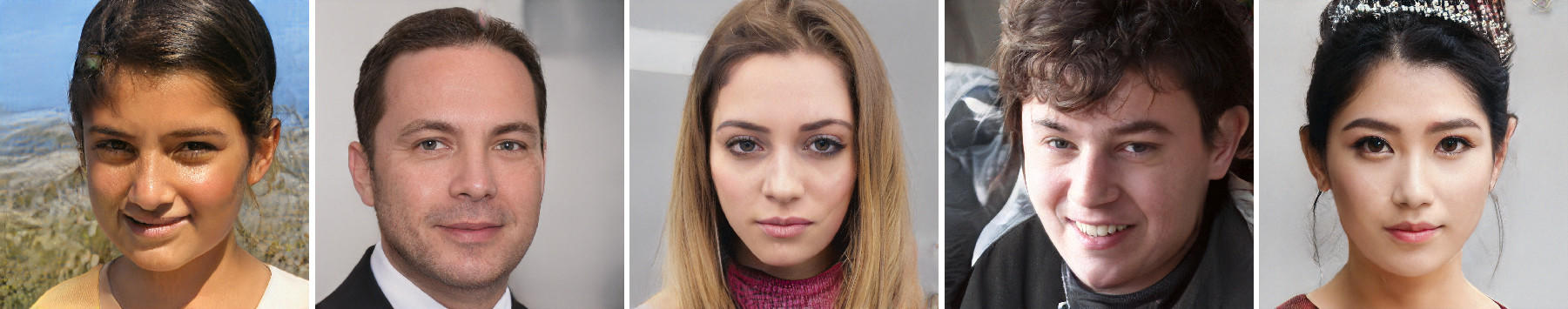}\\
		\includegraphics[width=0.47\textwidth]{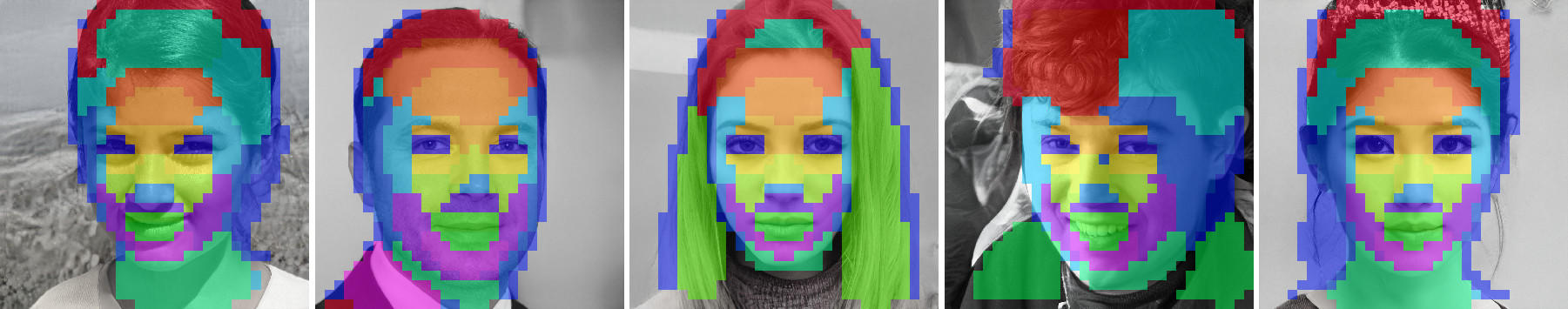}\\[1ex]
		\includegraphics[width=0.47\textwidth]{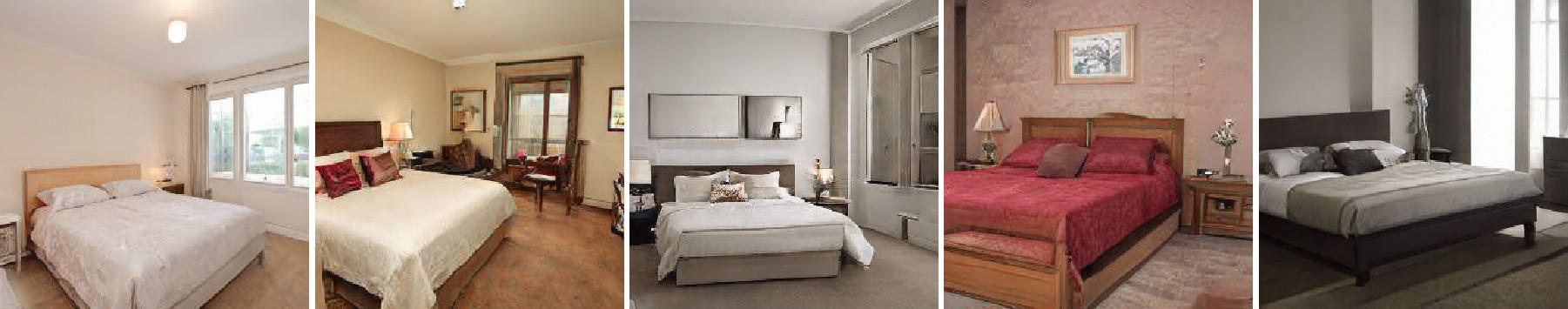}\\
		\includegraphics[width=0.47\textwidth]{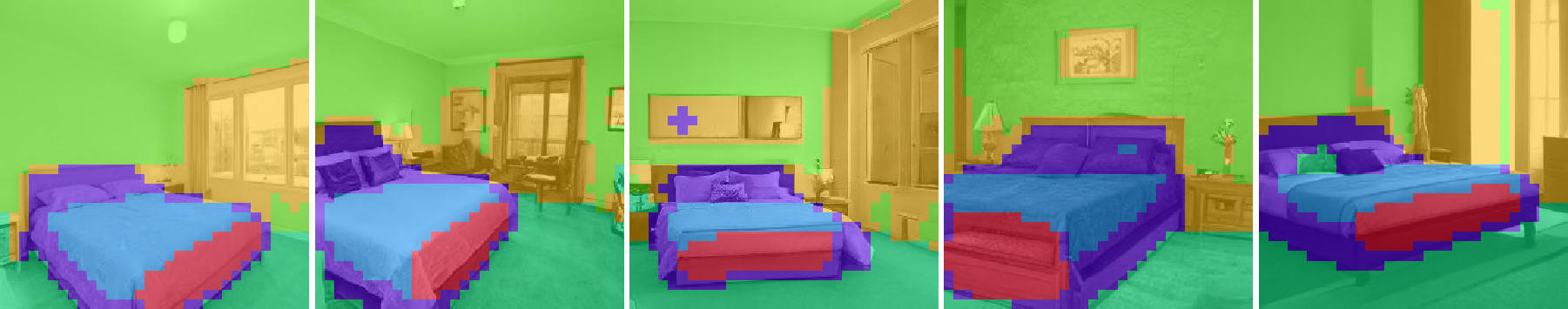}
		\caption{Applying {\tt k}-means to the hidden layer activations of the StyleGAN generator reveals a decomposition of the generated output into semantic objects and object-parts.}
		\label{fig:clustering}
	\end{figure}

	The main result of this analysis is that at certain layers of the generator, clusters correspond well to semantic objects and parts.
	Fig. \ref{fig:clustering} shows the clusters produced for a $32\times 32$ layer of StyleGAN generator networks trained on Flickr-Faces-HQ (FFHQ) \cite{karras2019style} and LSUN-Bedrooms \cite{yu15lsun}.
	Each pixel in the heatmap is color-coded to indicate its cluster. As can be seen, clusters spatially span coherent semantic objects and object-parts, such as \emph{eyes}, \emph{nose} and \emph{mouth} for faces, and \emph{bed}, \emph{pillows} and \emph{windows} for bedrooms. 
	
	The cluster membership encoded in $\tU$ allows us to compute the contribution $\mM_{k,c}$ of  channel $c$ towards each semantic cluster $k$ as follows:
	\begin{align}
	\mM_{k,c} = \frac{1}{N\dot H\dot W} \sum_{n,h,w} \tA_{n,c,h,w}^2 \odot \tU_{n,k,h,w}.
	\label{eq:channel2cluster}
	\end{align}.
	
	Assuming that the feature maps of $\tA_l$ have zero mean and unit variance, the contribution of each channel is bound between 0 and 1, i.e., $\mM\in[0,1]^{K\times C}$.
	
	Furthermore, by bilinearly up- or down-sampling the spatial dimensions of the tensor $\tU$ to an appropriate size, we are able to find a matrix $\mM$ for all layers in the generator, with respect to the same semantic clusters.
	
	Using this approach we produced a semantic catalog for each GAN. We chose at which layer and with which $K$ to apply spherical {\tt k}-means guided by a qualitative evaluation of the cluster membership maps. This process requires only minutes of human supervision.
	

	
	
	\begin{figure*}
		{
			\centering \setlength{\tabcolsep}{0pt} 
			\figurefont
			\renewcommand{\arraystretch}{0.1}
			\newcommand\figpath{figures/PartGrid/FFHQ_2001/}
			\newcommand\partone{eyes}
			\newcommand\parttwo{nose}
			\newcommand\partthree{mouth}
			\begin{tabular}{M{0.2\textwidth}M{0.18\textwidth}M{0.18\textwidth}M{0.18\textwidth}M{0.18\textwidth}}
				\textbf{Target} & \textbf{Reference 1} & \textbf{Reference 2} & \textbf{Reference 3}  & \textbf{Reference 4} \\
				\includegraphics[width=0.18\textwidth]{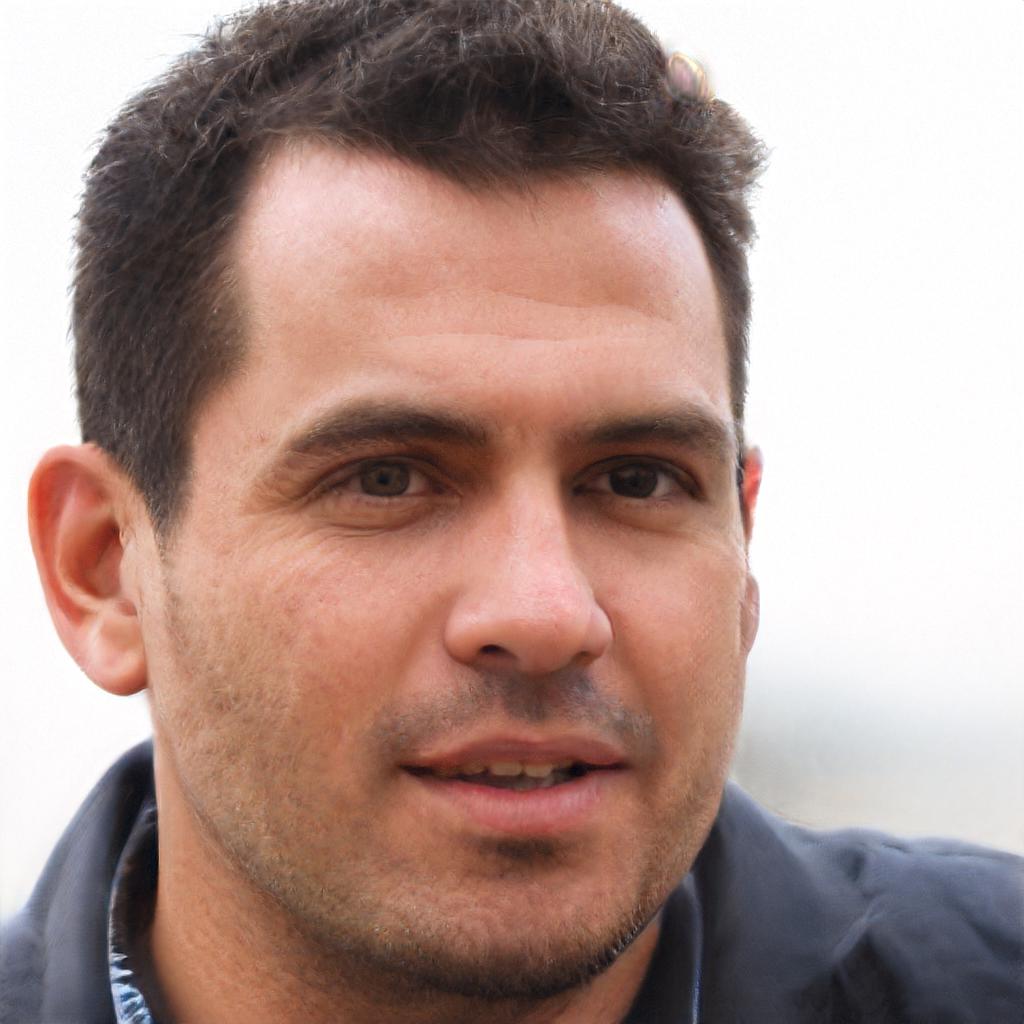}  &
				\includegraphics[width=0.18\textwidth]{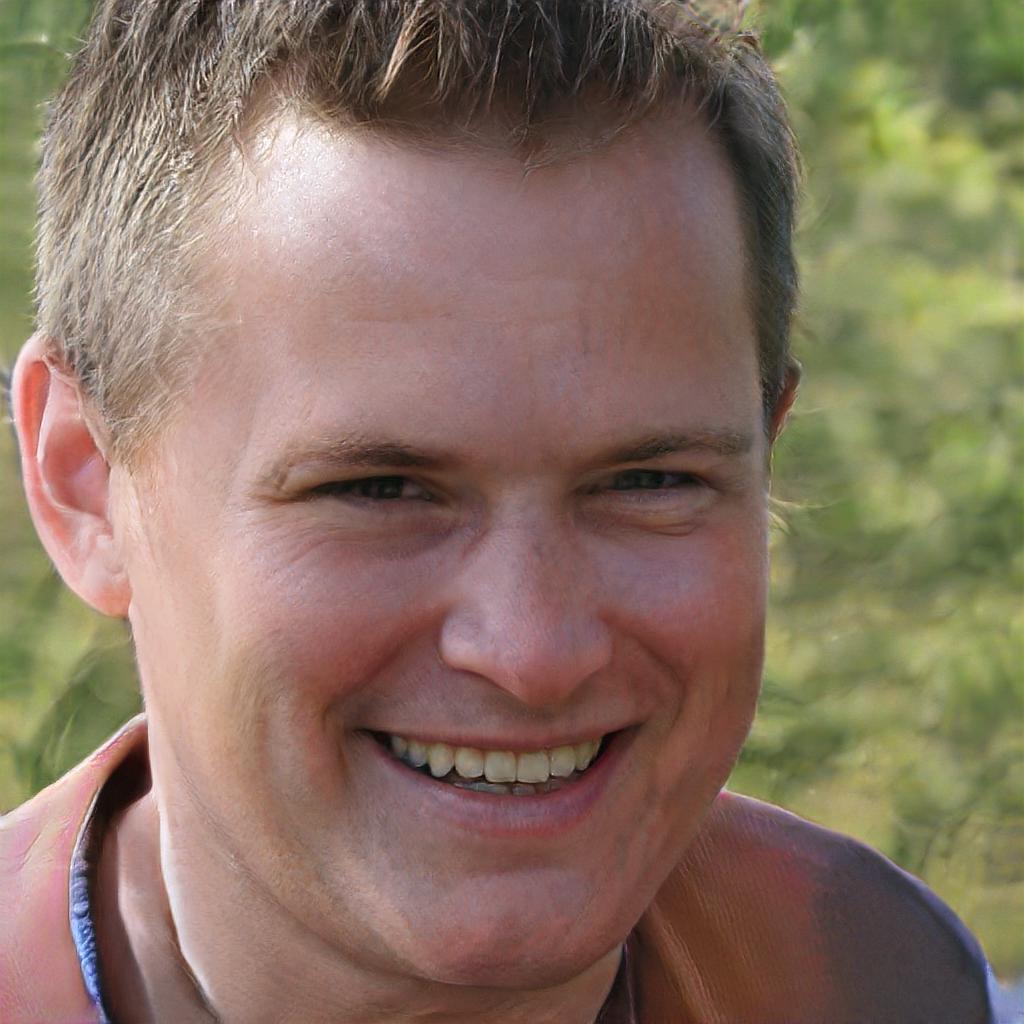} &
				\includegraphics[width=0.18\textwidth]{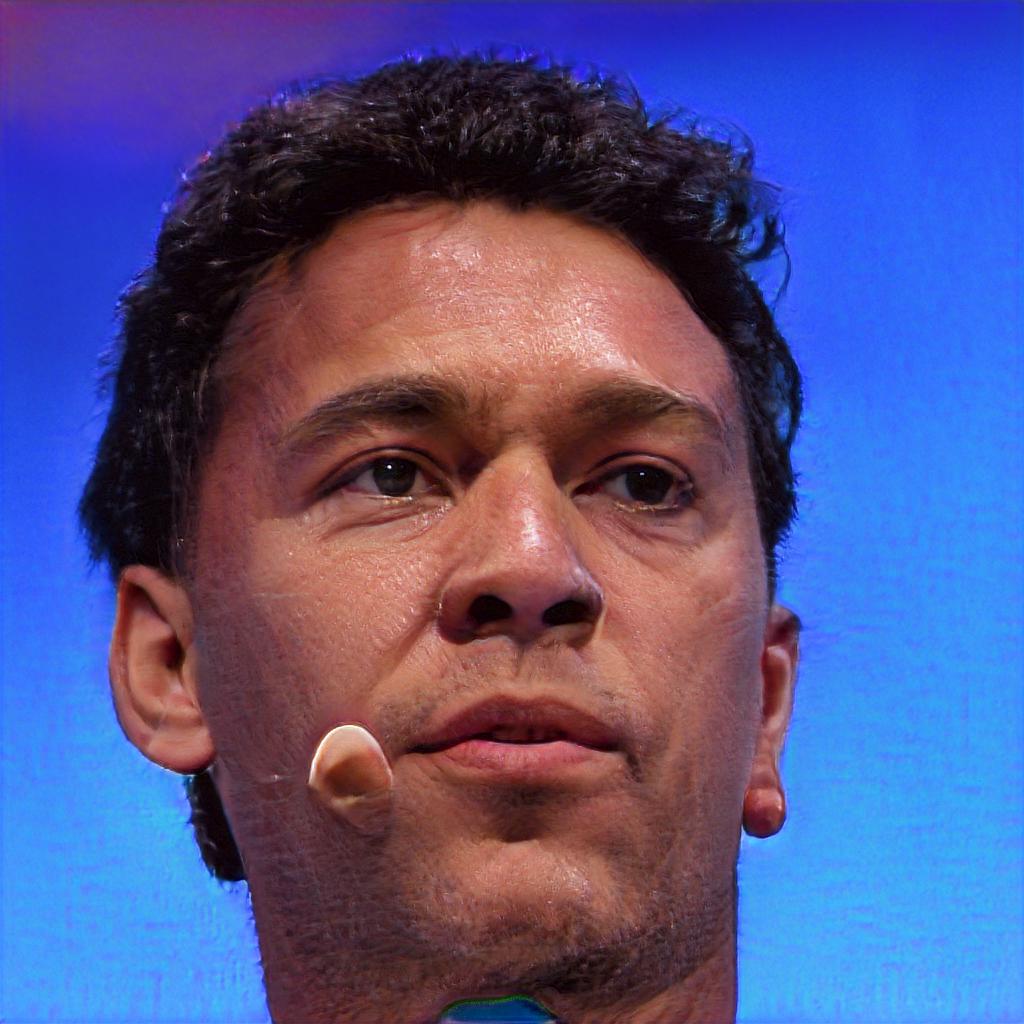} &
				\includegraphics[width=0.18\textwidth]{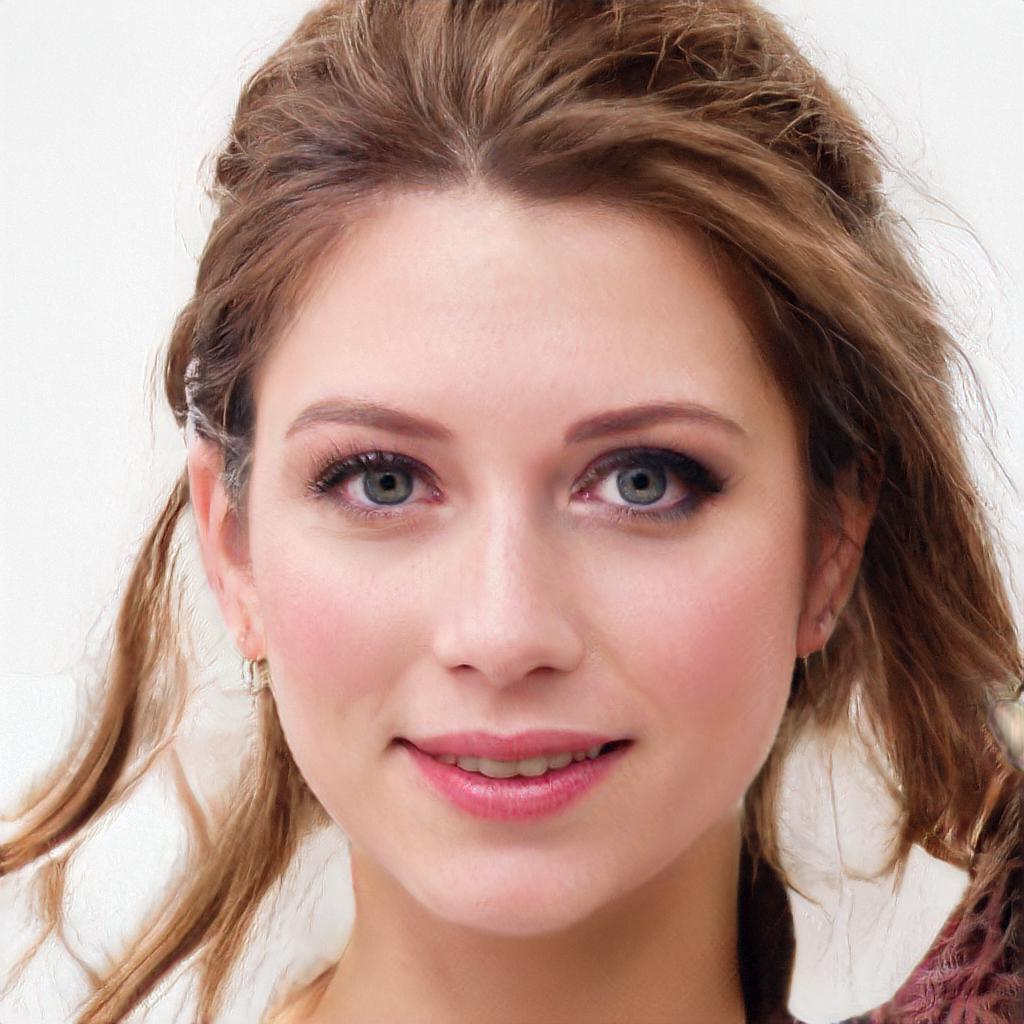} &
				\includegraphics[width=0.18\textwidth]{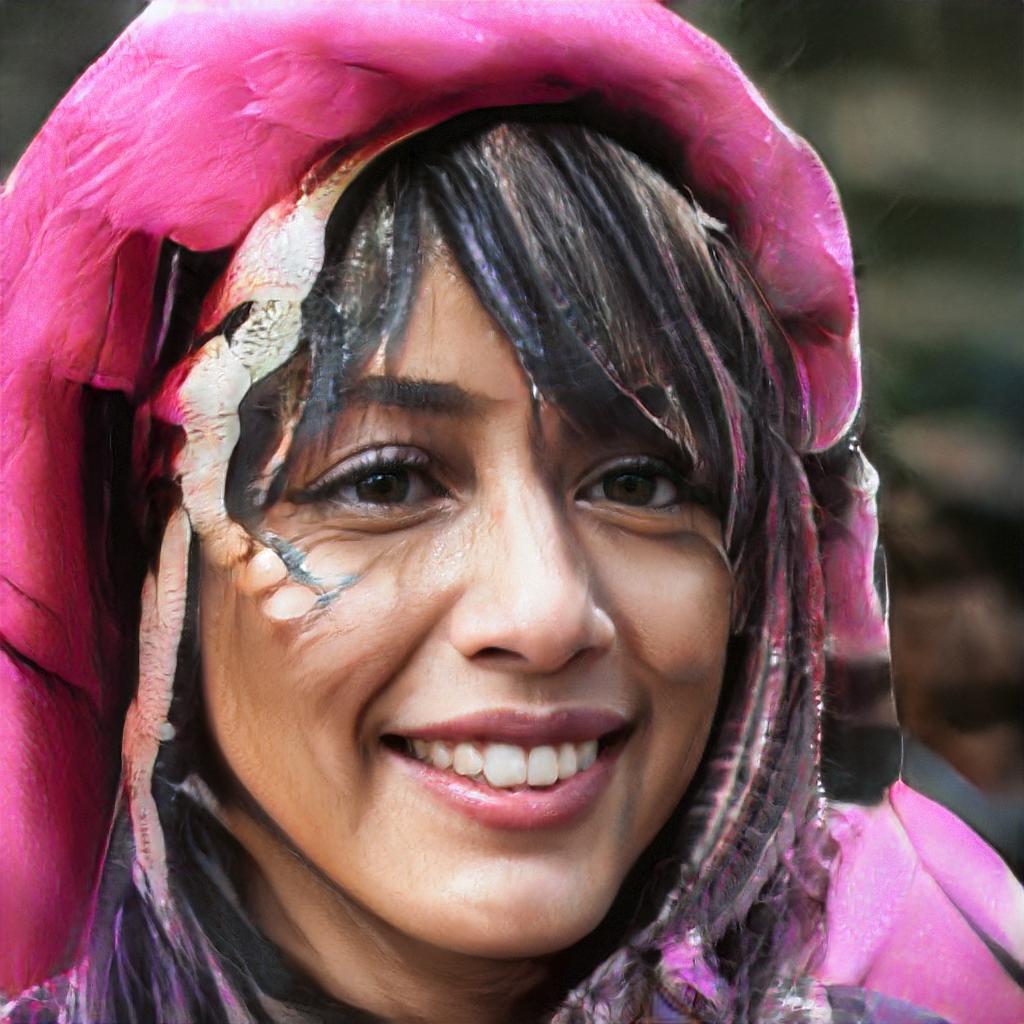}\\
				&&&& \\
				\textbf{\expandafter\MakeUppercase \partone:} &\includegraphics[width=0.18\textwidth]{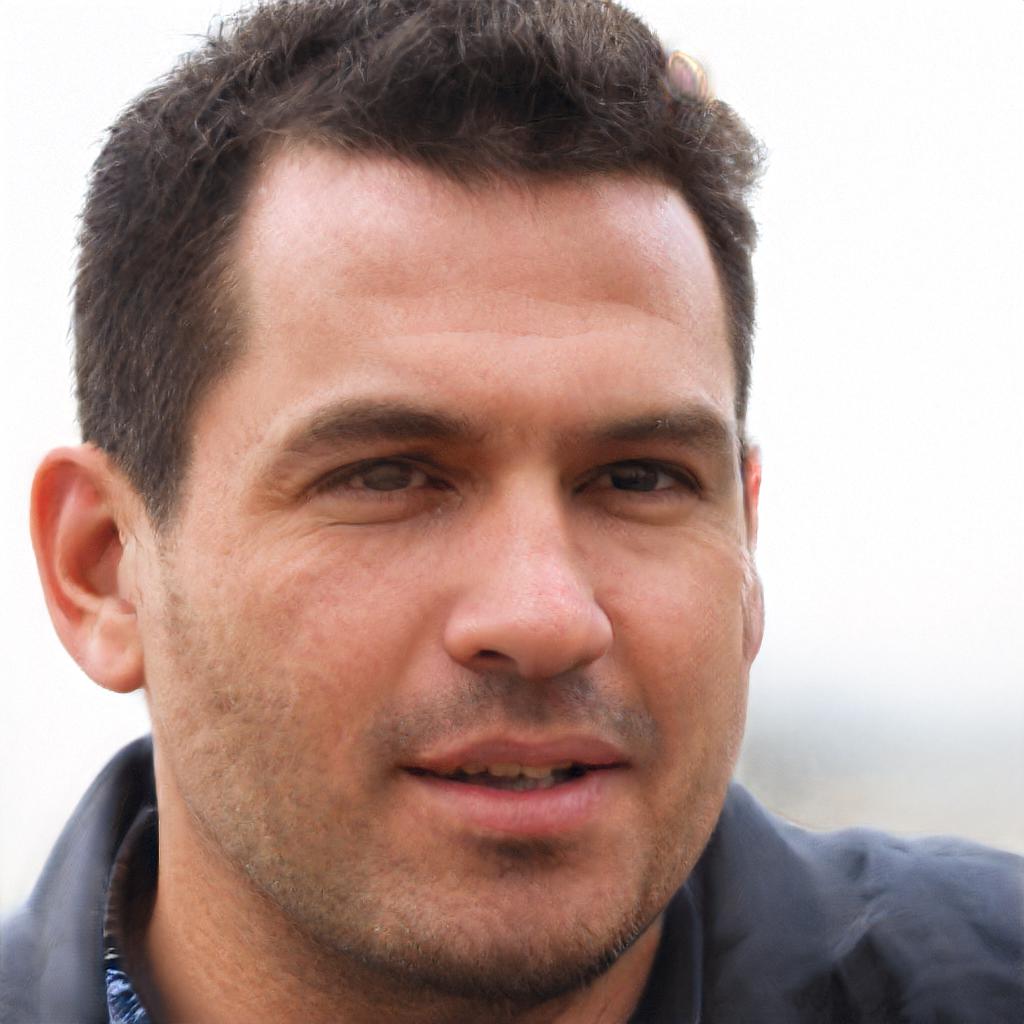} &
				\includegraphics[width=0.18\textwidth]{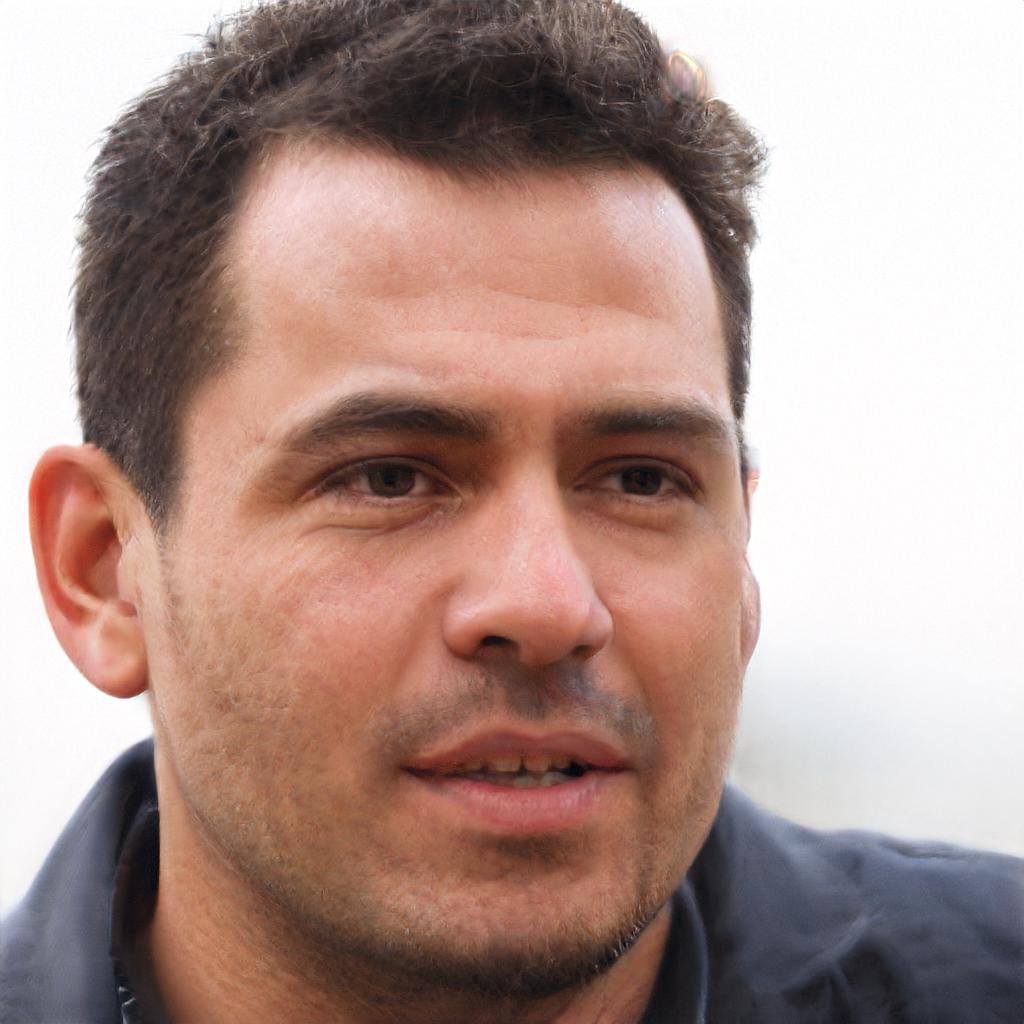} &
				\includegraphics[width=0.18\textwidth]{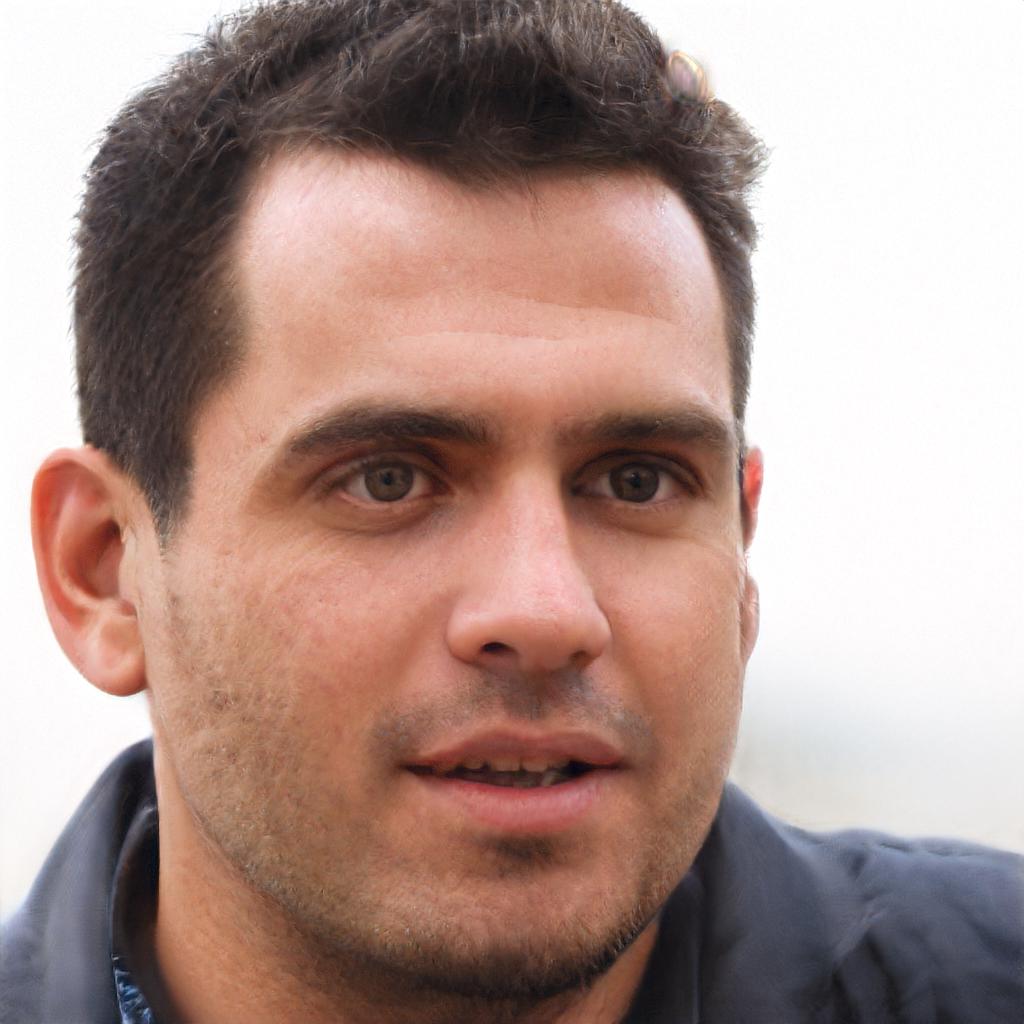} &
				\includegraphics[width=0.18\textwidth]{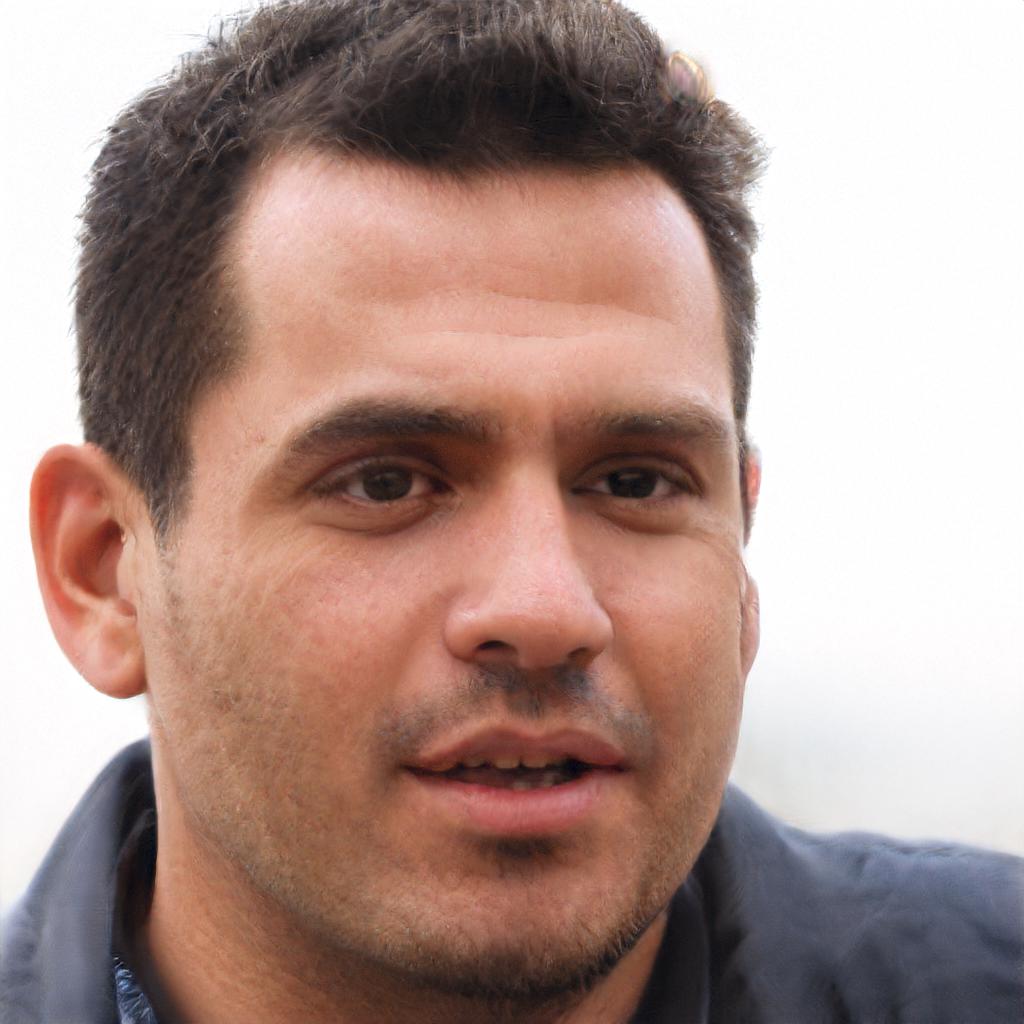}\\
				
				\textbf{\expandafter\MakeUppercase \parttwo:} &\includegraphics[width=0.18\textwidth]{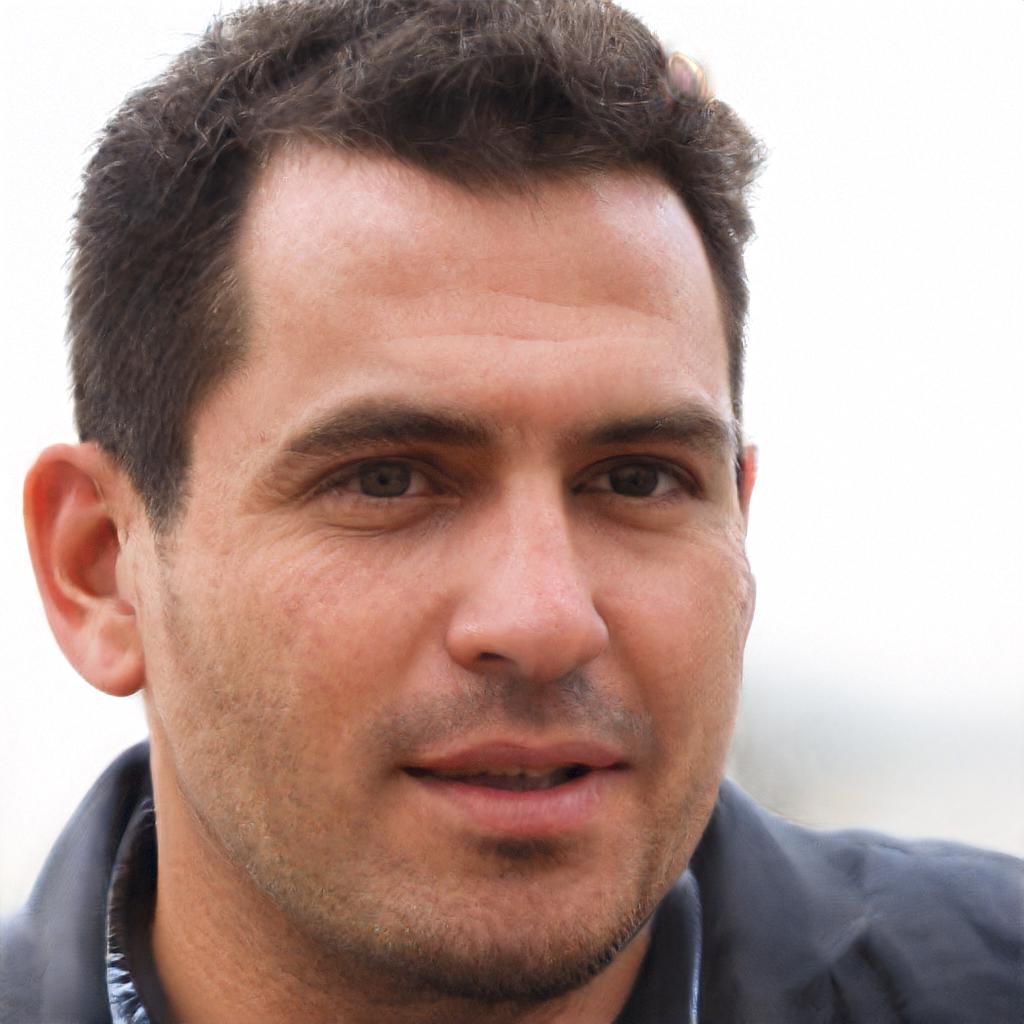} &
				\includegraphics[width=0.18\textwidth]{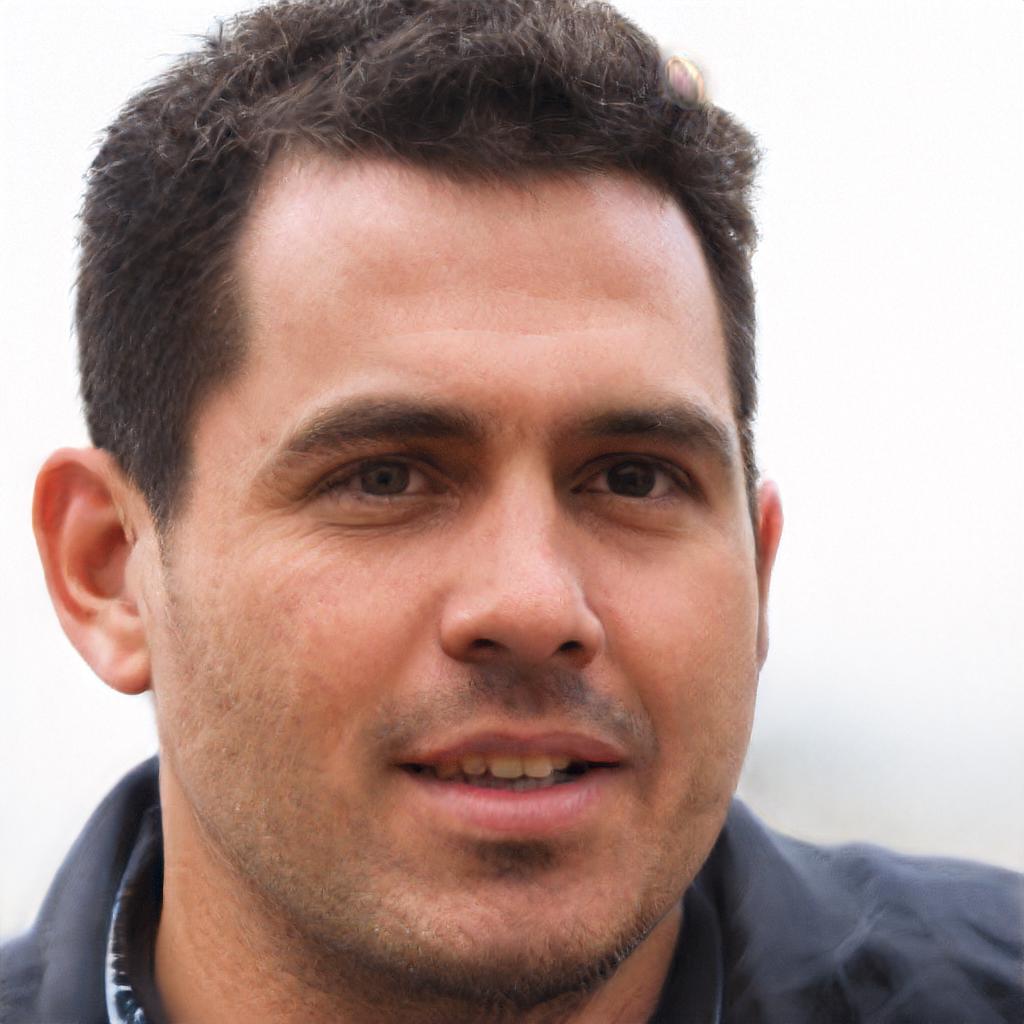} &
				\includegraphics[width=0.18\textwidth]{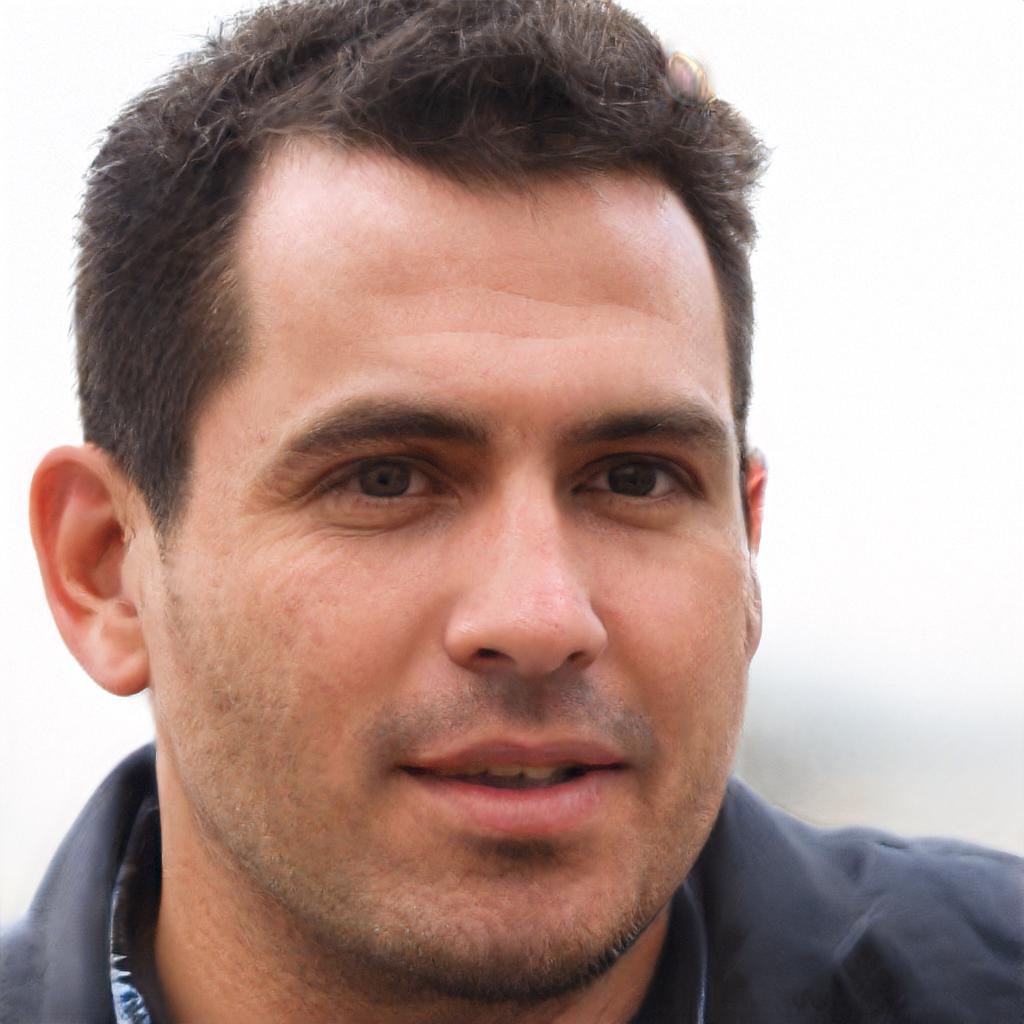} &
				\includegraphics[width=0.18\textwidth]{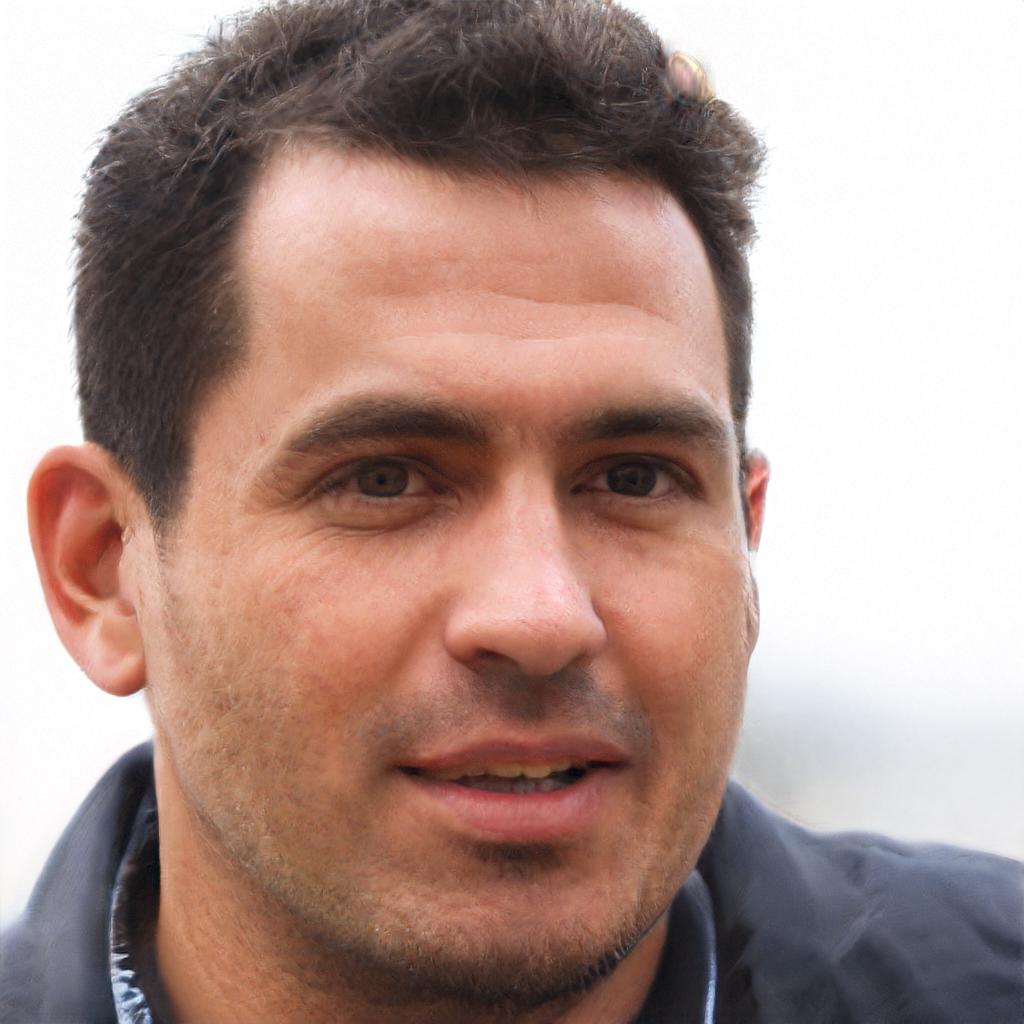}\\
				
				\textbf{\expandafter\MakeUppercase \partthree:} &\includegraphics[width=0.18\textwidth]{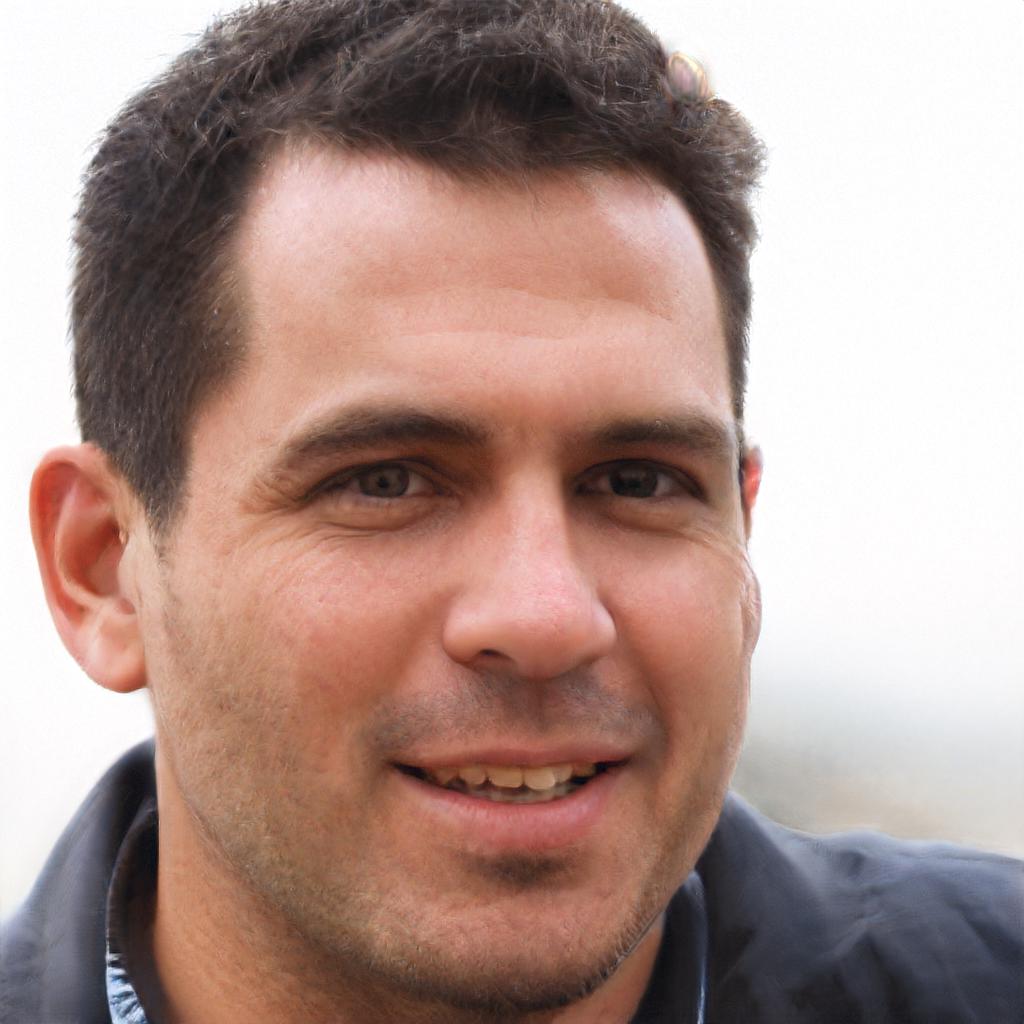} &
				\includegraphics[width=0.18\textwidth]{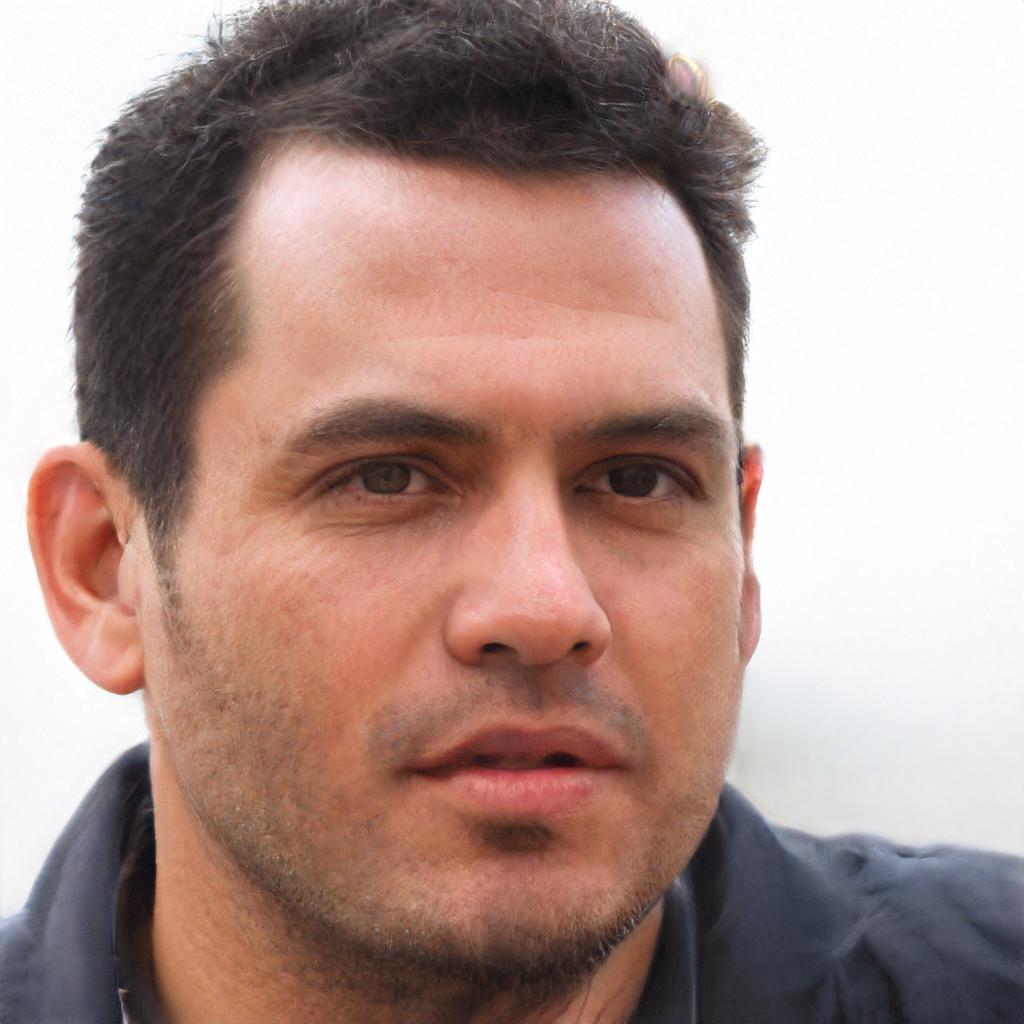} &
				\includegraphics[width=0.18\textwidth]{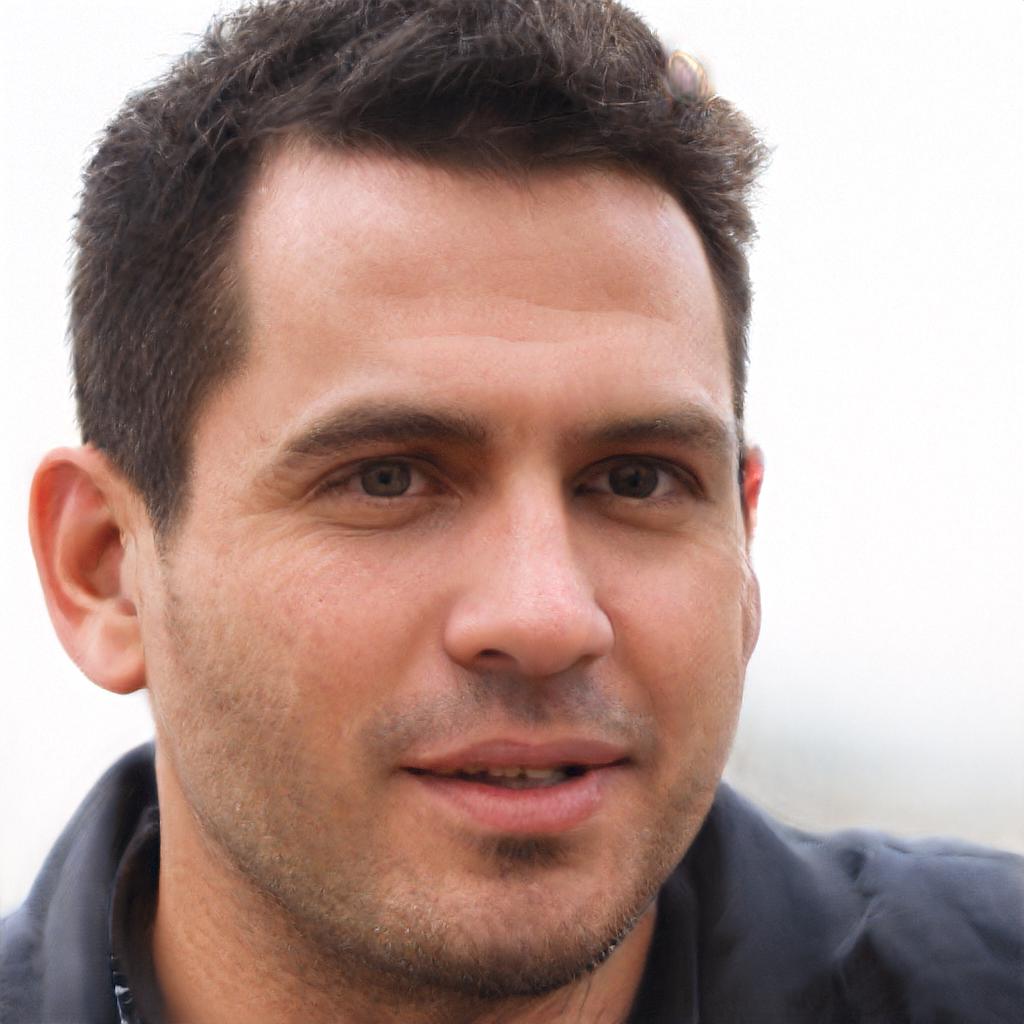} &
				\includegraphics[width=0.18\textwidth]{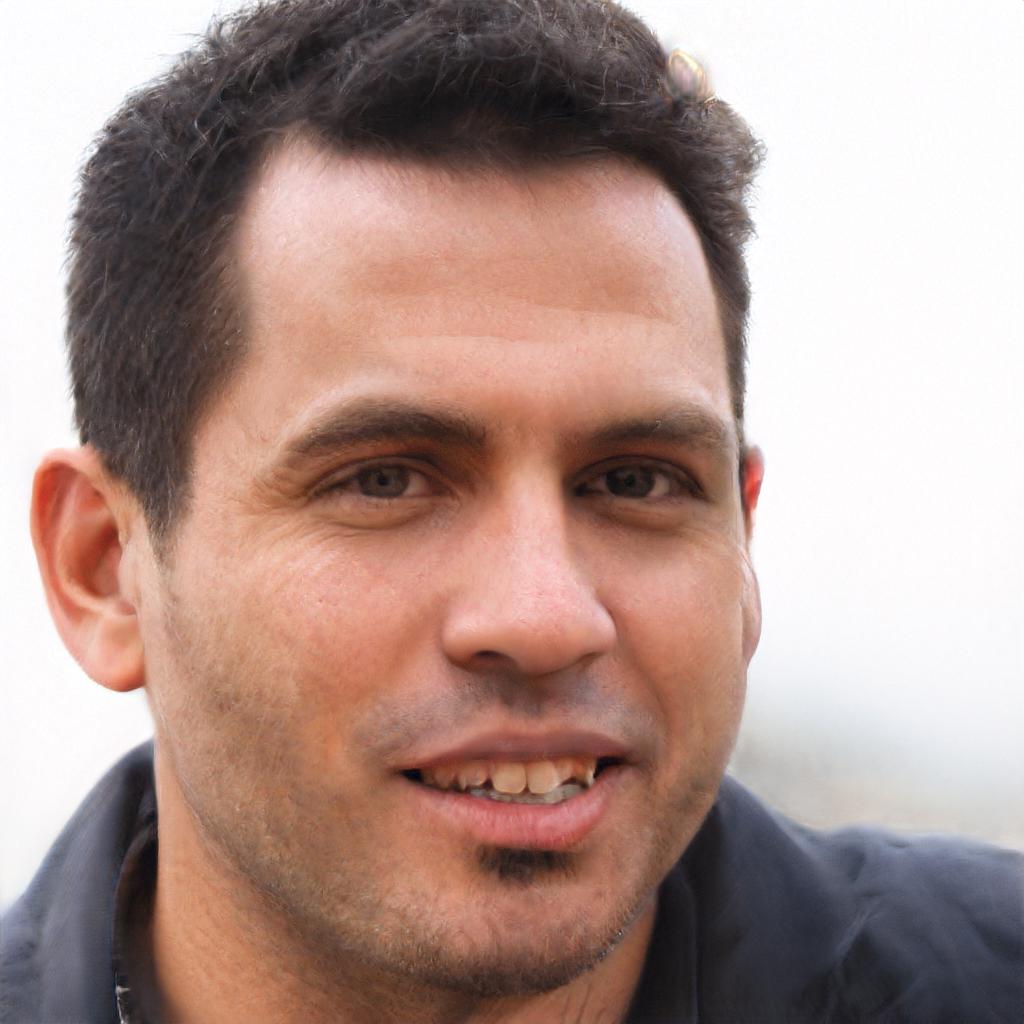}
		\end{tabular}}
		\caption{Our method localizes the edit made to the target image (top left) by conditioning the style transfer from the reference (top row) on a specific object of interest (left column). This gives users fine control over the appearance of objects in the synthesized images. Best viewed enlarged on screen.}
		\label{fig:main-ffhq}
	\end{figure*}

	\begin{figure*}
		{
			\centering \setlength{\tabcolsep}{0pt} 
			\figurefont
			\renewcommand{\arraystretch}{0.1}
			
			\newcommand\figpath{figures/PartGrid/bedrooms_6813/}
			\newcommand\partone{bed}
			\newcommand\parttwo{pillow}
			\newcommand\partthree{window}
			\begin{tabular}{M{0.2\textwidth}M{0.18\textwidth}M{0.18\textwidth}M{0.18\textwidth}M{0.18\textwidth}}
				\textbf{Target} & \textbf{Reference 1} & \textbf{Reference 2} & \textbf{Reference 3}  & \textbf{Reference 4} \\
				\includegraphics[width=0.18\textwidth]{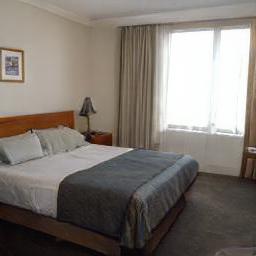}  &
				\includegraphics[width=0.18\textwidth]{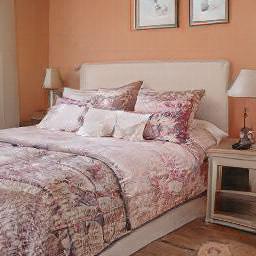} &
				\includegraphics[width=0.18\textwidth]{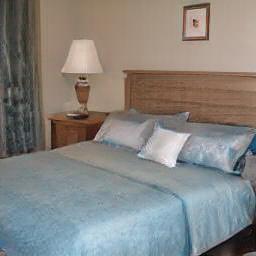} &
				\includegraphics[width=0.18\textwidth]{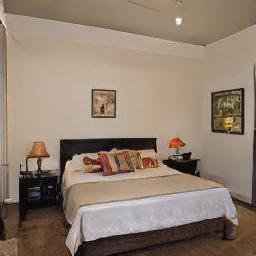} &
				\includegraphics[width=0.18\textwidth]{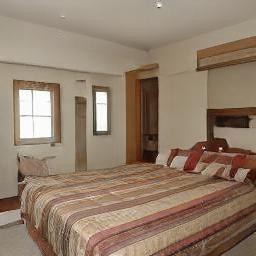}\\
				&&&& \\
				\textbf{\expandafter\MakeUppercase \partone:} &\includegraphics[width=0.18\textwidth]{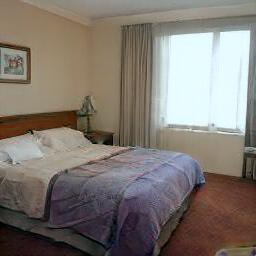} &
				\includegraphics[width=0.18\textwidth]{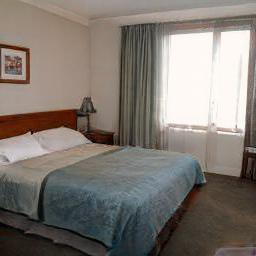} &
				\includegraphics[width=0.18\textwidth]{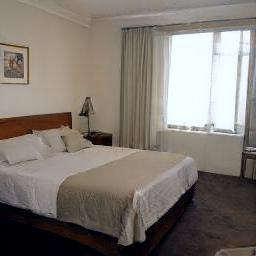} &
				\includegraphics[width=0.18\textwidth]{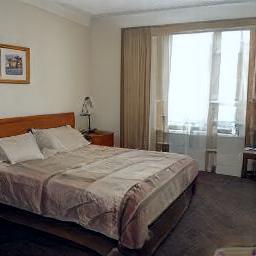}\\
				
				\textbf{\expandafter\MakeUppercase \parttwo:} &\includegraphics[width=0.18\textwidth]{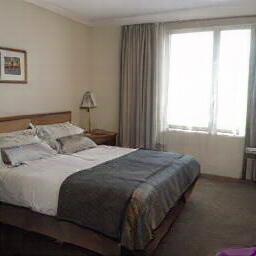} &
				\includegraphics[width=0.18\textwidth]{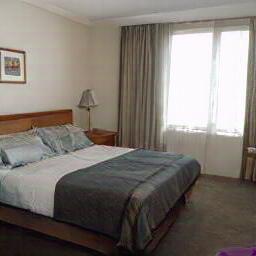} &
				\includegraphics[width=0.18\textwidth]{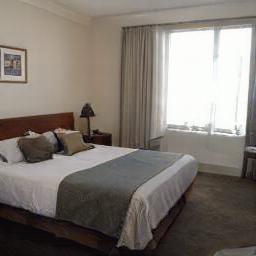} &
				\includegraphics[width=0.18\textwidth]{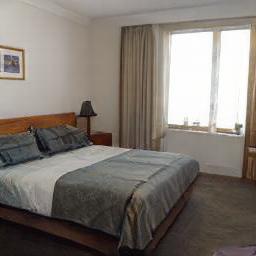}\\
				
				\textbf{\expandafter\MakeUppercase \partthree:} &\includegraphics[width=0.18\textwidth]{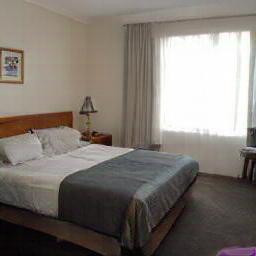} &
				\includegraphics[width=0.18\textwidth]{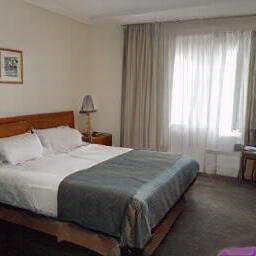} &
				\includegraphics[width=0.18\textwidth]{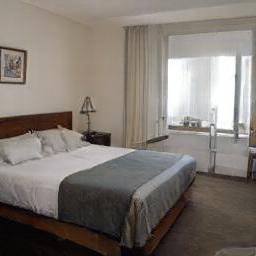} &
				\includegraphics[width=0.18\textwidth]{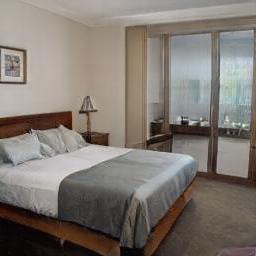}
		\end{tabular} }
		\caption{Unlike previous blending methods\cite{perez2003poisson,suzuki2018spatially}, our method does not require images to be aligned or of similar scale. In this case, the style of, e.g., the bed is successfully transferred from reference to target in spite of drastic changes in view point.}
		\label{fig:main-bedrooms}
	\end{figure*}
	
	\subsection{Local editing} \label{sec:Local editing}
	
	\subsubsection{StyleGAN overview} \label{sec:StyleGAN overview}
	We briefly review aspects of StyleGAN and StyleGAN2 relevant to our development. First, in order to generate a sample from the generator distribution, a latent vector $\vz$ is randomly sampled from the prior of the sampling space $\sZ$. Next, $\vz$ is transformed to an intermediate latent vector $\vw\in\sW$, that has been shown to exhibit better disentanglement properties \cite{karras2019style}, \cite{shen2019interpreting}. 
	
	The image generator is implemented as a convolutional neural network.
	Considering a batch consisting of a single image, let $\tA\in\sR^{(C\times H\times W)}$ be the input to a convolutional layer, which is assumed or explicitly normalized to have per-channel unit variance. Prior to the convolution operation, the vector $\vw$ alters the feature maps via a per-layer \emph{style}. Common to the application of style in both StyleGAN and StyleGAN2 is the use of per-channel \emph{scaling}, $\vsigma_{c}\tA_c$, where the layer-wise coefficients $\vsigma$ are obtained from a learned affine transformation of $\vw$.
	
	This style-based control mechanism is motivated by \emph{style transfer} \cite{gatys2016}, \cite{li2017}, where it has been shown that manipulating per-channel mean and variance is sufficient to control the style of an image \cite{huang2017arbitrary}. By fixing the input to the StyleGAN convolutional generator to be a constant image, the authors of StyleGAN showed that this mechanism is sufficient to determine all aspects of the generated image: the style at one layer determines the content at the next layer.

	\subsubsection{Conditioned interpolation} \label{sec:Conditioned interpolation}
	Given a target image $\mS$ and a reference image $\mR$, both GAN outputs, we would like to transfer the appearance of a specified local object or part from $\mR$ to $\mS$, creating the edited image $\mG$.
	Let $\vsigma^{\mS}$ and $\vsigma^{\mR}$ be two style scaling coefficients of the same layer corresponding to the two images.
	
	For global transfer, due to the properties of linearity and separability exhibited by StyleGAN's latent space, a mixed style $\vsigma^{\mG}$ produced by linear interpolation between $\vsigma^{\mS}$ and $\vsigma^{\mR}$\footnote{Karras et al. (2019) \cite{karras2019style} interpolate in the latent space of $\vw$, but the effect is similar.} produces plausible fluid morphings between the two images:
	\begin{align}
	\vsigma^{\mG} = \vsigma^{\mS} + \lambda(\vsigma^{\mR} - \vsigma^{\mS}) \label{eq:global-interp}
	\end{align}
	for $0\leq\lambda\leq1$. Doing so results in transferring \emph{all} the properties of $\vsigma^{\mR}$ onto $\vsigma^{\mG}$, eventually leaving no trace of $\vsigma^{\mS}$.
	
	To enable selective local editing, we control the style interpolation with a matrix transformation:
	\begin{align}
	\vsigma^{\mG} = \vsigma^{\mS} + \mQ(\vsigma^{\mR} - \vsigma^{\mS}) \label{eq:cond-interp}
	\end{align}
	where the matrix $\mQ$ is positive semi-definite and is chosen such that  $\vsigma^{\mG}$ effects a local style transfer from $ \vsigma^{\mR} $ to $ \vsigma^{\mS} $. In practice we choose $\mQ$ to be a diagonal matrix whose elements form $\vq\in\left[0,1\right]^{C}$, which we refer to as the query vector.
	
	\subsubsection{Choosing the query}
	For local editing, an appropriate choice for the query $\vq$ is one that favors channels that affect the region of interest (ROI), while ignoring channels that have an effect outside the ROI. When specifying the ROI using one of the semantic clusters computed in section  \ref{sec:Feature factorization}, say $k'$, the vector $\mM_{k',c}$ encodes exactly this information.
	
	A simple approach is to use $\mM_{k',c}$, computed offline from Eq. (\ref{eq:channel2cluster}) for a given genre and dataset of images, to control the slope of the interpolation, clipping at 1:
	\begin{align}
	\vq_c = \text{min}(1, \lambda\mM_{k',c}) \label{eq:simultaneous}
	\end{align}
	where $\vq_c$ is the $c$-th channel element of $\vq$, and $\lambda$, as in Eq. (\ref{eq:global-interp}), is the global strength of the interpolation. We refer to this approach as \emph{simultaneous} as it updates all channels at the same time. Intuitively, when $\lambda$ is small or intermediate, channels with large $\mM_{k',c}$ will have a higher weight, thus having an effect of localizing the interpolation.
	
	We propose an approach which achieves superior localization compared to Eq. (\ref{eq:simultaneous}), referred to as \emph{sequential}. We first set the most relevant channel to the maximum slope of 1, before raising the slope of the second-most relevant, third-most, etc. This definition of the query corresponds to solving for the following objective:
	\begin{align}
	\argmin_{\vq_c}~~ &\vq_c \left[\mM_{k',c} - \rho(1-\mM_{k',c})\right] \label{eq:sequential} \\ \nonumber
	\text{s.t.}~ &\sum_{c=1}^C \vq_c(1-\mM_{k',c})\leq \epsilon \\ &~~ 0\leq\vq_c\leq1 \nonumber
	\end{align}

	We solve this objective by sorting channels based on $\mM_{k'}$, 
	and greedily assigning $\vq_c = 1$ to the most relevant channels as long as the total effect outside the ROI is no more than some budget $\epsilon$. Additionally, a non-zero weight is only assigned to channels where $\mM_{k',c} > \frac{\rho}{1+\rho}$, which improves the robustness of local editing by ignoring irrelevant channels even when the budget $\epsilon$ allows more change.
	
	\begin{figure*}
		{\centering \setlength{\tabcolsep}{0pt} 
			\figurefont
			\renewcommand{\arraystretch}{0.1}
			\begin{tabular}{M{0.1\textwidth}M{0.18\textwidth}M{0.18\textwidth}M{0.18\textwidth}M{0.18\textwidth}}
				& & \textbf{Target} & \textbf{Reference} &   \\
				& & \includegraphics[width=0.18\textwidth]{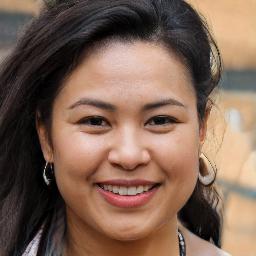}  &
				\includegraphics[width=0.18\textwidth]{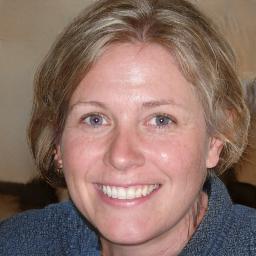} & \\
				
				&&&&\\[0.2cm]
				
				&  \textbf{Ours  ~~~~~~~~~~~~~~~~~~~~~~~~~~~~~~~~~~~~~~~}& \textbf{Feature blend $32\times32$}~~\cite{suzuki2018spatially} & \textbf{Feature blend $64\times64$}~~\cite{suzuki2018spatially} & \textbf{Poisson blend} ~~~~~~~~~~\cite{perez2003poisson} \\[0.5ex]
				
				\textbf{Eyes:} &\includegraphics[width=0.18\textwidth]{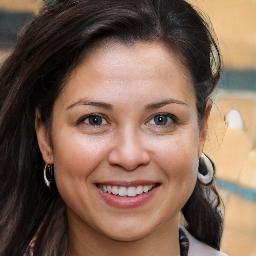} &
				\includegraphics[width=0.18\textwidth]{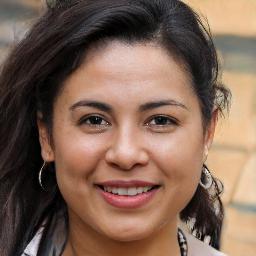} &
				\includegraphics[width=0.18\textwidth]{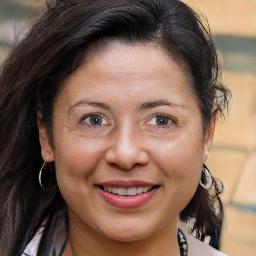} &
				\includegraphics[width=0.18\textwidth]{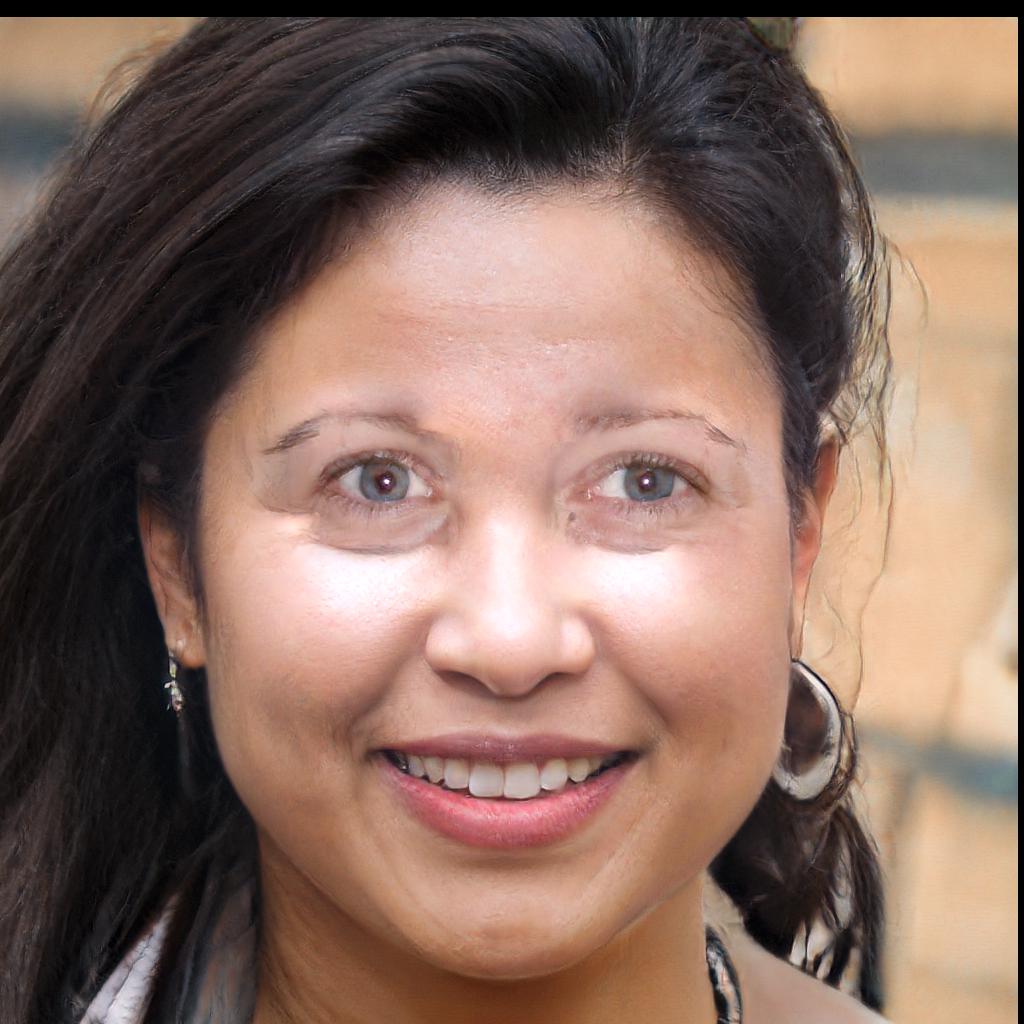}\\
				
				\textbf{Nose:} &\includegraphics[width=0.18\textwidth]{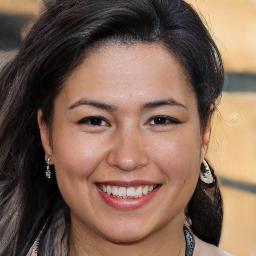} &
				\includegraphics[width=0.18\textwidth]{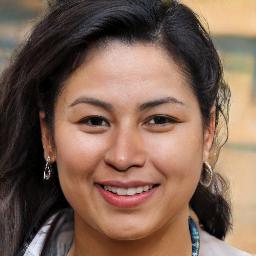} &
				\includegraphics[width=0.18\textwidth]{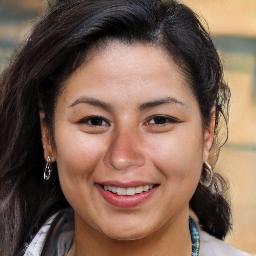} &
				\includegraphics[width=0.18\textwidth]{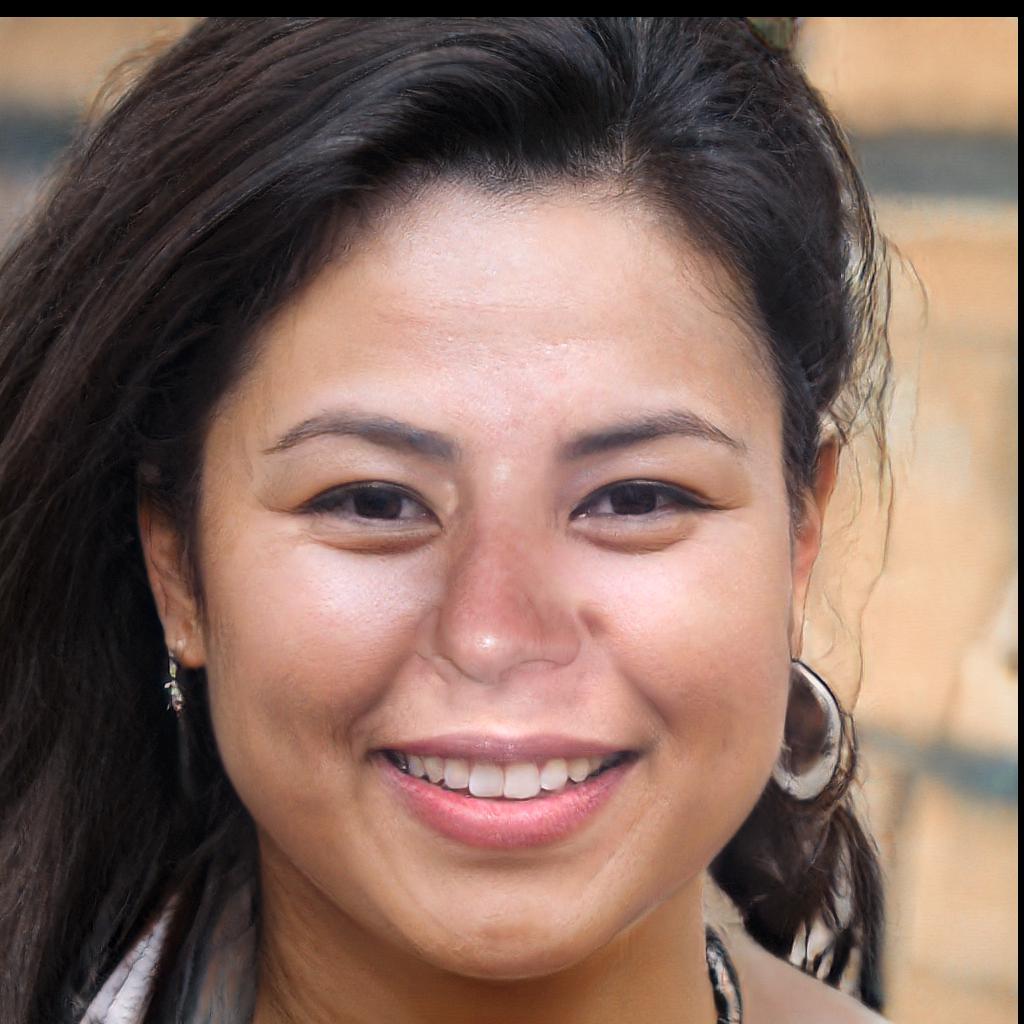}\\
				
				\textbf{Mouth:} &\includegraphics[width=0.18\textwidth]{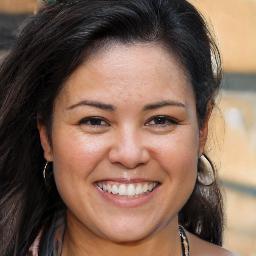} &
				\includegraphics[width=0.18\textwidth]{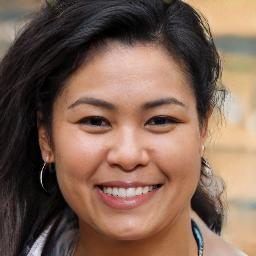} &
				\includegraphics[width=0.18\textwidth]{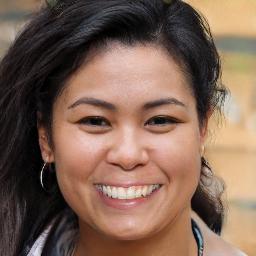} &
				\includegraphics[width=0.18\textwidth]{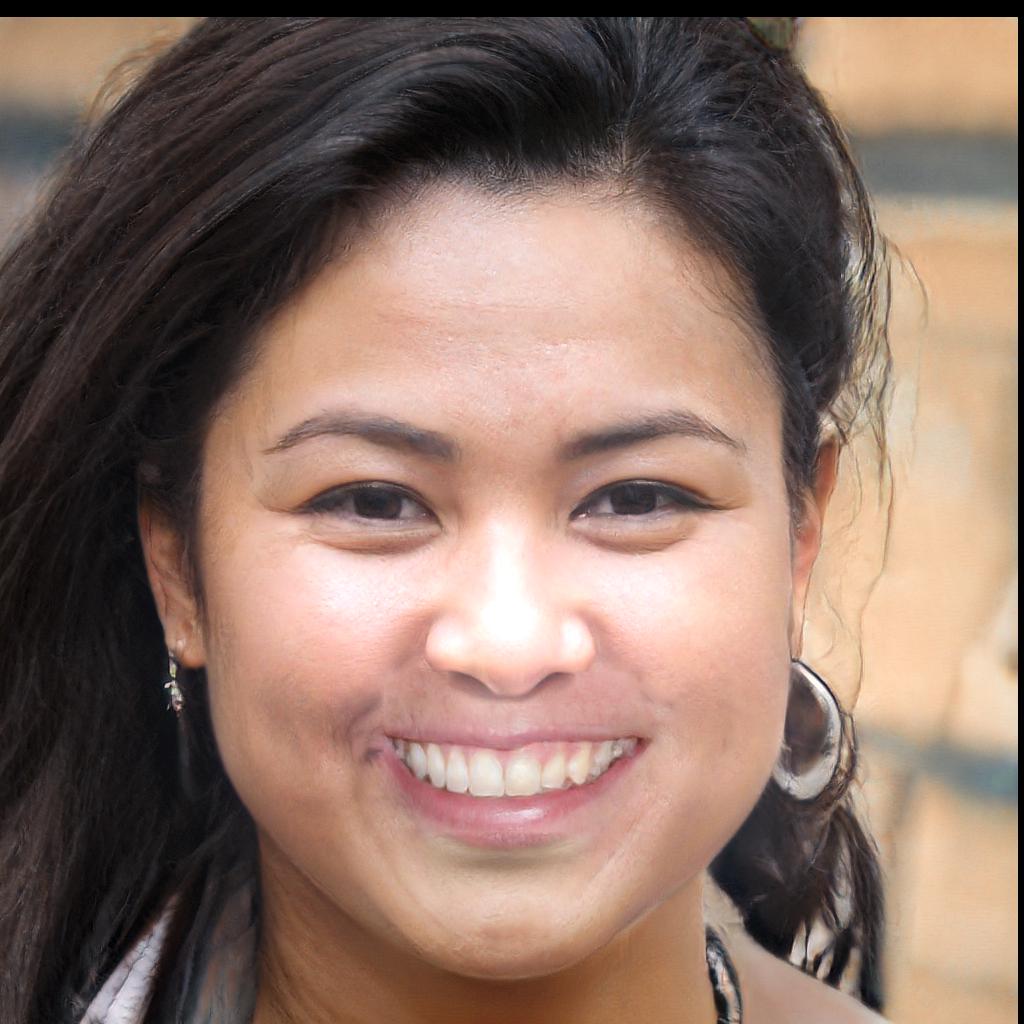}
		\end{tabular}}
		\caption{Even the well aligned FFHQ-generated faces prove challenging for existing blending methods, as they do not consider differences in pose and scale, and lack any notion of semantics or photorealism. In contrast, our method makes use of the correlation GANs learn from real data to maintain a natural appearance, while exploiting feature disentanglement for effectively localizing the change.}
		\label{fig:copy-paste-comapre}
	\end{figure*}

	\begin{figure*}
		{\centering \setlength{\tabcolsep}{0pt} 
			\figurefont
			\renewcommand{\arraystretch}{0.1}
			\begin{tabular}{M{0.2\textwidth}M{0.18\textwidth}M{0.18\textwidth}M{0.18\textwidth}M{0.18\textwidth}}
				\textbf{Target} & \textbf{Reference 1} & \textbf{Reference 2} & \textbf{Reference 3}  & \textbf{Reference 4} \\
				\includegraphics[width=0.18\textwidth]{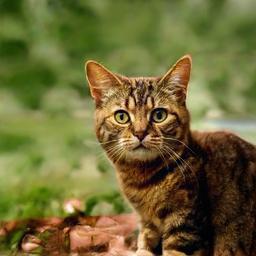}  &
				\includegraphics[width=0.18\textwidth]{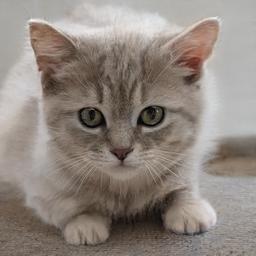} &
				\includegraphics[width=0.18\textwidth]{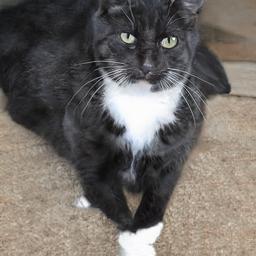} &
				\includegraphics[width=0.18\textwidth]{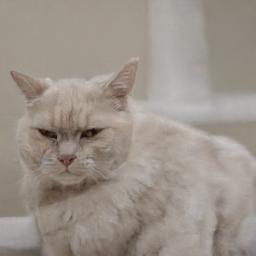} &
				\includegraphics[width=0.18\textwidth]{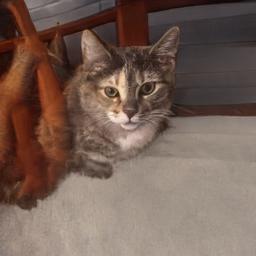}\\
				
				\textbf{Eyes:} &\includegraphics[width=0.18\textwidth]{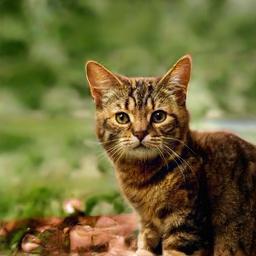} &
				\includegraphics[width=0.18\textwidth]{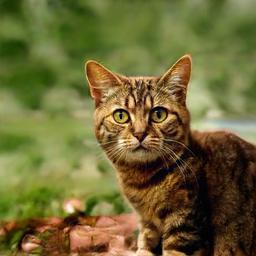} &
				\includegraphics[width=0.18\textwidth]{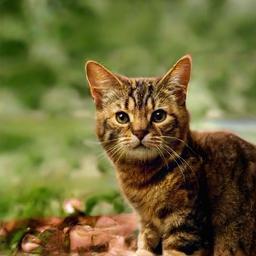} &
				\includegraphics[width=0.18\textwidth]{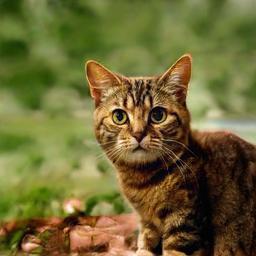}\\
				
				&&&& \\[0.3cm]
				
				\includegraphics[width=0.18\textwidth]{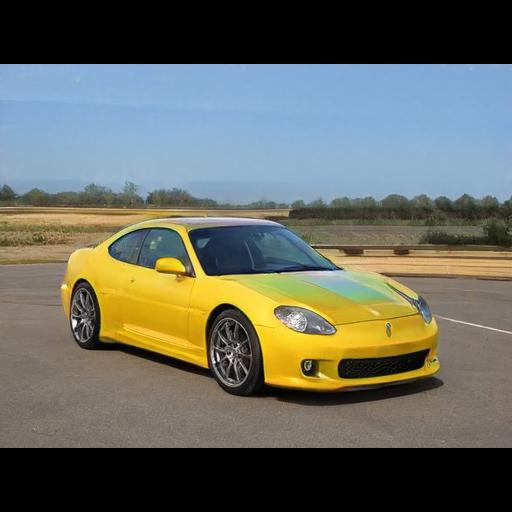}  &
				\includegraphics[width=0.18\textwidth]{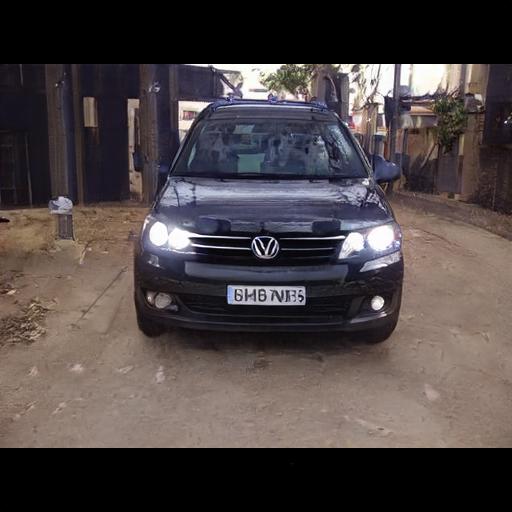} &
				\includegraphics[width=0.18\textwidth]{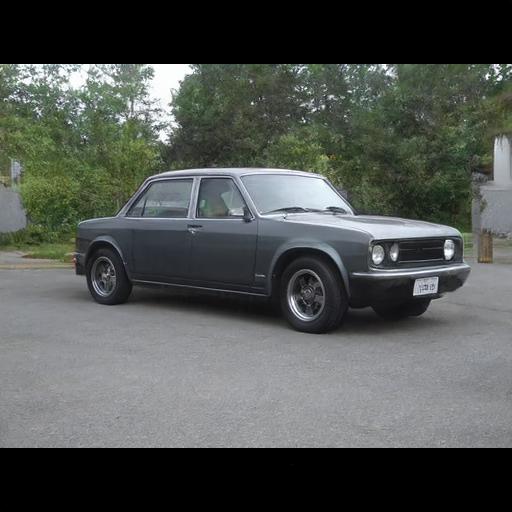} &
				\includegraphics[width=0.18\textwidth]{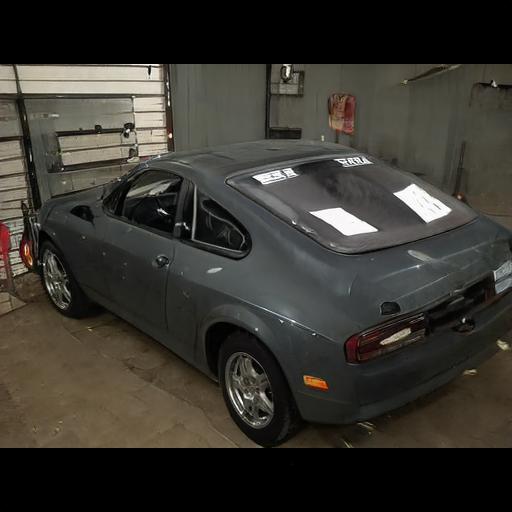} &
				\includegraphics[width=0.18\textwidth]{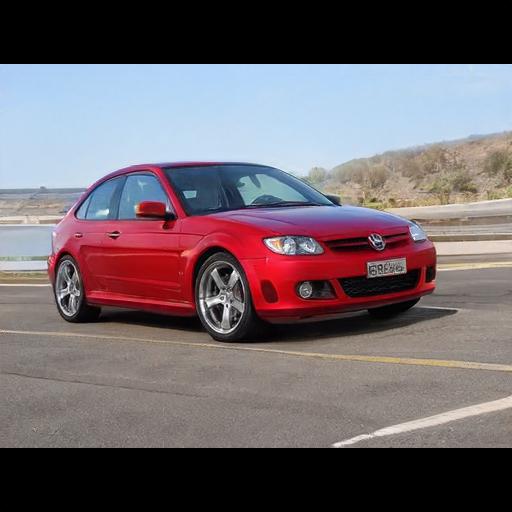}\\
				
				\textbf{Wheels:} &\includegraphics[width=0.18\textwidth]{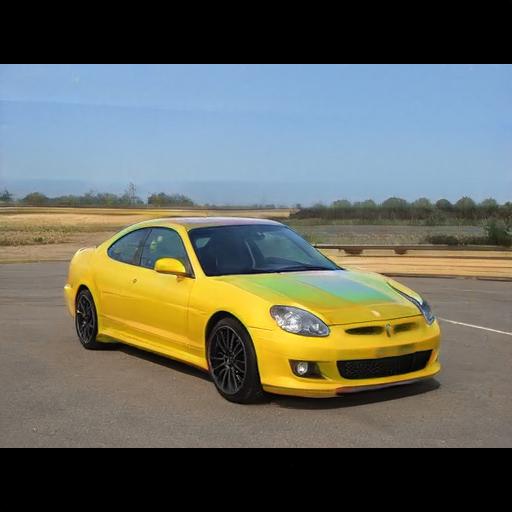} &
				\includegraphics[width=0.18\textwidth]{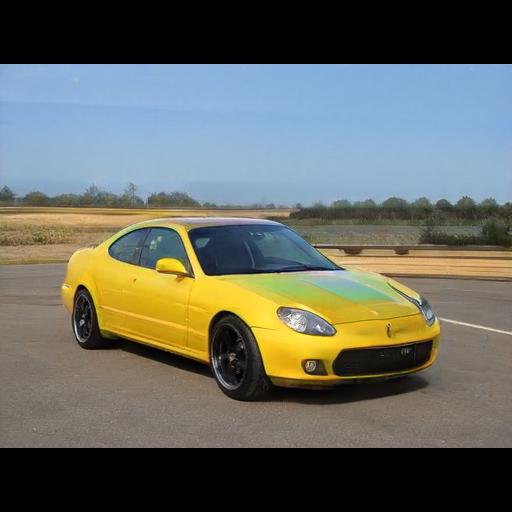} &
				\includegraphics[width=0.18\textwidth]{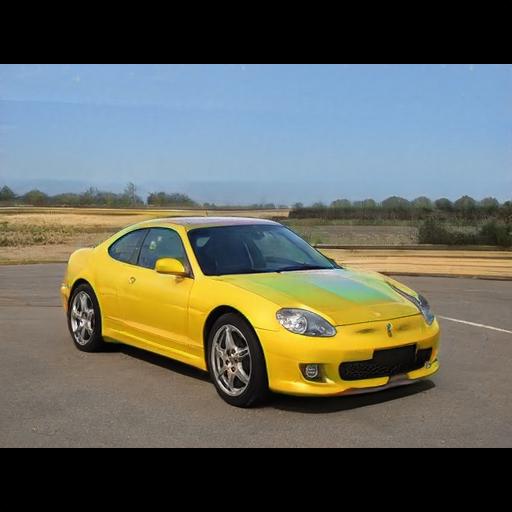} &
				\includegraphics[width=0.18\textwidth]{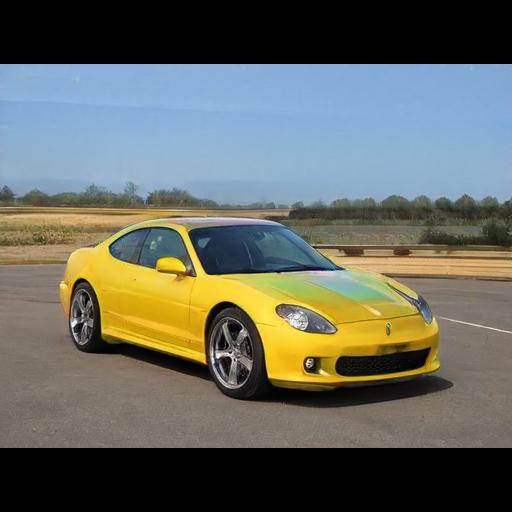}\\

		\end{tabular}}
		\caption{Our method applied to StyleGAN2 outputs. Photorealism is preserved while allowing fine control over highly-localized regions.}
		\label{fig:main-stylegan2}
	\end{figure*}
	
	\begin{figure*}
		{\centering \setlength{\tabcolsep}{0pt} 
			\begin{tabular}{cccc}
				\centering
				~~~~~\includegraphics[align=c,width=0.14\textwidth]{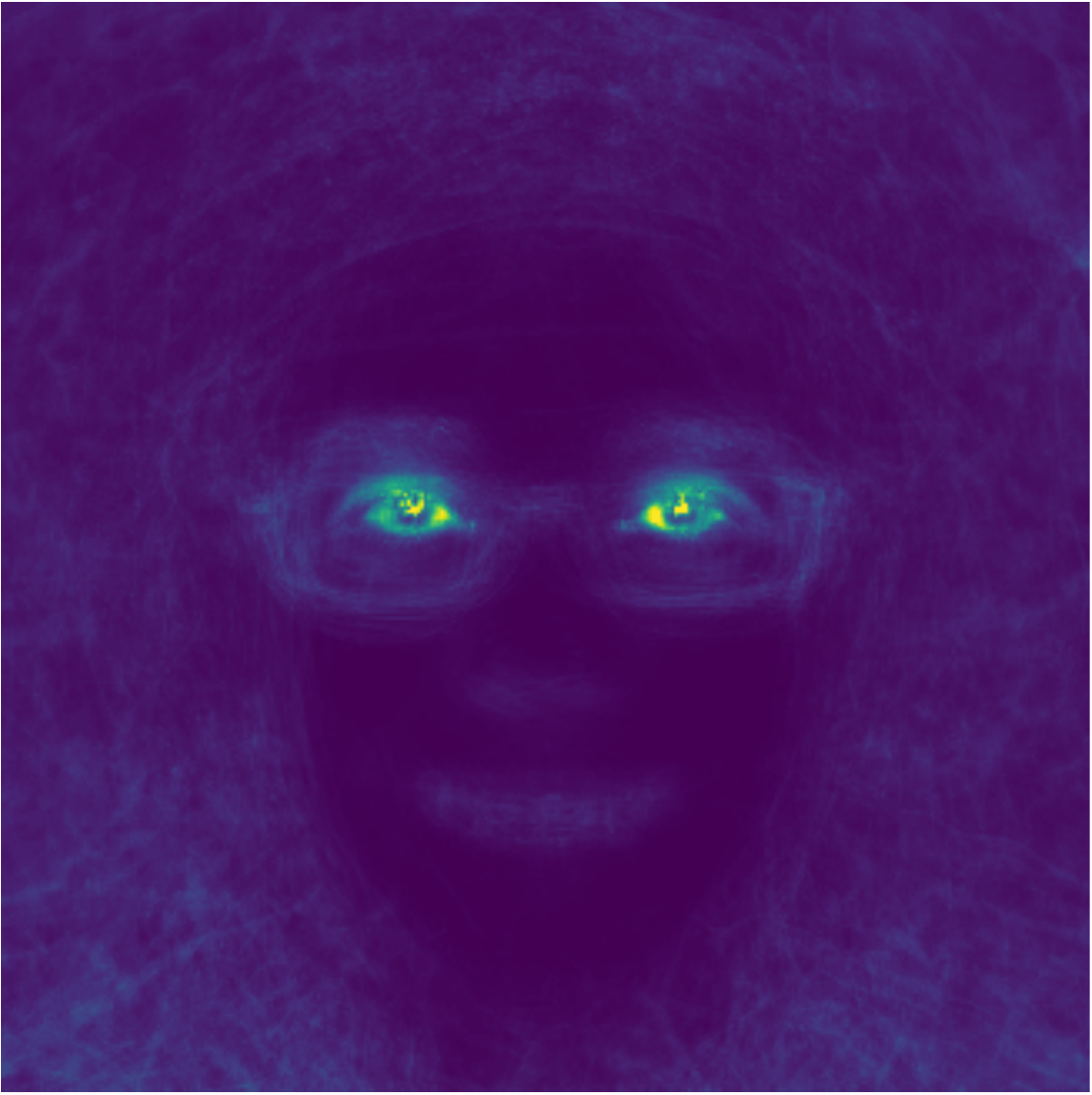}~%
				\includegraphics[align=c,width=0.14\textwidth]{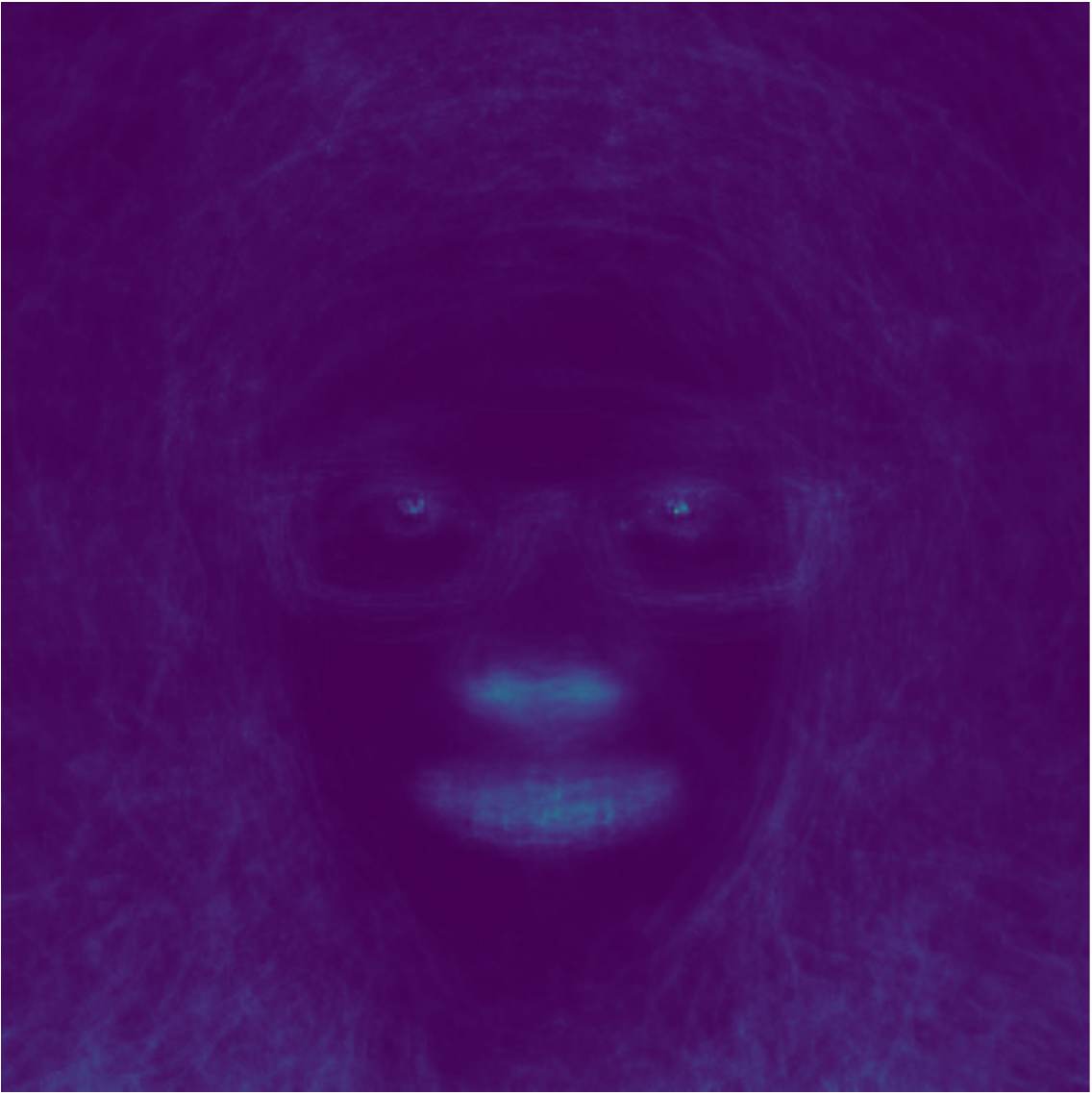}~%
				\includegraphics[align=c,width=0.14\textwidth]{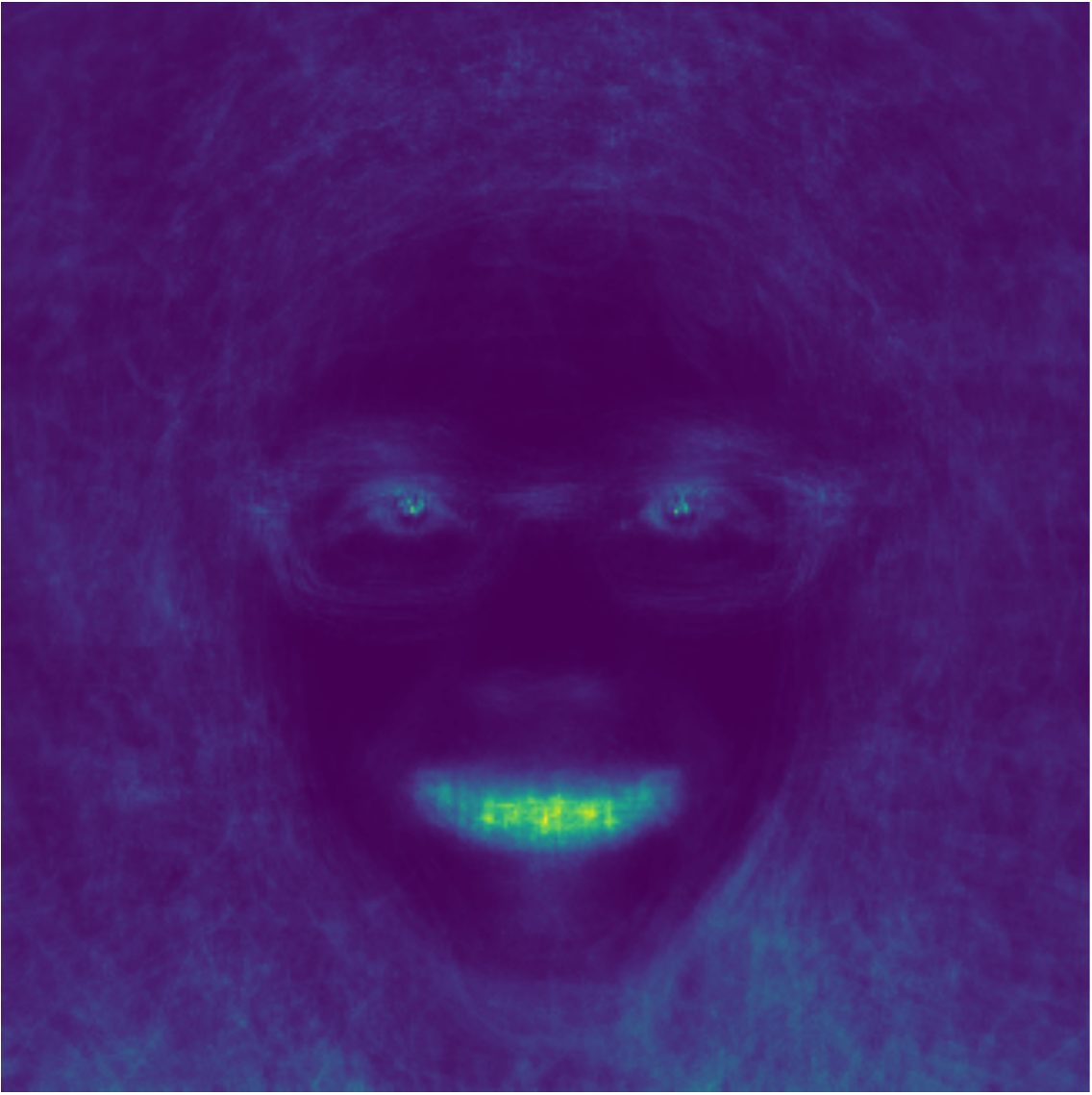}~~%
				\includegraphics[align=c,width=0.4cm]{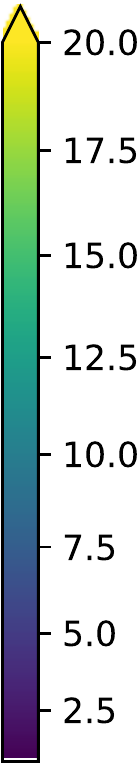}~~~~~~~~~
				&
				\rotatebox[origin=c]{90}{In-MSE}~~~
				&
				\includegraphics[align=c,width=0.2\textwidth]{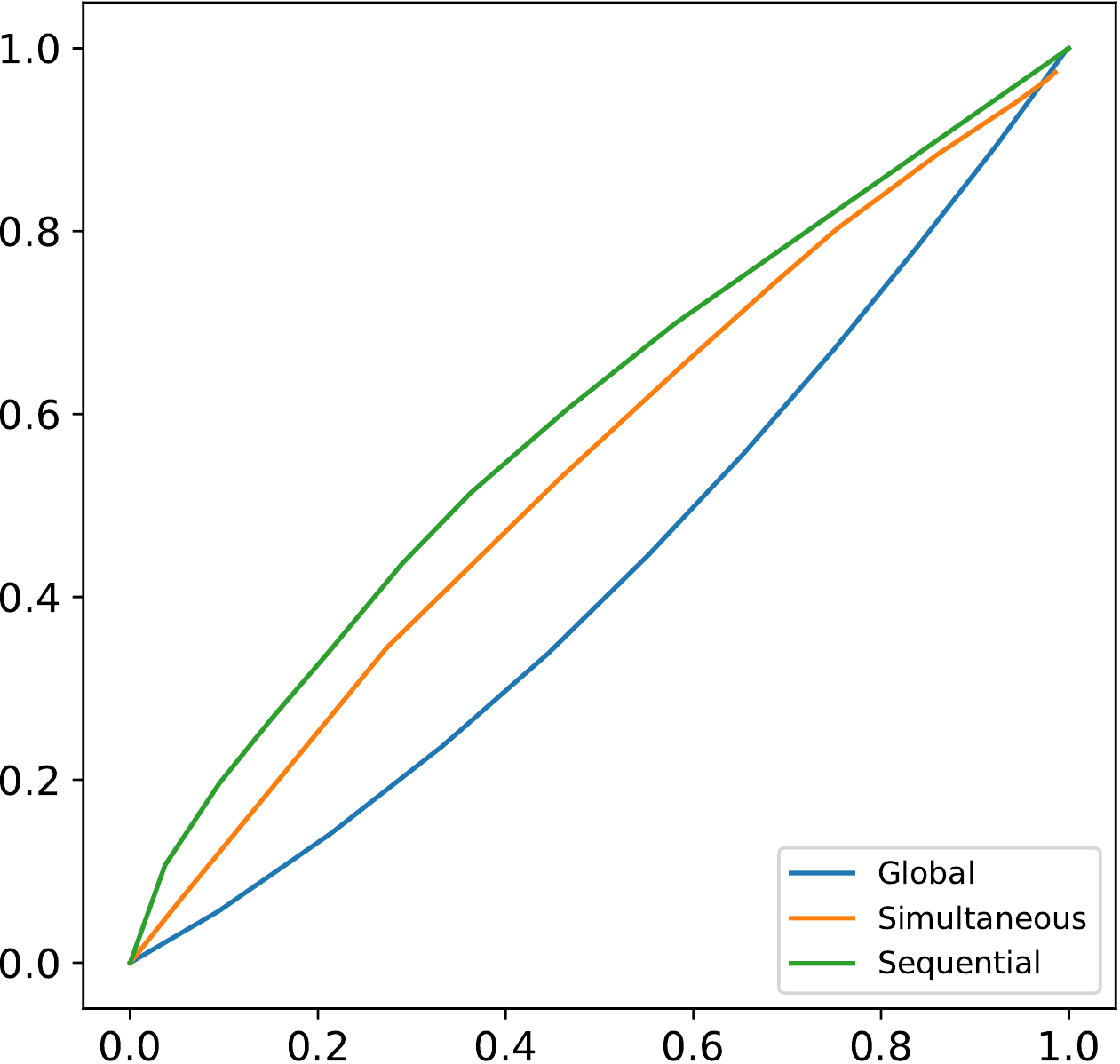}~~~
				&
				\includegraphics[align=c,width=0.2\textwidth]{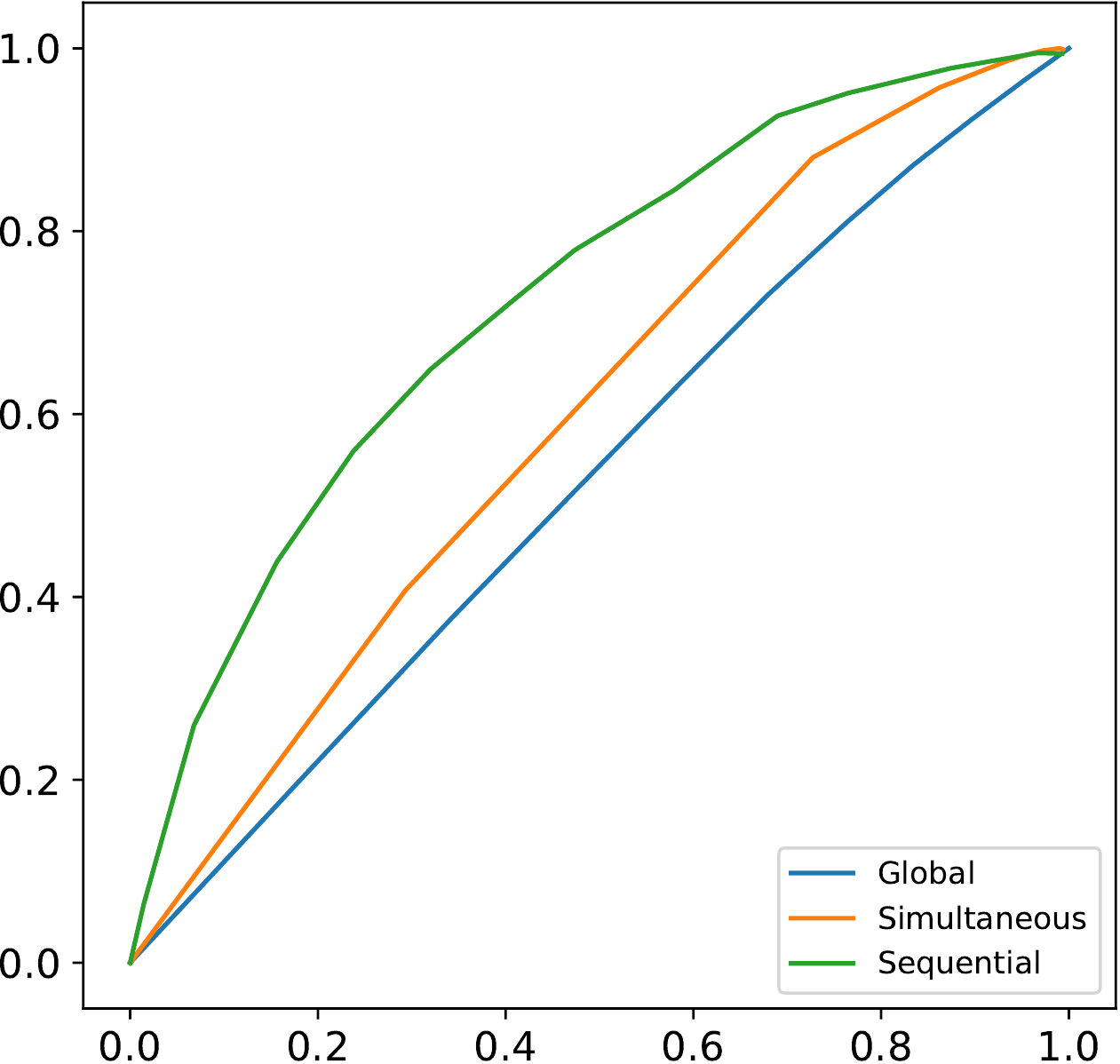}\vspace{5pt}\\
				& & \multicolumn{2}{c}{Out-MSE}\\
				(a) & & (b) & (c)
		\end{tabular}}
		\caption{(a) Mean squared-error (MSE) heatmaps computed between 50K FFHQ-StyleGAN outputs and their edited counterparts for \emph{eyes} (left), \emph{nose} (center) and \emph{mouth} (right). These heatmaps demonstrate that our method produces edits that are both perceptible and localized. (b) MSE \textbf{in}side the ROI vs. \textbf{out}side for various query parameters on FFHQ and (c) similarly for LSUN-Bedrooms. Our approach, \emph{sequential}, has the best trade-off between In-MSE (high is good) and Out-MSE (low is good).}
		\label{fig:locality}
	\end{figure*}
	
	\section{Experiments} \label{sec:Experiments}
	
	\subsection{Qualitative evaluation}
	In Figs. \ref{fig:main-ffhq} and \ref{fig:main-bedrooms} we demonstrate our editing method\footnote{Our code is available online at: \url{https://github.com/IVRL/GANLocalEditing}} with StyleGAN generators trained on two datasets: FFHQ \cite{karras2019style} comprising 70K facial images and LSUN-Bedrooms \cite{yu15lsun} comprising about 3M color images depicting bedrooms.
	
	In both datasets, we found the first $32\times 32$ resolution layer of the generator to be ``most semantic''. We therefore chose this layer to apply spherical {\tt k}-means clustering. We set $\rho$ such that $\frac{\rho}{1+\rho}=0.1$ and tune $20\leq\epsilon\leq100$ for best performance. We found that the tuning of $\epsilon$ depends mostly on the target image and object of interest, and not the style reference. Note that by nature of the \emph{local} edit, changes to the target image may be subtle, and best viewed on screen.
	
	Fig. \ref{fig:copy-paste-comapre} compares our method with feature-level blending \cite{suzuki2018spatially} and pixel-level (Poisson) blending \cite{perez2003poisson} methods. Feature blending is applied once to all layers of resolution $32\times32$ or lower, and once to those of $64\times 64$ or lower. While these approaches are strictly localized (see section \ref{sec:GAN-based Image Editing}), their outputs lack photorealism. For instance, the target and reference faces are facing slightly different directions, which causes a misalignment problem most visible in the \emph{nose}. In contrast, our editing method primarily affects the ROI, and yet maintains the photorealism of the baseline GAN by admitting \emph{some} necessary global changes. However, our method does not always copy the appearance of an object 'faithfully', as seen in the \emph{window} row of Fig. \ref{fig:main-bedrooms}.
	
	Fig. \ref{fig:main-stylegan2} demonstrates the applicability of our method to the recent StyleGAN2 model \cite{karras2019analyzing} trained on LSUN-Cats and LSUN-Cars \cite{yu15lsun}. Unlike traditional blending methods, our technique is able to transfer parts between unaligned images as seen here and in Fig. \ref{fig:main-bedrooms}.

	\subsection{Quantitative analysis}    
	
	We quantitatively evaluate the results of editing on two aspects of interest: locality and photorealism.
	
	\subsection{Locality}
	To evaluate the locality of editing, we examine the squared-error in pixel space between target images and their edited outputs. Fig. \ref{fig:locality} (a) shows the difference between unedited and edited images averaged over 50K FFHQ-StyleGAN samples, where at every pixel location we compute the squared distance in CIELAB color space. This figure indicates that the transfers are both perceptible and localized, and that not all object parts are equally disentangled. Compared to \emph{eyes} and \emph{mouth}, where edits are very localized, editing the \emph{nose} seems to force a subtle degree of correlation with the other face parts. Such correlations trade-off control on the appearance of individual parts versus plausibility and realism of the overall output.
	
	We further examine the localization ability of our method and variants described in Section \ref{sec:Local editing}. First, we obtain for each image the binary mask indicating the ROI, using the pre-computed spherical {\tt k}-means clusters of Section \ref{sec:Feature factorization}.
	Then, we perform interpolation with various values of $\lambda$ (Eqs. \ref{eq:global-interp} and \ref{eq:simultaneous}) and $\epsilon$ (Eq. \ref{eq:sequential}). For each such setting we measure the (normalized) In- and Out-MSE of each target-output pair, i.e., the MSE inside the ROI and MSE outside the ROI, respectively. 
	In Fig. \ref{fig:locality} (b) and (c), we show that for both FFHQ and LSUN-Bedrooms, respectively, our method (\emph{sequential)} has better localization, i.e., less change outside the ROI for the same amount of change inside the ROI.
	
	\subsection{Photorealism}
	Measuring photorealism is challenging, as there is not yet a direct computational way of assessing the \emph{perceived} photorealism of an image. The Fréchet Inception Distance (FID), however, \cite{heusel2017gans} has been shown to correlate well with human judgement and has become a standard metric for GAN evaluation.
	
	An aggregate statistic, FID compares the distributions of two image sets in the feature space of a deep CNN layer. In Table
	\ref{tab:FID} we report the FID of 50K edited images against the original FFHQ and LSUN-Bedrooms datasets. The FID scores indicate that our edited images are not significantly different from the vanilla output of the baseline GAN.
	
	However, the same result was achieved when we computed the FID of 50K FFHQ images edited with feature blending \cite{suzuki2018spatially}, although Fig. \ref{fig:copy-paste-comapre} shows qualitatively that these produced outputs lack photorealism. This reemphasizes the difficulty of correctly measuring photorealism in an automated way.
	We did not run a similar analysis with Poisson blending since the many failure cases we observed with this approach did not justify the heavy computational cost required to process a large collection of $1024\times1024$ images. For both feature blending and Poisson editing, we could not test the Bedrooms dataset since these methods are not suitable for unaligned image pairs.
	
	\begin{table}
		\centering
		\setlength{\tabcolsep}{5pt}
		\begin{tabular}{lcc}
			FID & FFHQ & Bedrooms \\ \hline
			
			StyleGAN\cite{karras2019style} & 4.4 & 2.8 \\
			Ours & 5.4 & 4.5 \\
			Feature blending $64\times64$\cite{suzuki2018spatially} & 5.4 & - \\
		\end{tabular}\\[2ex]
		\caption{Fréchet Inception Distance between ground truth images and outputs of StyleGAN, our method, and feature blending. In the case of LSUN-Bedrooms, the non-aligned nature of the images makes feature blending inapplicable.} \label{tab:FID}
	\end{table}

	\section{Conclusion} \label{sec:Conclusion}
	We have demonstrated that StyleGAN's latent representations spatially disentangle semantic objects and parts. We leverage this finding to introduce a simple method for local semantic part editing in StyleGAN images. The core idea is to let the latent object representation guide the style interpolation to produce realistic part transfers without introducing any artifacts not already inherent to StyleGAN. The locality of the result depends on the extent to which an object's representation is disentangled from other object representations, which in the case of StyleGAN is significant. Importantly, our technique does not involve external supervision by semantic segmentation models, or complex spatial operations to define the edit region and ensure a seamless transition from edited to unedited regions. 
	
	For future investigation, our observations open the door to explicitly incorporate editing capabilities into the adversarial training itself, which we believe will improve the extent of disentanglement between semantic objects, and yield even better localization.
	
	Finally, the method can, in principle, be extended to semantic editing of real images by leveraging the frameworks of \cite{abdal2019}, \cite{karras2019analyzing} to first map natural images into the latent space of StyleGAN. This opens up interesting applications in photo enhancement, augmented reality, visualization for plastic surgery, and privacy preservation.

	{\small
		\bibliographystyle{ieee_fullname}
		\bibliography{egbib}
	}
	
	\clearpage
	\appendix
	\twocolumn[{%
		\renewcommand\twocolumn[1][]{#1}%
		\begin{center}
			\centering
			{\Huge Appendix}\\[0.5ex]
			\includegraphics[width=0.95\textwidth]{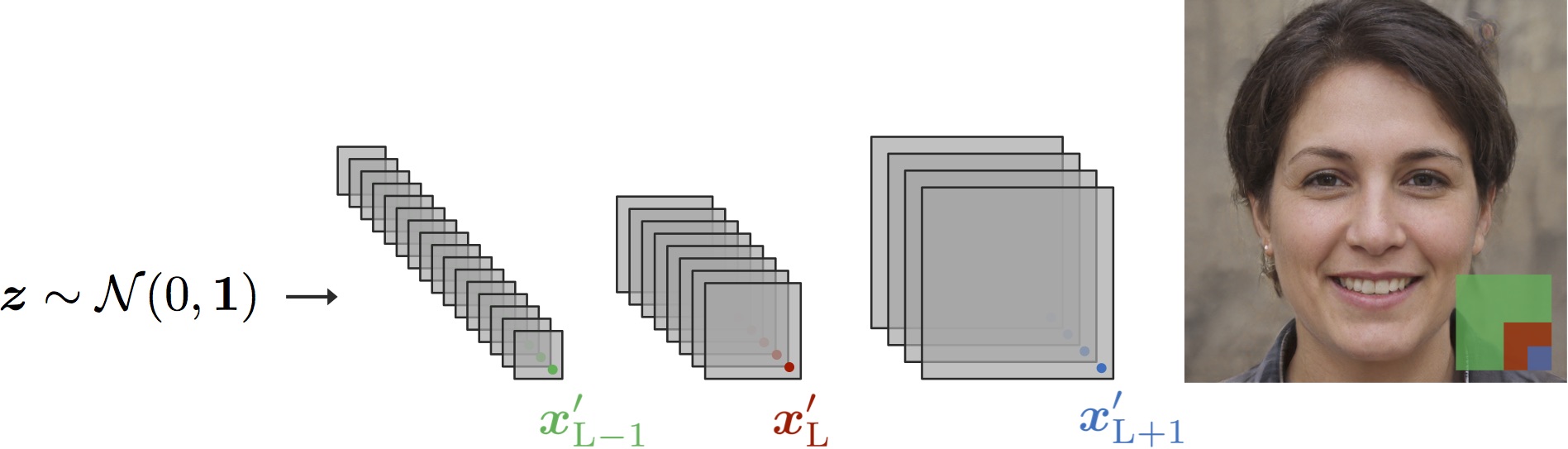}%
			\vspace{5pt}
			\captionsetup{type=figure}
			\caption{In convolutional generative networks, the vector $x'$ at a single spatial position on a hidden feature map at some layer $\mathrm{L}$ corresponds to a whole \emph{patch} on the RGB image. The lower the resolution of the hidden layer, the larger the patch in pixel space.} \label{fig:patch-embeddings}
			\vspace{5pt}
			
		\end{center}%
	}]

	\section{Spherical {\tt k}-means for semantic clustering} \label{sec:kmeans}
	
	In this section we elaborate on the layer-wise analysis described in Section \ref{sec:Local Semantics}.
	
	For a hidden layer $\mathrm{L}$ with $C$ channels, let $\tA\in\sR^{(N\times C\times H\times W)}$ be a tensor of zero-mean unit-variance activations for a batch of $N$ images, where at each channel the feature map has spatial dimensions $H\times W$.
	
	As show in Fig. \ref{fig:patch-embeddings}, a vector $\va\in\sR^{C}$ sampled at a single spatial location on $\tA$ represents a whole patch (e.g., $32\times32$) in the RGB image, and acts as a \emph{patch embedding}.
	
	We apply spherical {\tt k}-means to this $C$-dimensional space by first partially flattening $\tA$ to $\mA\in\sR^{(N\cdot H\cdot W)\times C}$, i.e., to a ``bag-of-patch-embeddings'' representation, with no explicit encoding of spatial position or even partitioning into different samples of the batch.
	The process can thus be viewed as clustering patches whose embeddings at layer $\mathrm{L}$ are similar, in the cosine similarity sense.
	
	Performing spherical {\tt k}-means with $K$ clusters can be viewed as a matrix factorization $\mA\approx\mU\mV$, where the binary matrix $\mU\in\{0,1\}^{(N\cdot H\cdot W)\times K}$ encodes cluster membership and the matrix $\mV\in\sR^{K\times C}$ encodes the unit-length centroid of each cluster.
	
	The matrix $\mU$ can be reshaped to a tensor $\tU\in\{0,1\}^{N\times K\times H\times W}$ which represents $K$ sets of $N$ \emph{masks} (one per image), where each mask spatially shows the cluster memberships.
	
\begin{figure*}
		\centering
		{\small \rotatebox[origin=c]{90}{RGB output}} \includegraphics[align=c,width=0.89\textwidth]{figures/kmeans/ffhq_input.jpg}\\
		{\small \rotatebox[origin=c]{90}{Layer 3, $K=20$}} \includegraphics[align=c,width=0.89\textwidth]{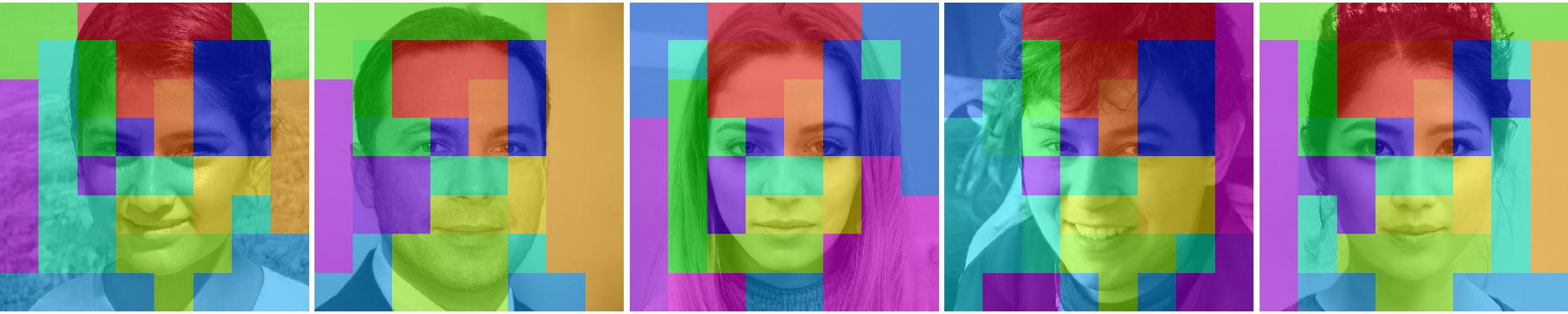}\\
		{\small \rotatebox[origin=c]{90}{Layer 4, $K=15$}}
		\includegraphics[align=c,width=0.89\textwidth]{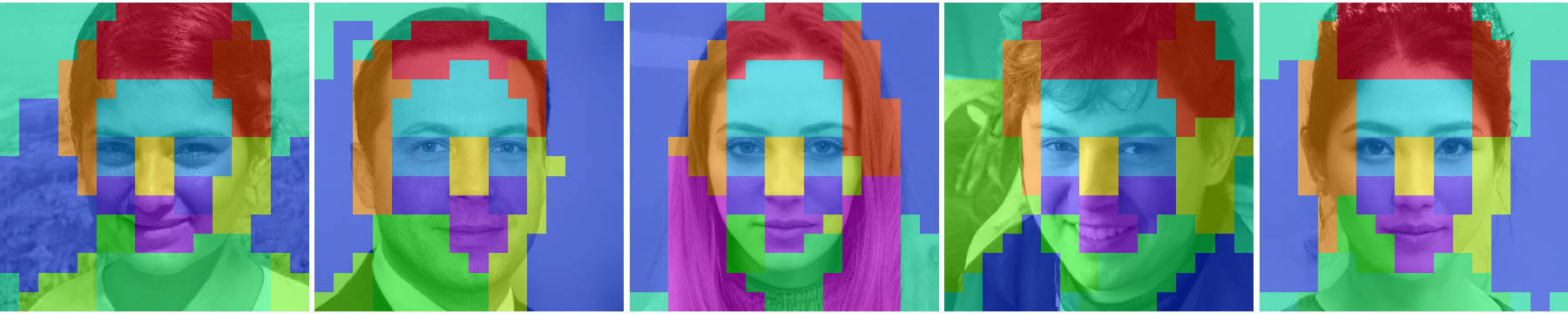}\\
		{\small \rotatebox[origin=c]{90}{Layer 5, $K=15$}}
		\includegraphics[align=c,width=0.89\textwidth]{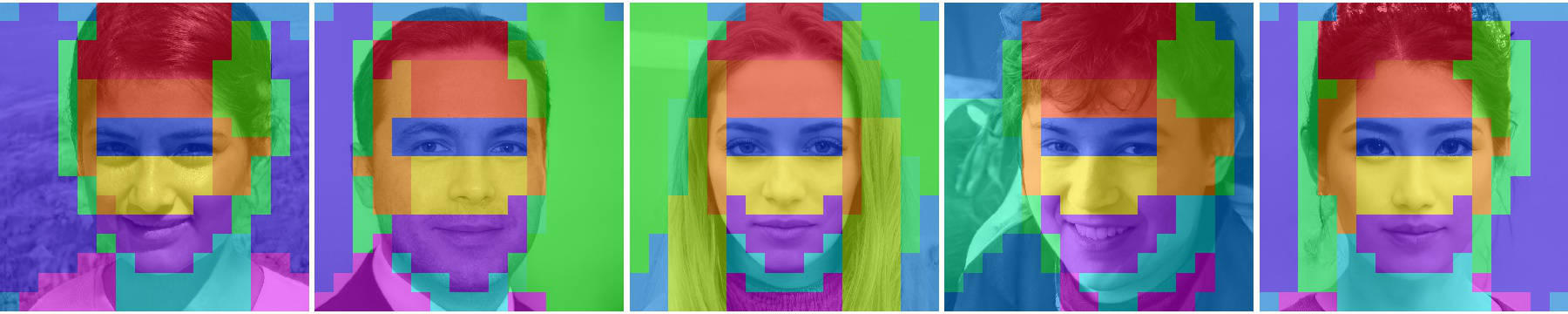}\\
		{\small \rotatebox[origin=c]{90}{Layer 6, $K=15$}}
		\includegraphics[align=c,width=0.89\textwidth]{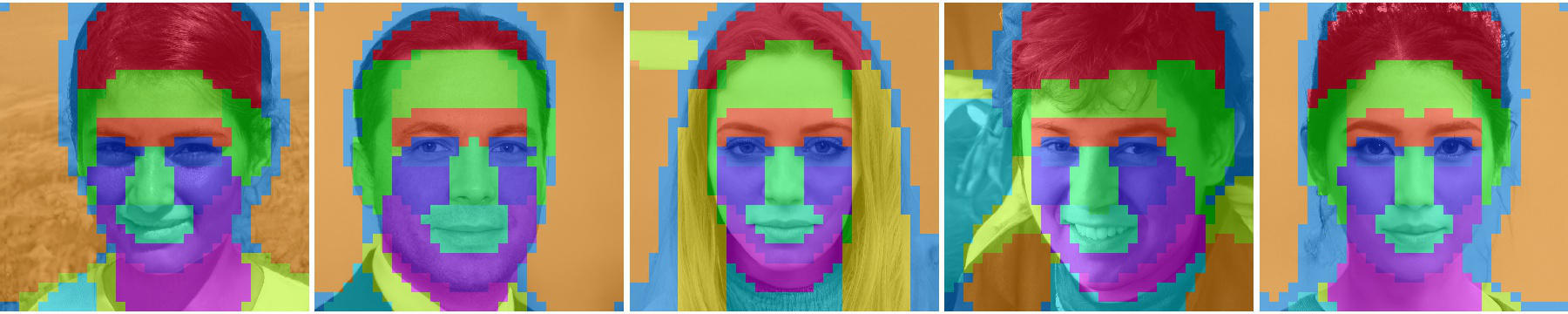}\\
		{\small \rotatebox[origin=c]{90}{Layer 8, $K=15$}}
		\includegraphics[align=c,width=0.89\textwidth]{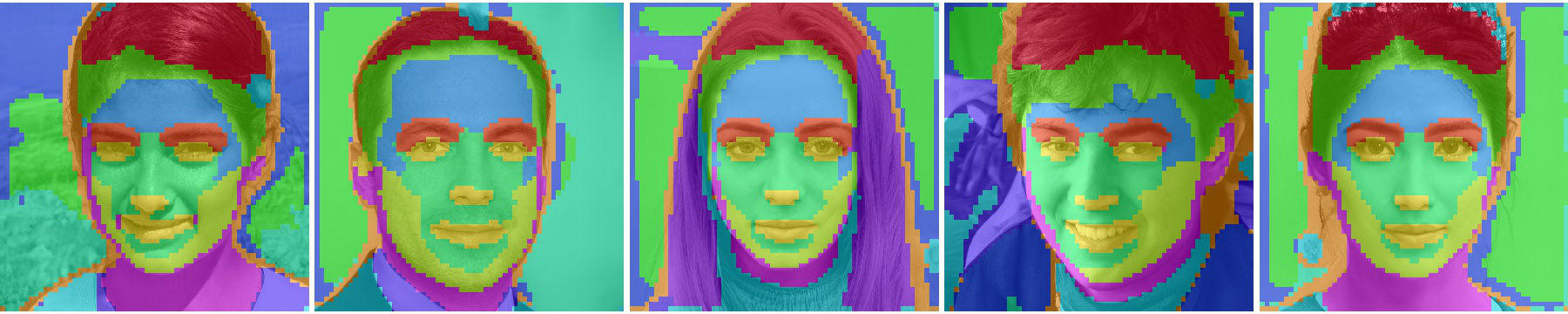}\\
		{\small \rotatebox[origin=c]{90}{Layer 11, $K=15$}}
		\includegraphics[align=c,width=0.89\textwidth]{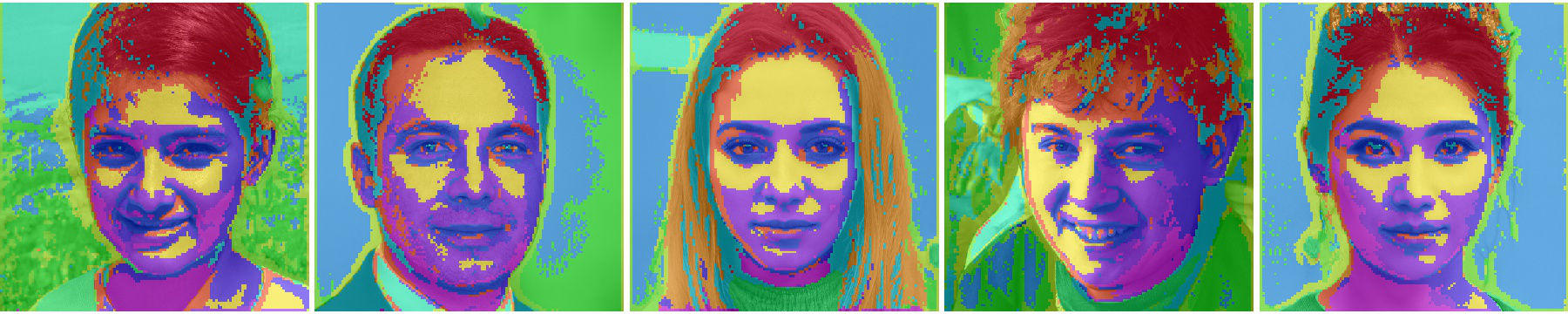}\\
		\caption{Spherical {\tt k}-means cluster membership maps for various FFHQ-StyleGAN layers. Color-coding signifies different clusters, and is arbitrarily determined per layer.}
		\label{fig:kmeans-SG-FFHQ}
	\end{figure*}

\begin{figure*}
		\centering
		{\small \rotatebox[origin=c]{90}{RGB output}} \includegraphics[align=c,width=0.89\textwidth]{figures/kmeans/bedroom_input.jpg}\\
		{\small \rotatebox[origin=c]{90}{Layer 3, $K=10$}} \includegraphics[align=c,width=0.89\textwidth]{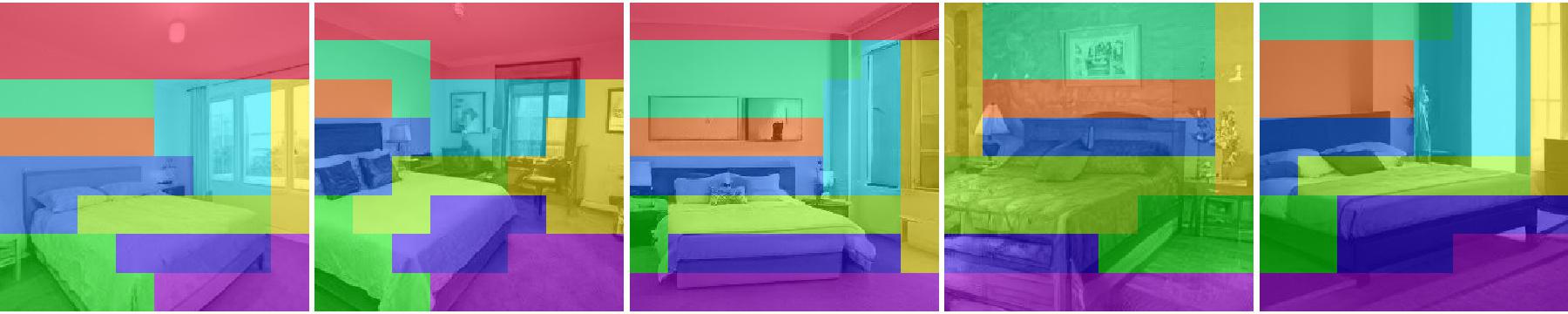}\\
		{\small \rotatebox[origin=c]{90}{Layer 4, $K=15$}}
		\includegraphics[align=c,width=0.89\textwidth]{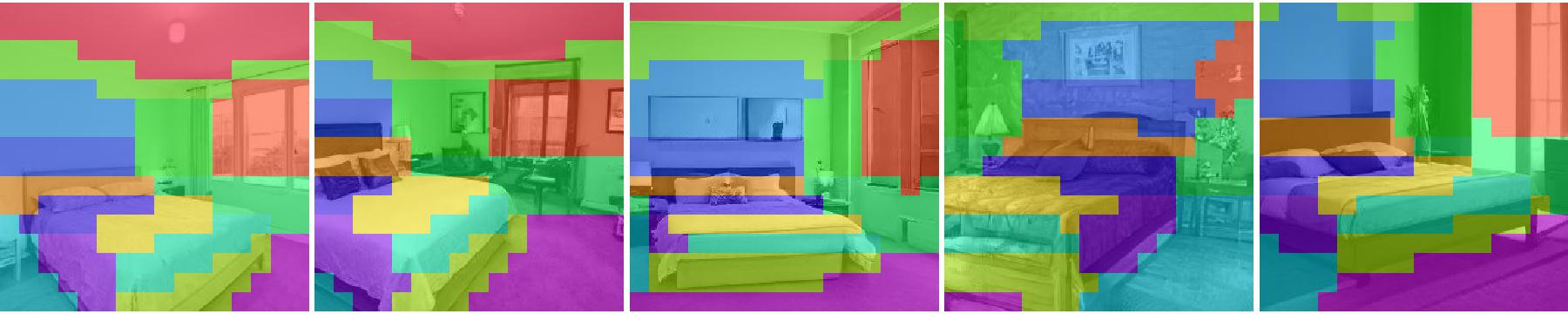}\\
		{\small \rotatebox[origin=c]{90}{Layer 5, $K=15$}}
		\includegraphics[align=c,width=0.89\textwidth]{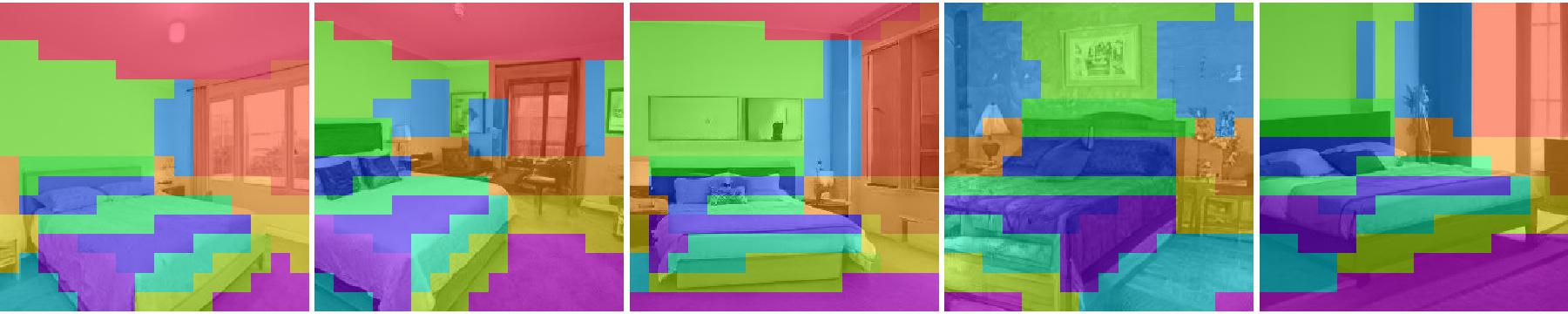}\\
		{\small \rotatebox[origin=c]{90}{Layer 6, $K=17$}}
		\includegraphics[align=c,width=0.89\textwidth]{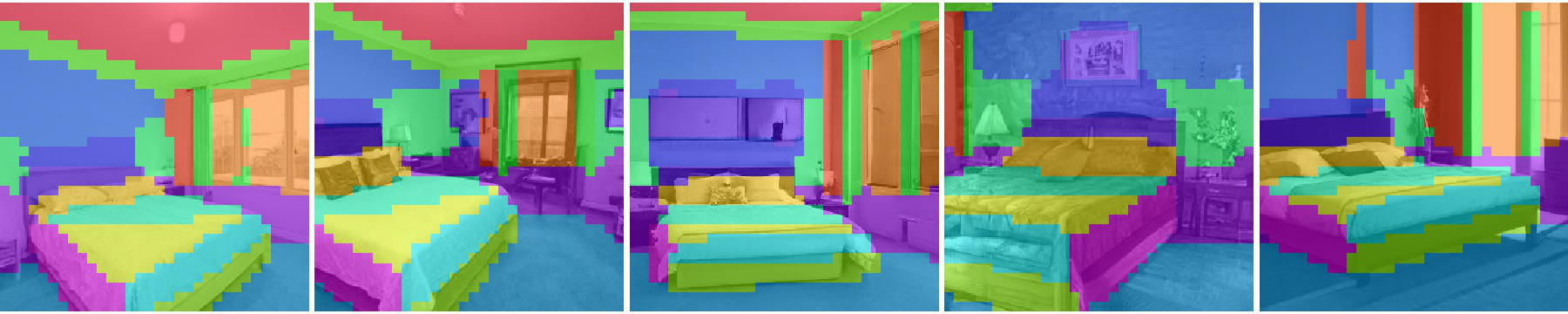}\\
		{\small \rotatebox[origin=c]{90}{Layer 8, $K=15$}}
		\includegraphics[align=c,width=0.89\textwidth]{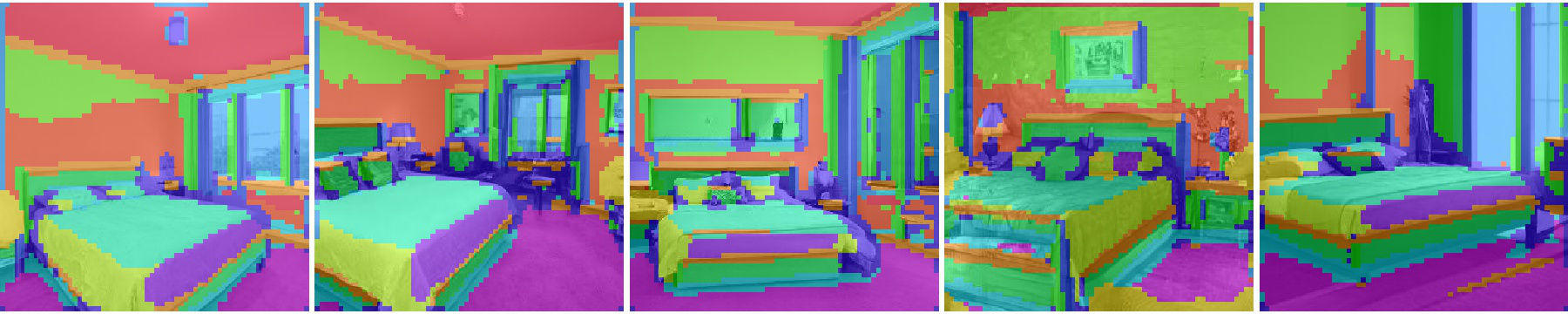}\\
		{\small \rotatebox[origin=c]{90}{Layer 9, $K=15$}}
		\includegraphics[align=c,width=0.89\textwidth]{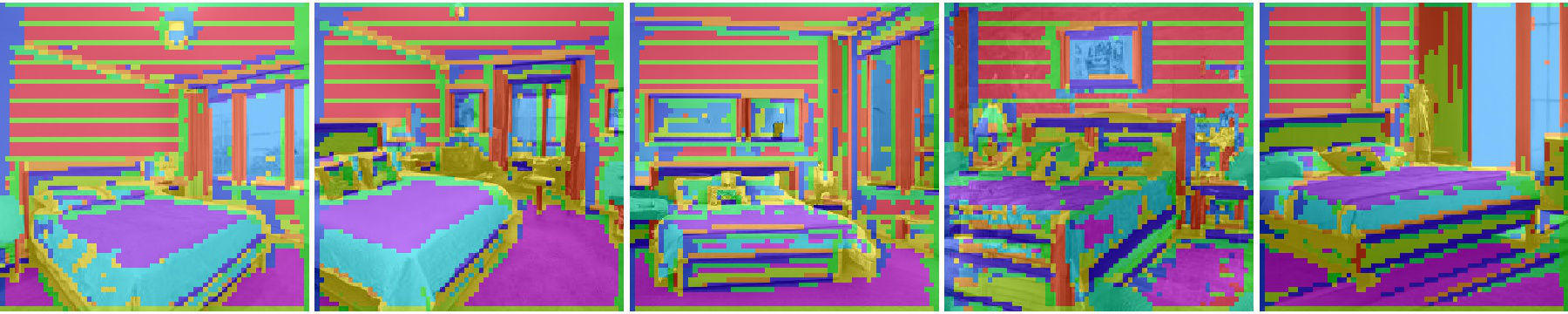}\\
		\caption{Spherical {\tt k}-means cluster membership maps for various LSUN-Bedroom-StyleGAN layer. Color-coding signifies different clusters, and is arbitrarily determined per layer.}
		\label{fig:kmeans-SG-Bedroom}
	\end{figure*}

\begin{figure*}
		\centering
		{\small \rotatebox[origin=c]{90}{RGB output}} \includegraphics[align=c,width=0.89\textwidth]{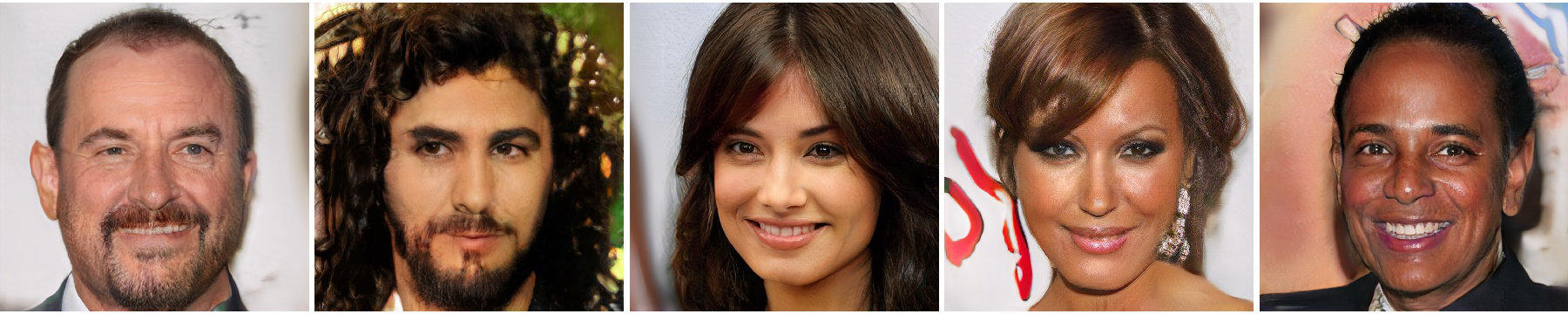}\\
		{\small \rotatebox[origin=c]{90}{Layer 3, $K=15$}} \includegraphics[align=c,width=0.89\textwidth]{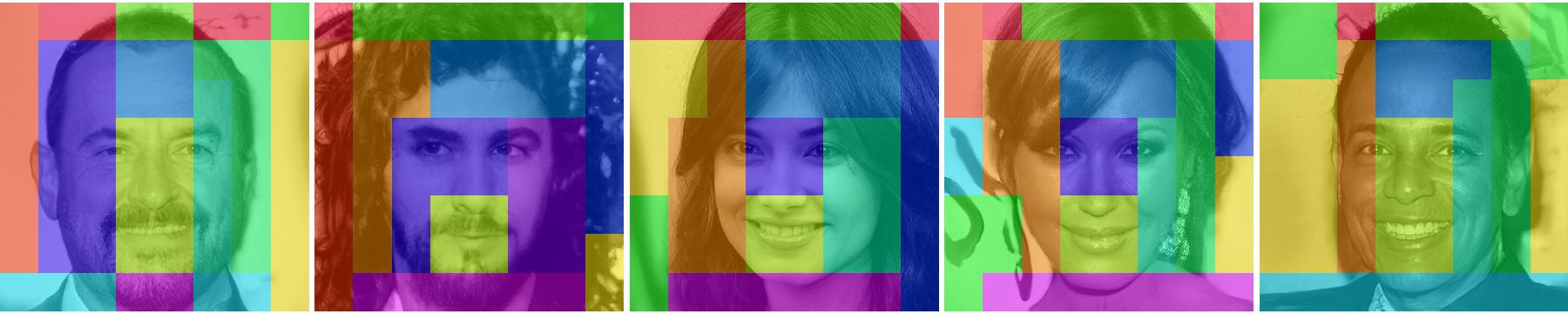}\\
		{\small \rotatebox[origin=c]{90}{Layer 4, $K=15$}}
		\includegraphics[align=c,width=0.89\textwidth]{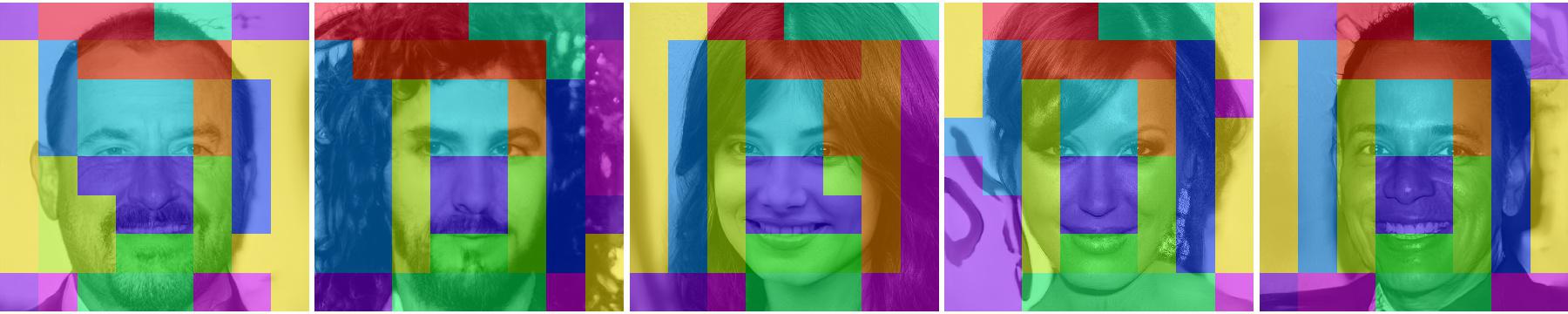}\\
		{\small \rotatebox[origin=c]{90}{Layer 5, $K=15$}}
		\includegraphics[align=c,width=0.89\textwidth]{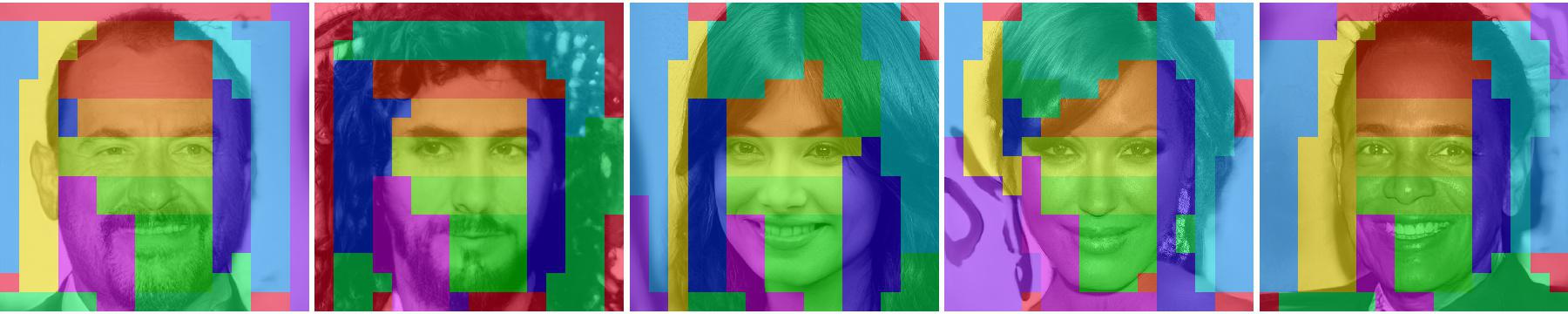}\\
		{\small \rotatebox[origin=c]{90}{Layer 6, $K=15$}}
		\includegraphics[align=c,width=0.89\textwidth]{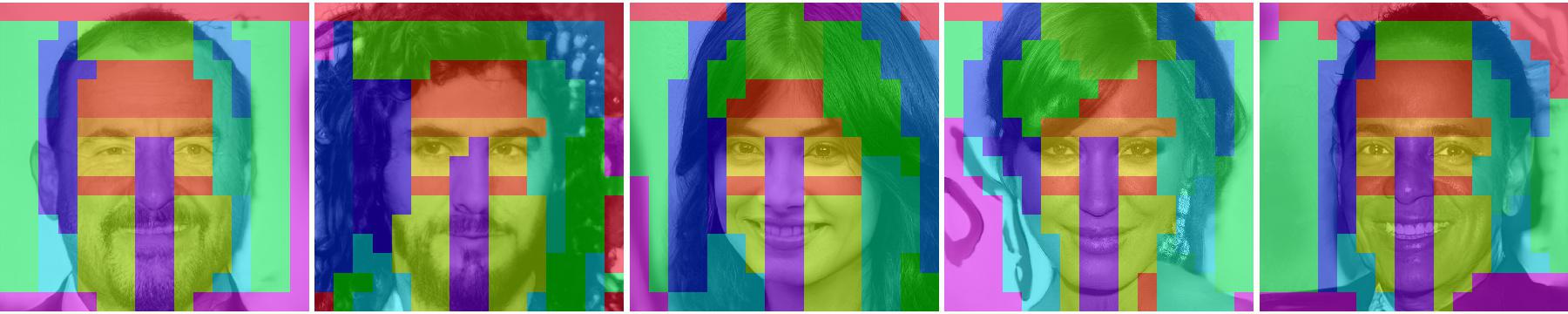}\\
		{\small \rotatebox[origin=c]{90}{Layer 8, $K=15$}}
		\includegraphics[align=c,width=0.89\textwidth]{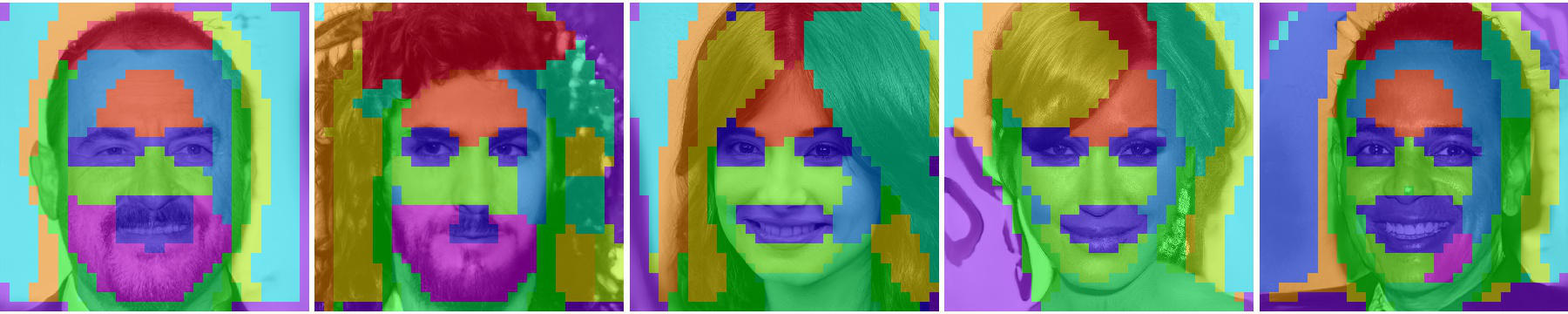}\\
		{\small \rotatebox[origin=c]{90}{Layer 11, $K=15$}}
		\includegraphics[align=c,width=0.89\textwidth]{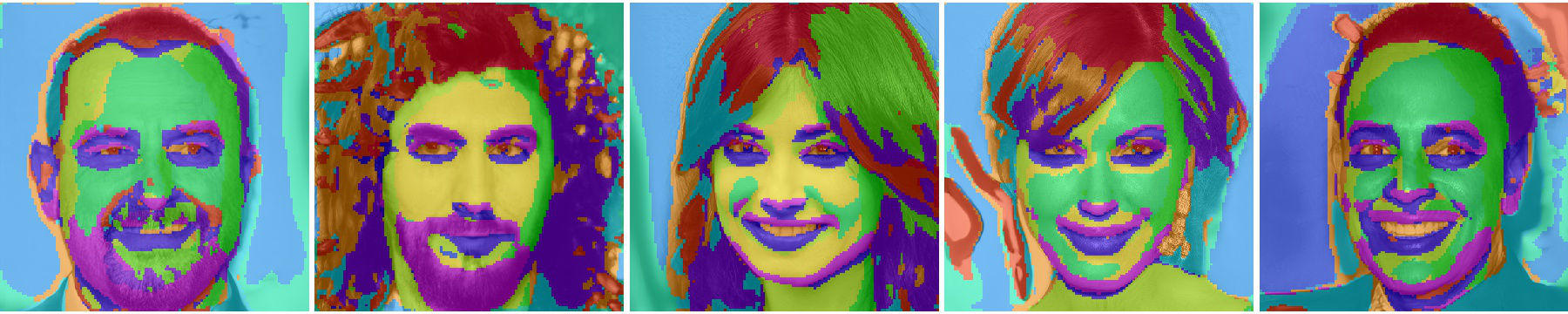}\\
		\caption{Spherical {\tt k}-means cluster membership maps for various CelebA-HQ-ProgGAN layer. Color-coding signifies different clusters, and is arbitrarily determined per layer.}
		\label{fig:kmeans-PG-Celeb}
	\end{figure*}

	
	In Figs. \ref{fig:kmeans-SG-FFHQ}, \ref{fig:kmeans-SG-Bedroom} we show examples produced with StyleGAN \cite{karras2019style}, where the tensor $\tU$ is up-sampled and overlaid on RGB images for ease of interpretation. The color-coding in these figures indicates to which cluster a spatial position belongs. In Fig. \ref{fig:kmeans-PG-Celeb} we similarly show results for ProgGAN \cite{karras2018progressive} on CelebA-HQ \cite{karras2018progressive}.
	
	The main observation emerging from this analysis is that at certain layers (e.g., the $32\times32$ layer 6 of StyleGAN), activations capture abstract semantic concepts (e.g., \emph{eyes} for faces, \emph{pillow} for bedrooms).
	
	By manually examining the cluster membership masks of a few (five to ten) samples, an annotator can easily label a cluster as representing a certain object.
	Thus, we randomly generated $N=200$ samples and recorded all their activations. We tested several layers and rank $K$ combinations and selected the one that qualitatively yielded the most semantic decomposition into objects, as shown in Figures \ref{fig:kmeans-SG-FFHQ} and \ref{fig:kmeans-SG-Bedroom}. We then manually labeled the resulting clusters. In the case that multiple clusters matched a part of interest, we merged their masks into a single mask. Note that this process is a one-time, offline process (per dataset/GAN) that then drives a fully automated semantic editing operation.
	
	\begin{figure*}
		\centering
		{\small \rotatebox[origin=c]{90}{RGB output}} \includegraphics[align=c,width=0.89\textwidth]{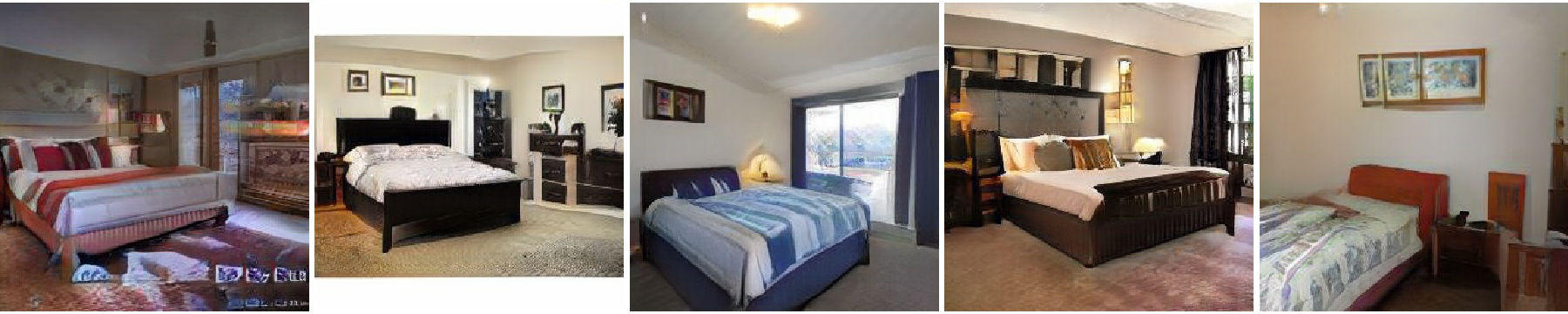}\\
		{\small \rotatebox[origin=c]{90}{Layer 3, $K=15$}} \includegraphics[align=c,width=0.89\textwidth]{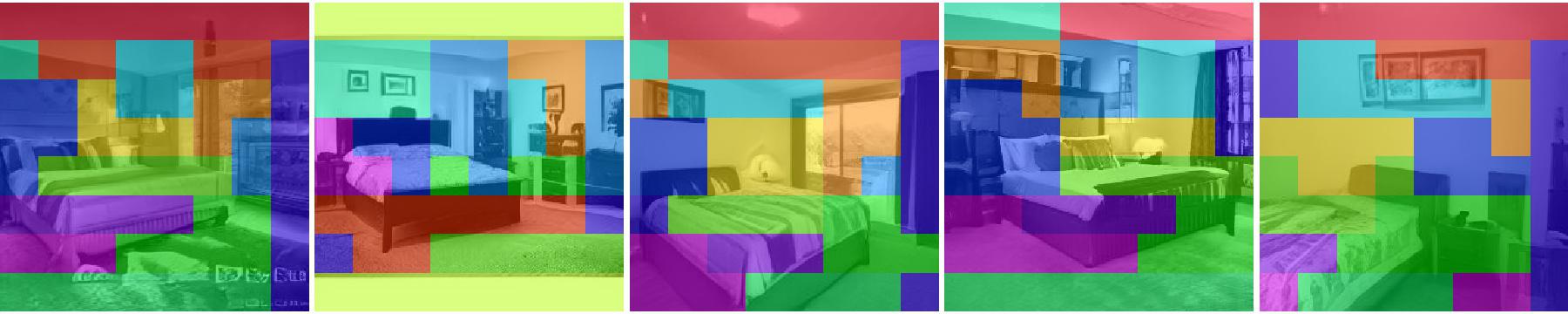}\\
		{\small \rotatebox[origin=c]{90}{Layer 4, $K=15$}}
		\includegraphics[align=c,width=0.89\textwidth]{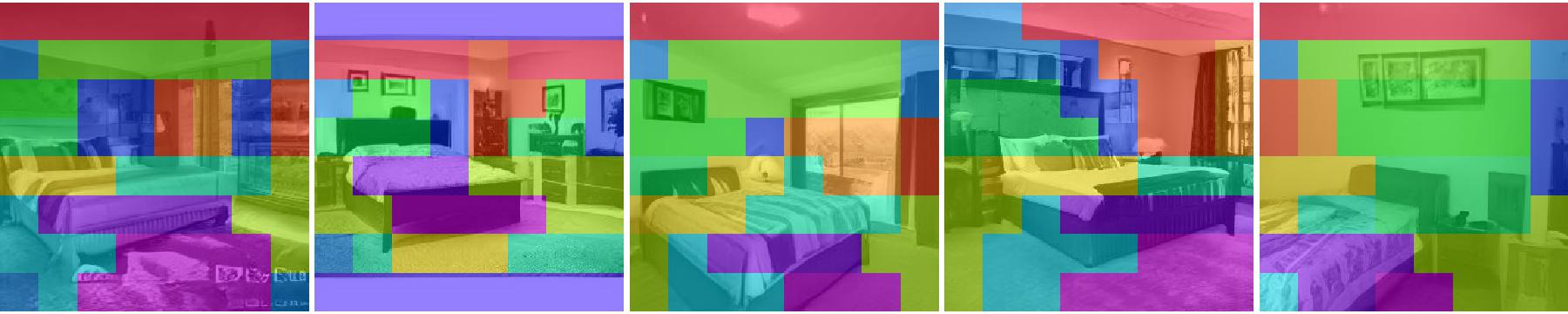}\\
		{\small \rotatebox[origin=c]{90}{Layer 5, $K=15$}}
		\includegraphics[align=c,width=0.89\textwidth]{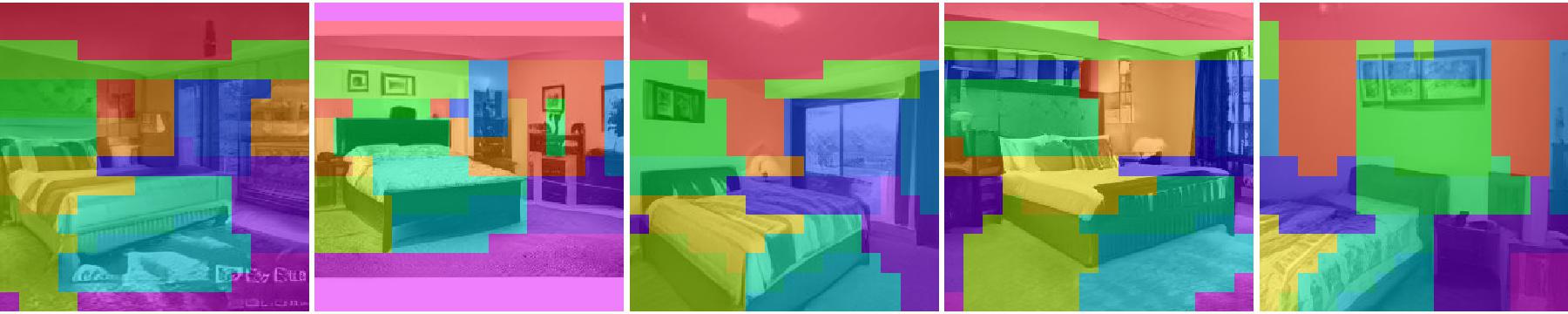}\\
		{\small \rotatebox[origin=c]{90}{Layer 6, $K=15$}}
		\includegraphics[align=c,width=0.89\textwidth]{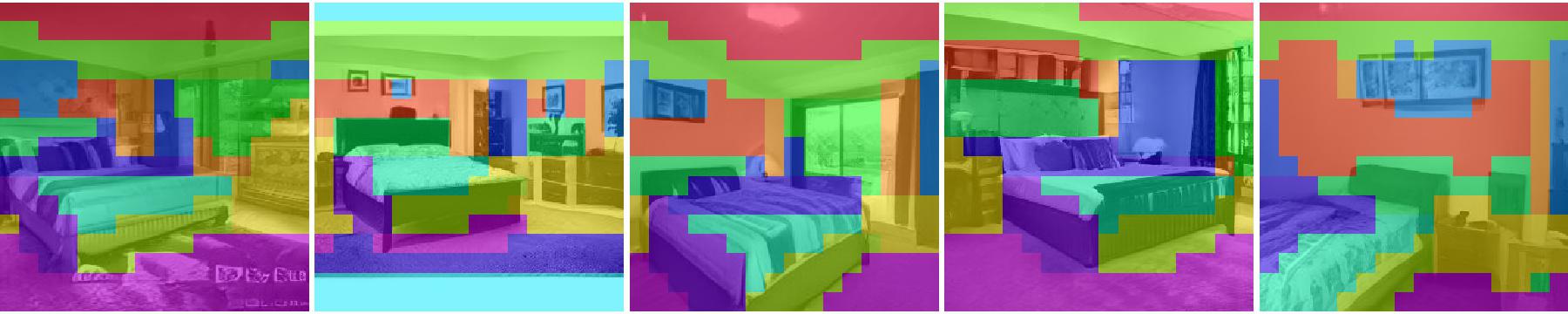}\\
		{\small \rotatebox[origin=c]{90}{Layer 8, $K=15$}}
		\includegraphics[align=c,width=0.89\textwidth]{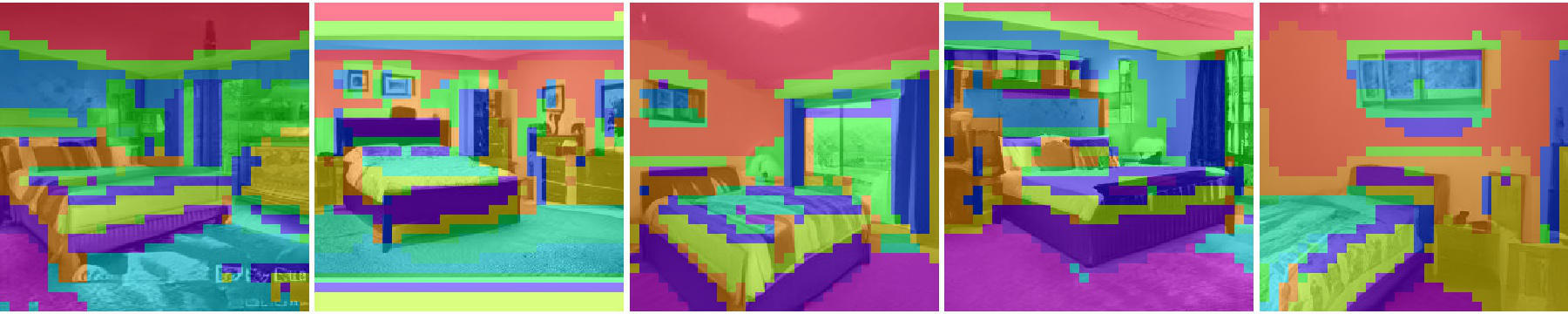}\\
		{\small \rotatebox[origin=c]{90}{Layer 9, $K=15$}}
		\includegraphics[align=c,width=0.89\textwidth]{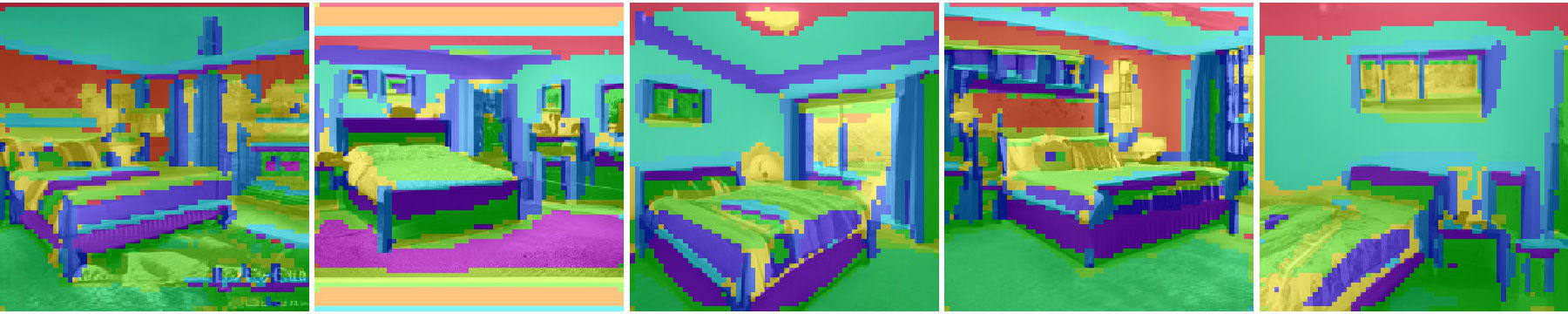}\\
		\caption{Spherical {\tt k}-means cluster membership maps for various LSUN-Bedroom-ProgGAN layer. Color-coding signifies different clusters, and is arbitrarily determined per layer.}
		\label{fig:kmeans-PG-Bedroom}
	\end{figure*}
	
	\begin{figure*}[!b]
		{\centering \setlength{\tabcolsep}{0pt} 
			\figurefont
			\renewcommand{\arraystretch}{0.1}
			\newcommand\figpathorig{figures/PartGrid/FFHQ_2001/}
			\newcommand\figpath{figures/PartGrid/FFHQ_2001_diff/}
			\newcommand\partone{eyes}
			\newcommand\parttwo{nose}
			\newcommand\partthree{mouth}
			\begin{tabular}{M{0.2\textwidth}M{0.18\textwidth}M{0.18\textwidth}M{0.18\textwidth}M{0.18\textwidth}}
				\textbf{Target} & \textbf{Reference 1} & \textbf{Reference 2} & \textbf{Reference 3}  & \textbf{Reference 4} \\
				\includegraphics[width=0.18\textwidth]{\figpathorig original.jpg}  &
				\includegraphics[width=0.18\textwidth]{\figpathorig style1.jpg} &
				\includegraphics[width=0.18\textwidth]{\figpathorig style2.jpg} &
				\includegraphics[width=0.18\textwidth]{\figpathorig style3.jpg} &
				\includegraphics[width=0.18\textwidth]{\figpathorig style4.jpg}\\
				&&&& \\
				\textbf{\expandafter\MakeUppercase \partone:} &\includegraphics[width=0.18\textwidth]{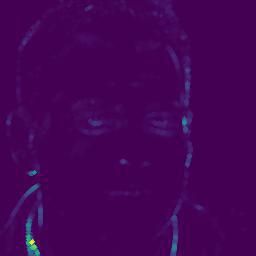} &
				\includegraphics[width=0.18\textwidth]{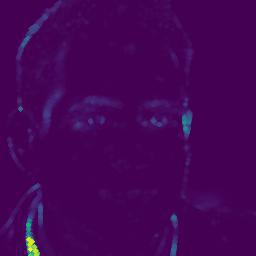} &
				\includegraphics[width=0.18\textwidth]{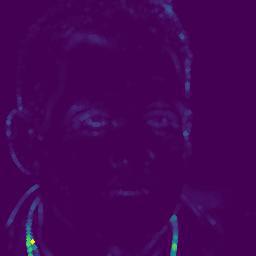} &
				\includegraphics[width=0.18\textwidth]{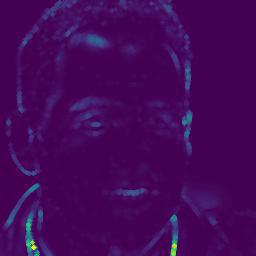}\\
				
				\textbf{\expandafter\MakeUppercase \parttwo:} &\includegraphics[width=0.18\textwidth]{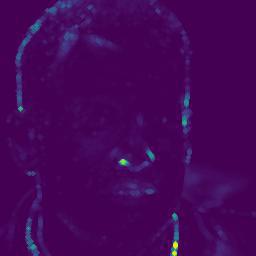} &
				\includegraphics[width=0.18\textwidth]{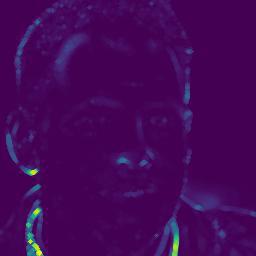} &
				\includegraphics[width=0.18\textwidth]{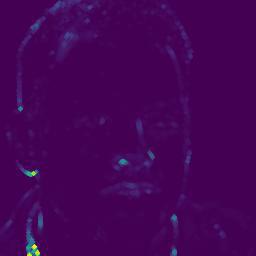} &
				\includegraphics[width=0.18\textwidth]{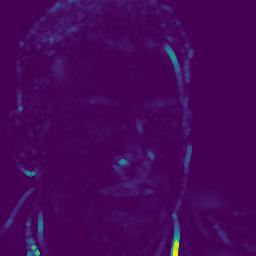}\\
				
				\textbf{\expandafter\MakeUppercase \partthree:} &\includegraphics[width=0.18\textwidth]{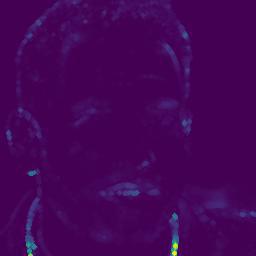} &
				\includegraphics[width=0.18\textwidth]{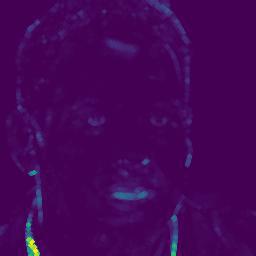} &
				\includegraphics[width=0.18\textwidth]{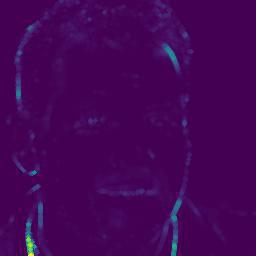} &
				\includegraphics[width=0.18\textwidth]{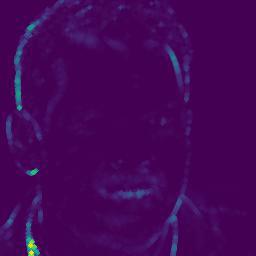}
		\end{tabular}}
		\caption{Mean-squared error maps between the edited outputs shown in Fig. \ref{fig:main-ffhq} and the target image, shown in the same figure. Editing is primarily focused on the object of interest, though some subtle changes do occur elsewhere in the scene.}
		\label{fig:main-ffhq-diff}
	\end{figure*}
	
	\section{Squared-error maps}
	Squared-error ``diff'' maps between edited outputs and the target image help detect changes between the two images and evaluate the locality of the edit operation. We compute the error in CIELAB color-space.
	
\begin{figure*}
		
		{\centering \setlength{\tabcolsep}{0pt} 
			\figurefont
			\renewcommand{\arraystretch}{0.1}
			\newcommand\figpathorig{figures/PartGrid/bedrooms_6813/}
			\newcommand\figpath{figures/PartGrid/bedrooms_6813_diff/}
			\newcommand\partone{bed}
			\newcommand\parttwo{pillow}
			\newcommand\partthree{window}
			\begin{tabular}{M{0.2\textwidth}M{0.18\textwidth}M{0.18\textwidth}M{0.18\textwidth}M{0.18\textwidth}}
				\textbf{Target} & \textbf{Reference 1} & \textbf{Reference 2} & \textbf{Reference 3}  & \textbf{Reference 4} \\
				\includegraphics[width=0.18\textwidth]{\figpathorig original.jpg}  &
				\includegraphics[width=0.18\textwidth]{\figpathorig style1.jpg} &
				\includegraphics[width=0.18\textwidth]{\figpathorig style2.jpg} &
				\includegraphics[width=0.18\textwidth]{\figpathorig style3.jpg} &
				\includegraphics[width=0.18\textwidth]{\figpathorig style4.jpg}\\
				&&&& \\
				\textbf{\expandafter\MakeUppercase \partone:} &\includegraphics[width=0.18\textwidth]{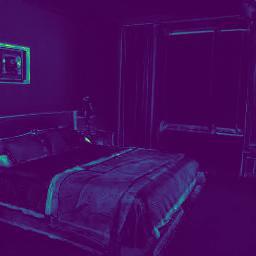} &
				\includegraphics[width=0.18\textwidth]{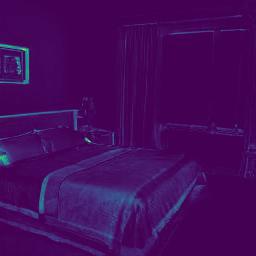} &
				\includegraphics[width=0.18\textwidth]{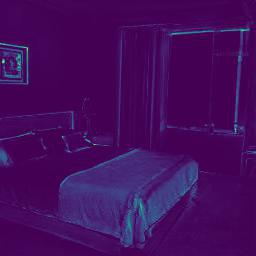} &
				\includegraphics[width=0.18\textwidth]{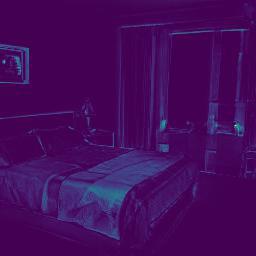}\\
				
				\textbf{\expandafter\MakeUppercase \parttwo:} &\includegraphics[width=0.18\textwidth]{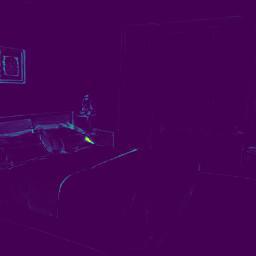} &
				\includegraphics[width=0.18\textwidth]{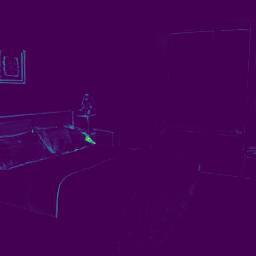} &
				\includegraphics[width=0.18\textwidth]{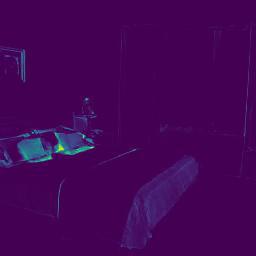} &
				\includegraphics[width=0.18\textwidth]{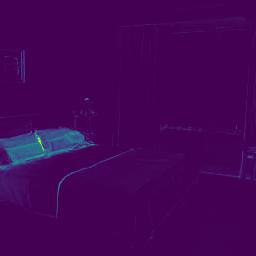}\\
				
				\textbf{\expandafter\MakeUppercase \partthree:} &\includegraphics[width=0.18\textwidth]{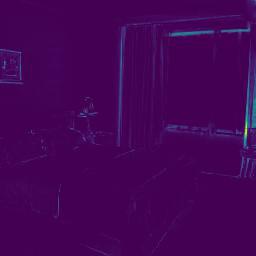} &
				\includegraphics[width=0.18\textwidth]{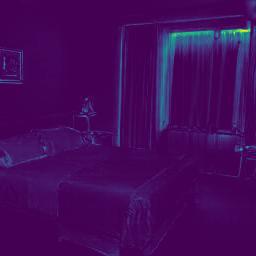} &
				\includegraphics[width=0.18\textwidth]{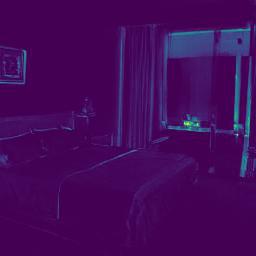} &
				\includegraphics[width=0.18\textwidth]{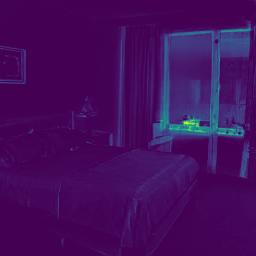}
		\end{tabular}}

		\caption{Mean-squared error maps corresponding to  Fig. \ref{fig:main-bedrooms}. Correlations learned and respected by the GAN sometimes lead to unintentional changes, e.g., changes to the picture on the wall when editing the bed (first row).}
		\label{fig:main-bedrooms-diff}
	\end{figure*}

	In Figs. \ref{fig:main-ffhq-diff} and \ref{fig:main-bedrooms-diff} we show the diff maps corresponding to Figs. 3 and 4 respectively.
	
	\section{Additional qualitative results with StyleGAN2}
\begin{figure*}
		
		{\centering \setlength{\tabcolsep}{0pt} 
			\figurefont
			\renewcommand{\arraystretch}{0.1}
			\newcommand\figpath{figures/PartGrid/stylegan2_cat_2002/}
			\newcommand\partone{eyes}
			\newcommand\parttwo{ears}
			\newcommand\partthree{snout}
			\newcommand\partfour{body}
			\begin{tabular}{M{0.2\textwidth}M{0.18\textwidth}M{0.18\textwidth}M{0.18\textwidth}M{0.18\textwidth}}
				\textbf{Target} & \textbf{Reference 1} & \textbf{Reference 2} & \textbf{Reference 3}  & \textbf{Reference 4} \\
				\includegraphics[width=0.18\textwidth]{\figpath original.jpg}  &
				\includegraphics[width=0.18\textwidth]{\figpath style1.jpg} &
				\includegraphics[width=0.18\textwidth]{\figpath style2.jpg} &
				\includegraphics[width=0.18\textwidth]{\figpath style3.jpg} &
				\includegraphics[width=0.18\textwidth]{\figpath style4.jpg}\\
				&&&& \\
				\textbf{\expandafter\MakeUppercase \partone:} &\includegraphics[width=0.18\textwidth]{\figpath interp1_\partone.jpg} &
				\includegraphics[width=0.18\textwidth]{\figpath interp2_\partone.jpg} &
				\includegraphics[width=0.18\textwidth]{\figpath interp3_\partone.jpg} &
				\includegraphics[width=0.18\textwidth]{\figpath interp4_\partone.jpg}\\
				
				\textbf{\expandafter\MakeUppercase \parttwo:} &\includegraphics[width=0.18\textwidth]{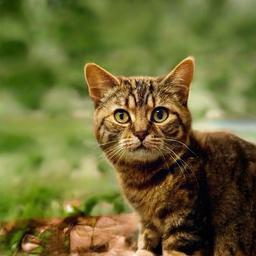} &
				\includegraphics[width=0.18\textwidth]{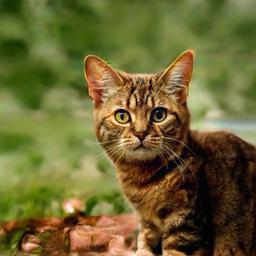} &
				\includegraphics[width=0.18\textwidth]{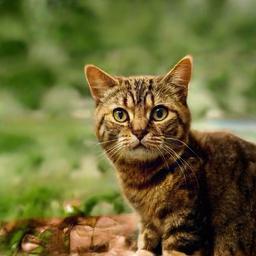} &
				\includegraphics[width=0.18\textwidth]{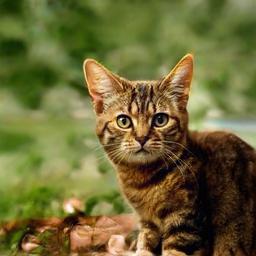}\\
				
				\textbf{\expandafter\MakeUppercase \partthree:} &\includegraphics[width=0.18\textwidth]{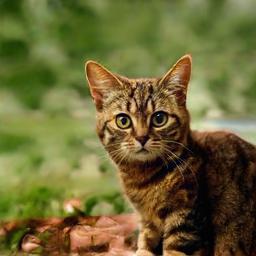} &
				\includegraphics[width=0.18\textwidth]{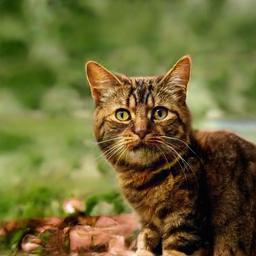} &
				\includegraphics[width=0.18\textwidth]{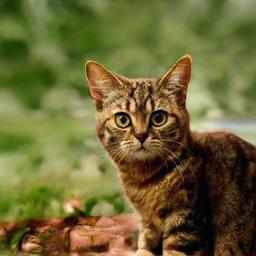} &
				\includegraphics[width=0.18\textwidth]{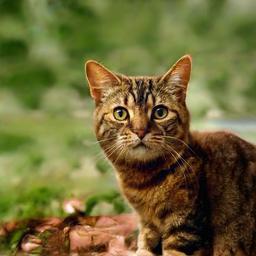}\\
				
				\textbf{\expandafter\MakeUppercase \partfour:} &\includegraphics[width=0.18\textwidth]{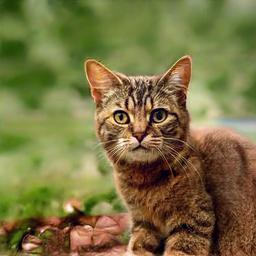} &
				\includegraphics[width=0.18\textwidth]{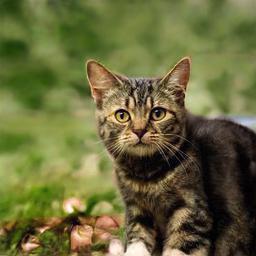} &
				\includegraphics[width=0.18\textwidth]{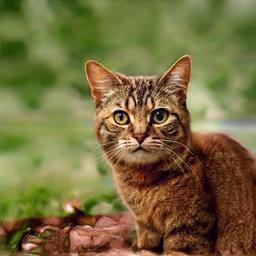} &
				\includegraphics[width=0.18\textwidth]{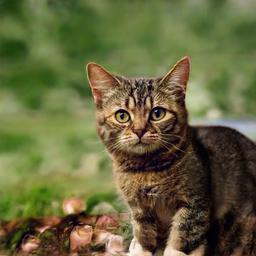}
		\end{tabular}}
		
		\caption{Our local editing method applied to StyleGAN2 trained on LSUN-Cats.}
		\label{fig:main-cats}
	\end{figure*}

\begin{figure*}
		{\centering \setlength{\tabcolsep}{0pt} 
			\figurefont
			\renewcommand{\arraystretch}{0.1}
			\newcommand\figpathorig{figures/PartGrid/stylegan2_cat_2002/}
			\newcommand\figpath{figures/PartGrid/stylegan2_cat_2002_diff/}
			\newcommand\partone{eyes}
			\newcommand\parttwo{ears}
			\newcommand\partthree{snout}
			\newcommand\partfour{body}
			\begin{tabular}{M{0.2\textwidth}M{0.18\textwidth}M{0.18\textwidth}M{0.18\textwidth}M{0.18\textwidth}}
				\textbf{Target} & \textbf{Reference 1} & \textbf{Reference 2} & \textbf{Reference 3}  & \textbf{Reference 4} \\
				\includegraphics[width=0.18\textwidth]{\figpathorig original.jpg}  &
				\includegraphics[width=0.18\textwidth]{\figpathorig style1.jpg} &
				\includegraphics[width=0.18\textwidth]{\figpathorig style2.jpg} &
				\includegraphics[width=0.18\textwidth]{\figpathorig style3.jpg} &
				\includegraphics[width=0.18\textwidth]{\figpathorig style4.jpg}\\
				&&&& \\
				\textbf{\expandafter\MakeUppercase \partone:} &\includegraphics[width=0.18\textwidth]{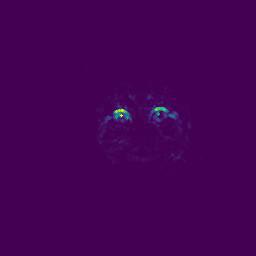} &
				\includegraphics[width=0.18\textwidth]{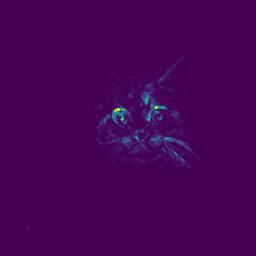} &
				\includegraphics[width=0.18\textwidth]{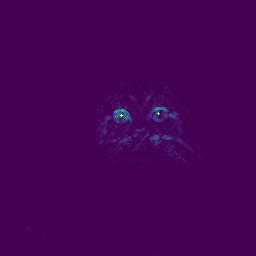} &
				\includegraphics[width=0.18\textwidth]{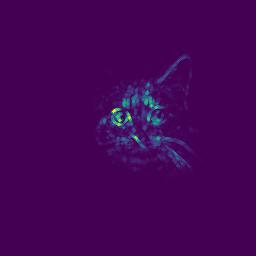}\\
				
				\textbf{\expandafter\MakeUppercase \parttwo:} &\includegraphics[width=0.18\textwidth]{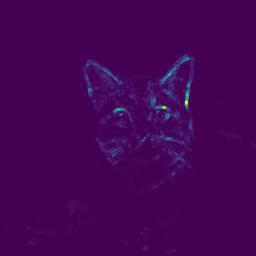} &
				\includegraphics[width=0.18\textwidth]{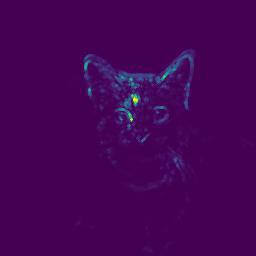} &
				\includegraphics[width=0.18\textwidth]{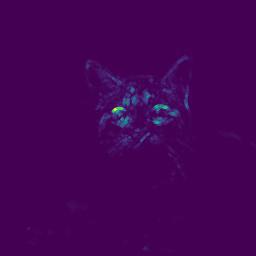} &
				\includegraphics[width=0.18\textwidth]{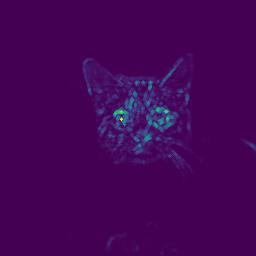}\\
				
				\textbf{\expandafter\MakeUppercase \partthree:} &\includegraphics[width=0.18\textwidth]{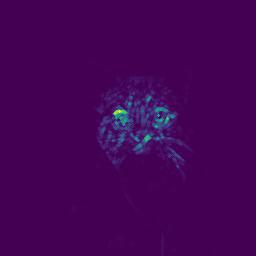} &
				\includegraphics[width=0.18\textwidth]{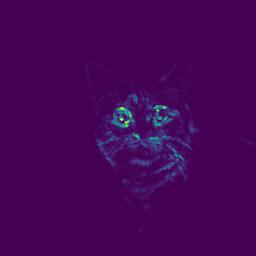} &
				\includegraphics[width=0.18\textwidth]{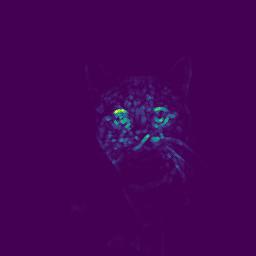} &
				\includegraphics[width=0.18\textwidth]{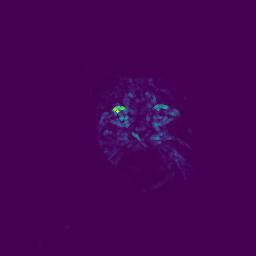}\\
				
				\textbf{\expandafter\MakeUppercase \partfour:} &\includegraphics[width=0.18\textwidth]{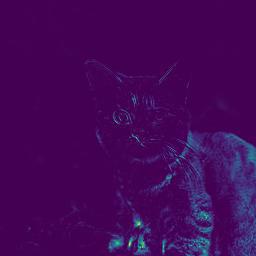} &
				\includegraphics[width=0.18\textwidth]{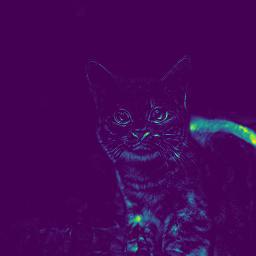} &
				\includegraphics[width=0.18\textwidth]{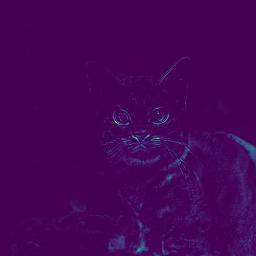} &
				\includegraphics[width=0.18\textwidth]{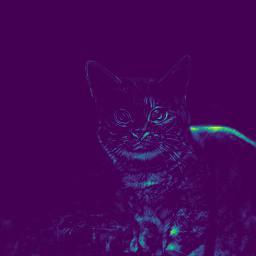}
		\end{tabular}}
		
		\caption{Diff maps corresponding to the results in Fig. \ref{fig:main-cats}, for StyleGAN2 trained on LSUN-Cats.}
		\label{fig:main-cats-diff}
	\end{figure*}

\begin{figure*}
		
		{\centering \setlength{\tabcolsep}{0pt} 
			\figurefont
			\renewcommand{\arraystretch}{0.1}
			\newcommand\figpath{figures/PartGrid/stylegan2_car_2005a/}
			\newcommand\partone{wheels}
			\newcommand\parttwo{bumper}
			\newcommand\partthree{hood}
			\newcommand\partfour{cabin}
			\begin{tabular}{M{0.2\textwidth}M{0.18\textwidth}M{0.18\textwidth}M{0.18\textwidth}M{0.18\textwidth}}
				\textbf{Target} & \textbf{Reference 1} & \textbf{Reference 2} & \textbf{Reference 3}  & \textbf{Reference 4} \\
				\includegraphics[width=0.18\textwidth]{\figpath original.jpg}  &
				\includegraphics[width=0.18\textwidth]{\figpath style1.jpg} &
				\includegraphics[width=0.18\textwidth]{\figpath style2.jpg} &
				\includegraphics[width=0.18\textwidth]{\figpath style3.jpg} &
				\includegraphics[width=0.18\textwidth]{\figpath style4.jpg}\\
				&&&& \\
				\textbf{\expandafter\MakeUppercase \partone:} &\includegraphics[width=0.18\textwidth]{\figpath interp1_\partone.jpg} &
				\includegraphics[width=0.18\textwidth]{\figpath interp2_\partone.jpg} &
				\includegraphics[width=0.18\textwidth]{\figpath interp3_\partone.jpg} &
				\includegraphics[width=0.18\textwidth]{\figpath interp4_\partone.jpg}\\
				
				\textbf{\expandafter\MakeUppercase \parttwo:} &\includegraphics[width=0.18\textwidth]{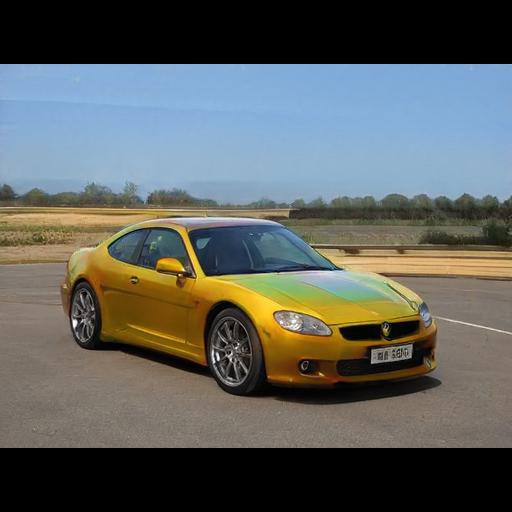} &
				\includegraphics[width=0.18\textwidth]{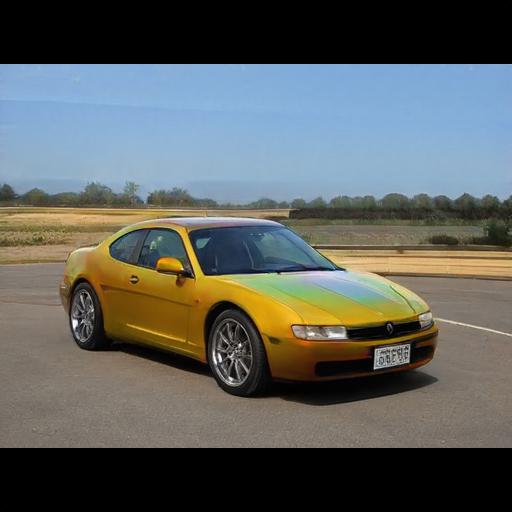} &
				\includegraphics[width=0.18\textwidth]{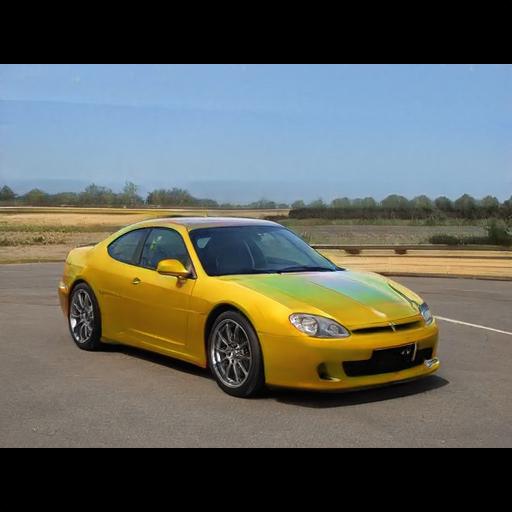} &
				\includegraphics[width=0.18\textwidth]{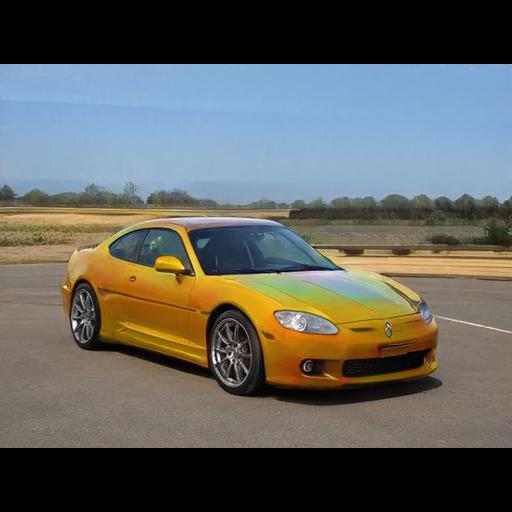}\\
				
				\textbf{\expandafter\MakeUppercase \partthree:} &\includegraphics[width=0.18\textwidth]{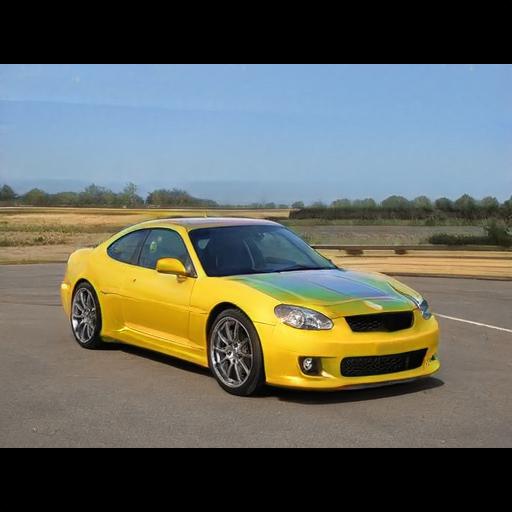} &
				\includegraphics[width=0.18\textwidth]{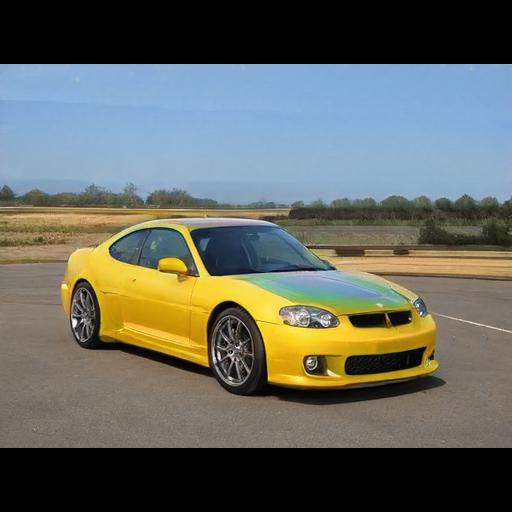} &
				\includegraphics[width=0.18\textwidth]{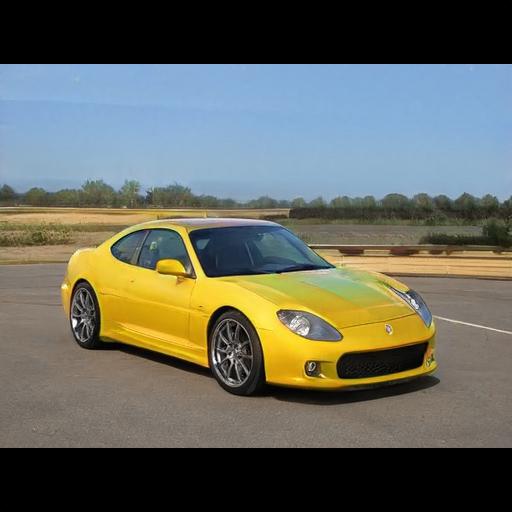} &
				\includegraphics[width=0.18\textwidth]{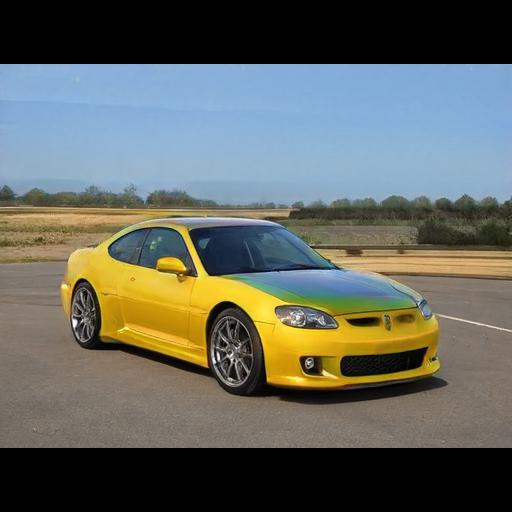}\\
				
				\textbf{\expandafter\MakeUppercase \partfour:} &\includegraphics[width=0.18\textwidth]{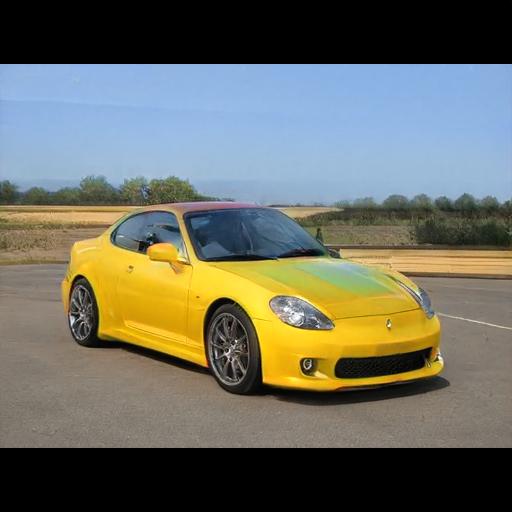} &
				\includegraphics[width=0.18\textwidth]{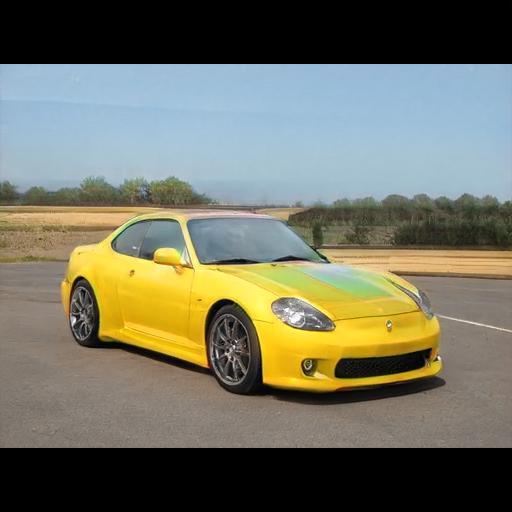} &
				\includegraphics[width=0.18\textwidth]{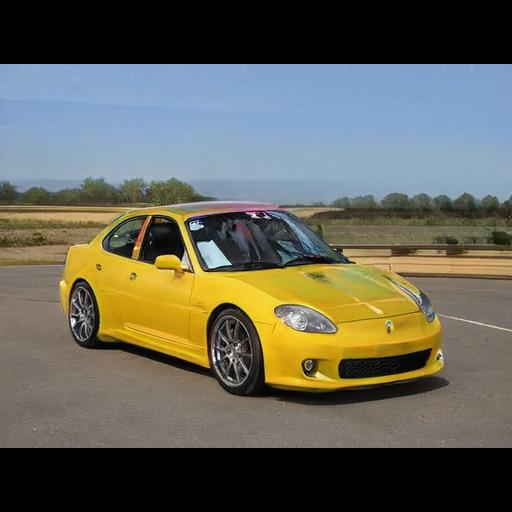} &
				\includegraphics[width=0.18\textwidth]{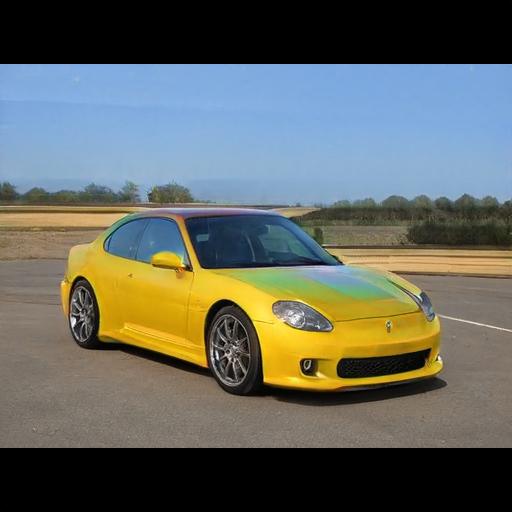}
		\end{tabular}}
		
		\caption{Our local editing method applied to StyleGAN2 trained on LSUN-Cars.}
		\label{fig:main-cars}
	\end{figure*}

\begin{figure*}
		{\centering \setlength{\tabcolsep}{0pt} 
			\figurefont
			\renewcommand{\arraystretch}{0.1}
			\newcommand\figpathorig{figures/PartGrid/stylegan2_car_2005a/}
			\newcommand\figpath{figures/PartGrid/stylegan2_car_2005a_diff/}
			\newcommand\partone{wheels}
			\newcommand\parttwo{bumper}
			\newcommand\partthree{hood}
			\newcommand\partfour{cabin}
			\begin{tabular}{M{0.2\textwidth}M{0.18\textwidth}M{0.18\textwidth}M{0.18\textwidth}M{0.18\textwidth}}
				\textbf{Target} & \textbf{Reference 1} & \textbf{Reference 2} & \textbf{Reference 3}  & \textbf{Reference 4} \\
				\includegraphics[width=0.18\textwidth]{\figpathorig original.jpg}  &
				\includegraphics[width=0.18\textwidth]{\figpathorig style1.jpg} &
				\includegraphics[width=0.18\textwidth]{\figpathorig style2.jpg} &
				\includegraphics[width=0.18\textwidth]{\figpathorig style3.jpg} &
				\includegraphics[width=0.18\textwidth]{\figpathorig style4.jpg}\\
				&&&& \\
				\textbf{\expandafter\MakeUppercase \partone:} &\includegraphics[width=0.18\textwidth]{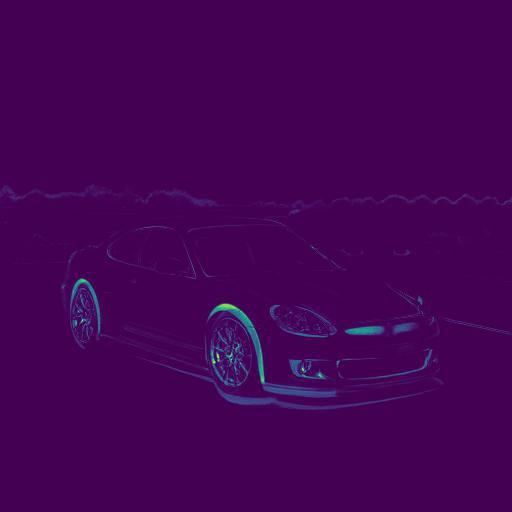} &
				\includegraphics[width=0.18\textwidth]{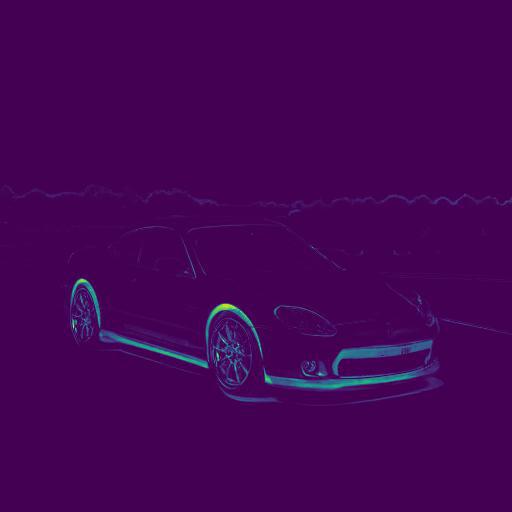} &
				\includegraphics[width=0.18\textwidth]{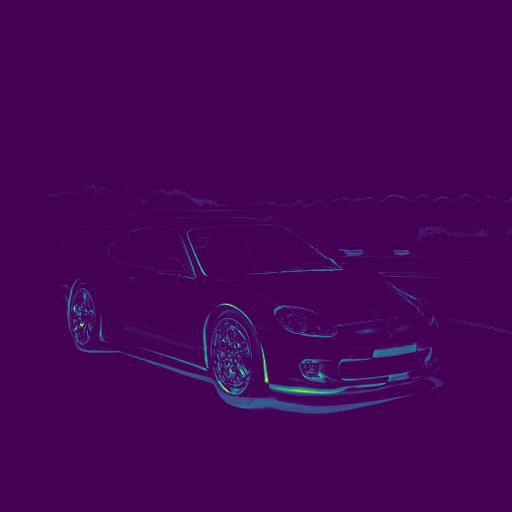} &
				\includegraphics[width=0.18\textwidth]{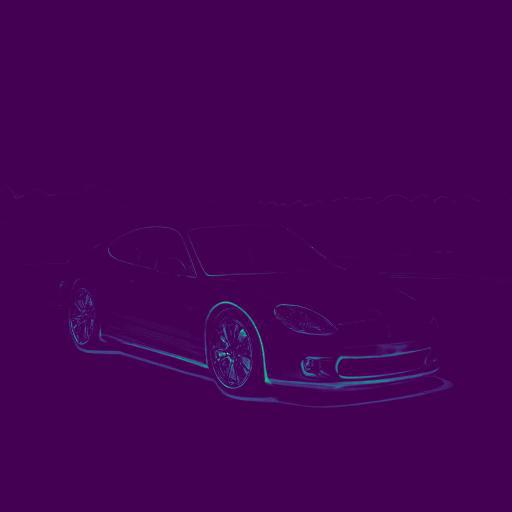}\\
				
				\textbf{\expandafter\MakeUppercase \parttwo:} &\includegraphics[width=0.18\textwidth]{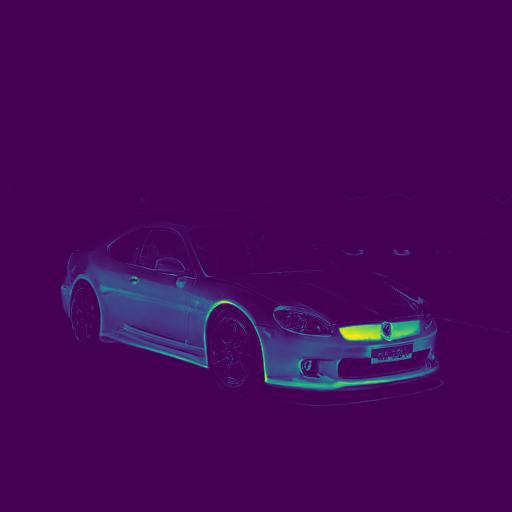} &
				\includegraphics[width=0.18\textwidth]{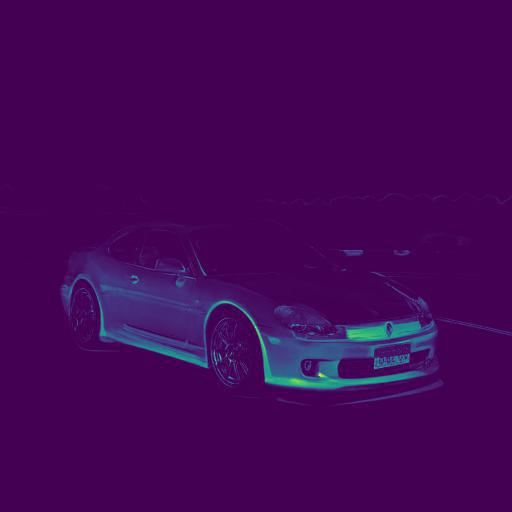} &
				\includegraphics[width=0.18\textwidth]{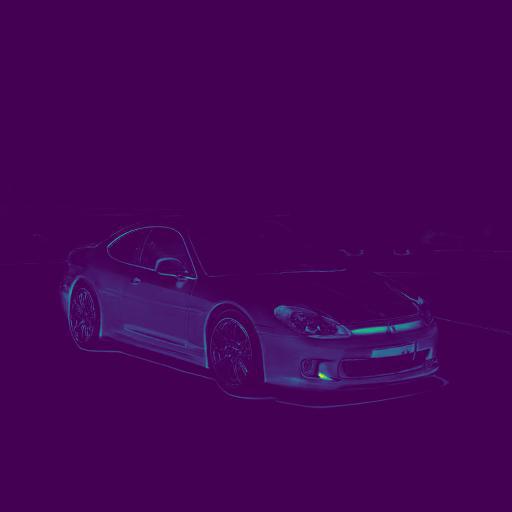} &
				\includegraphics[width=0.18\textwidth]{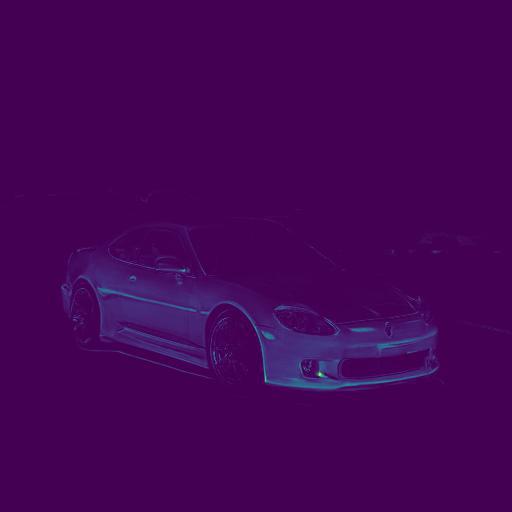}\\
				
				\textbf{\expandafter\MakeUppercase \partthree:} &\includegraphics[width=0.18\textwidth]{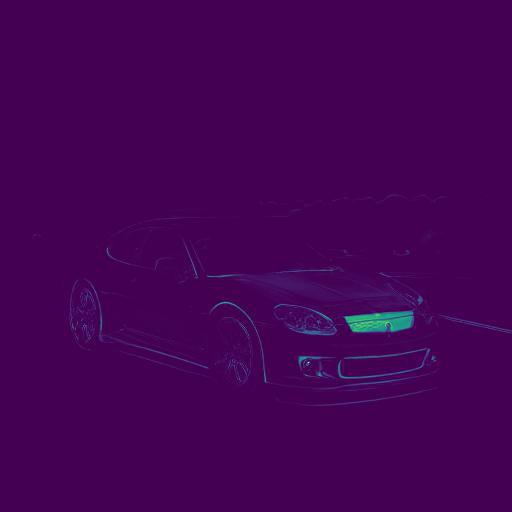} &
				\includegraphics[width=0.18\textwidth]{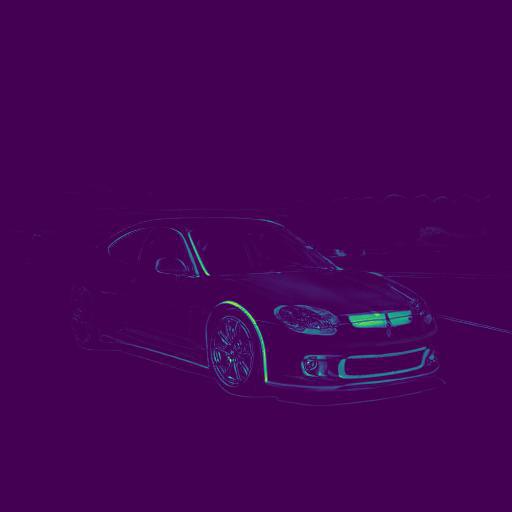} &
				\includegraphics[width=0.18\textwidth]{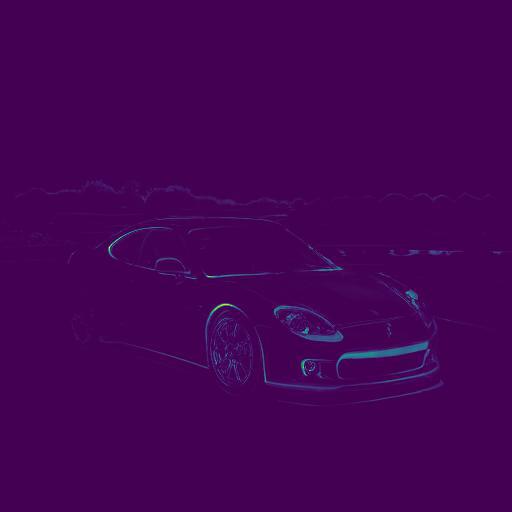} &
				\includegraphics[width=0.18\textwidth]{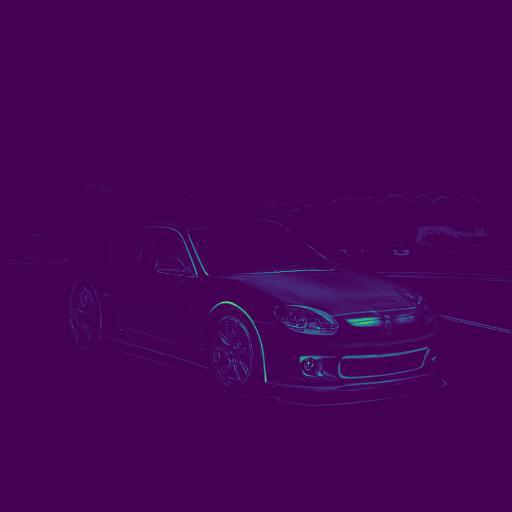}\\
				
				\textbf{\expandafter\MakeUppercase \partfour:} &\includegraphics[width=0.18\textwidth]{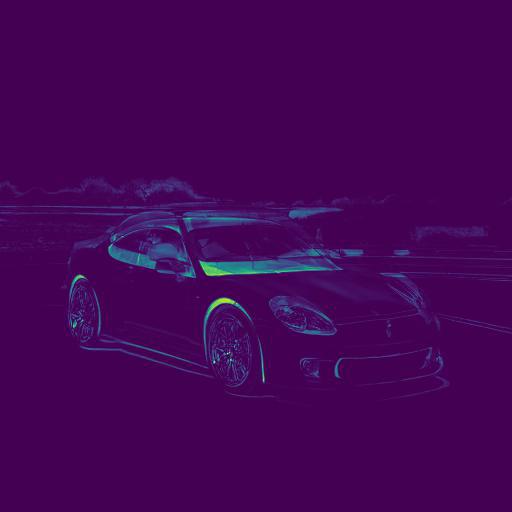} &
				\includegraphics[width=0.18\textwidth]{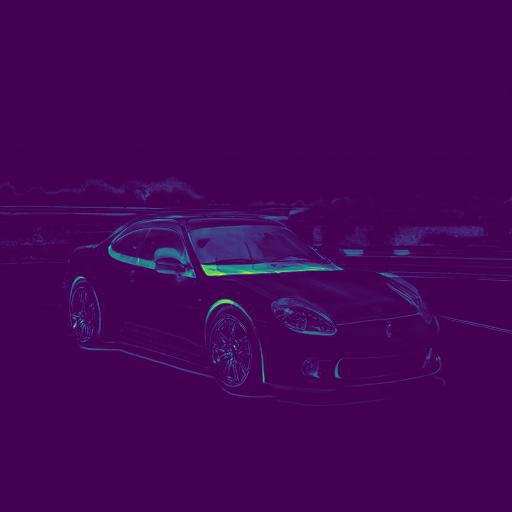} &
				\includegraphics[width=0.18\textwidth]{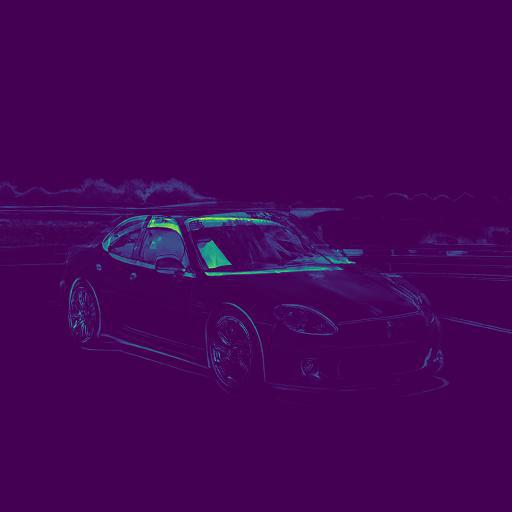} &
				\includegraphics[width=0.18\textwidth]{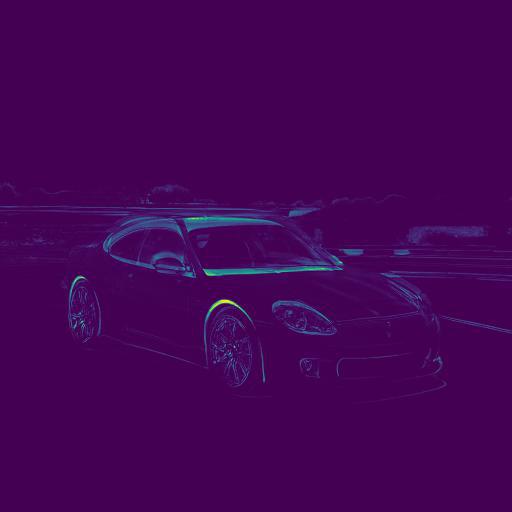}
		\end{tabular}}
		
		\caption{Diff maps corresponding to the results in Fig. \ref{fig:main-cars}, for StyleGAN2 trained on LSUN-Cars.}
		\label{fig:main-cars-diff}
	\end{figure*}

\begin{figure*}
		
		{\centering \setlength{\tabcolsep}{0pt} 
			\figurefont
			\renewcommand{\arraystretch}{0.1}
			\newcommand\figpath{figures/PartGrid/stylegan2_FFHQ_5858/}
			\newcommand\partone{eyes}
			\newcommand\parttwo{nose}
			\newcommand\partthree{mouth}
			\begin{tabular}{M{0.2\textwidth}M{0.18\textwidth}M{0.18\textwidth}M{0.18\textwidth}M{0.18\textwidth}}
				\textbf{Target} & \textbf{Reference 1} & \textbf{Reference 2} & \textbf{Reference 3}  & \textbf{Reference 4} \\
				\includegraphics[width=0.18\textwidth]{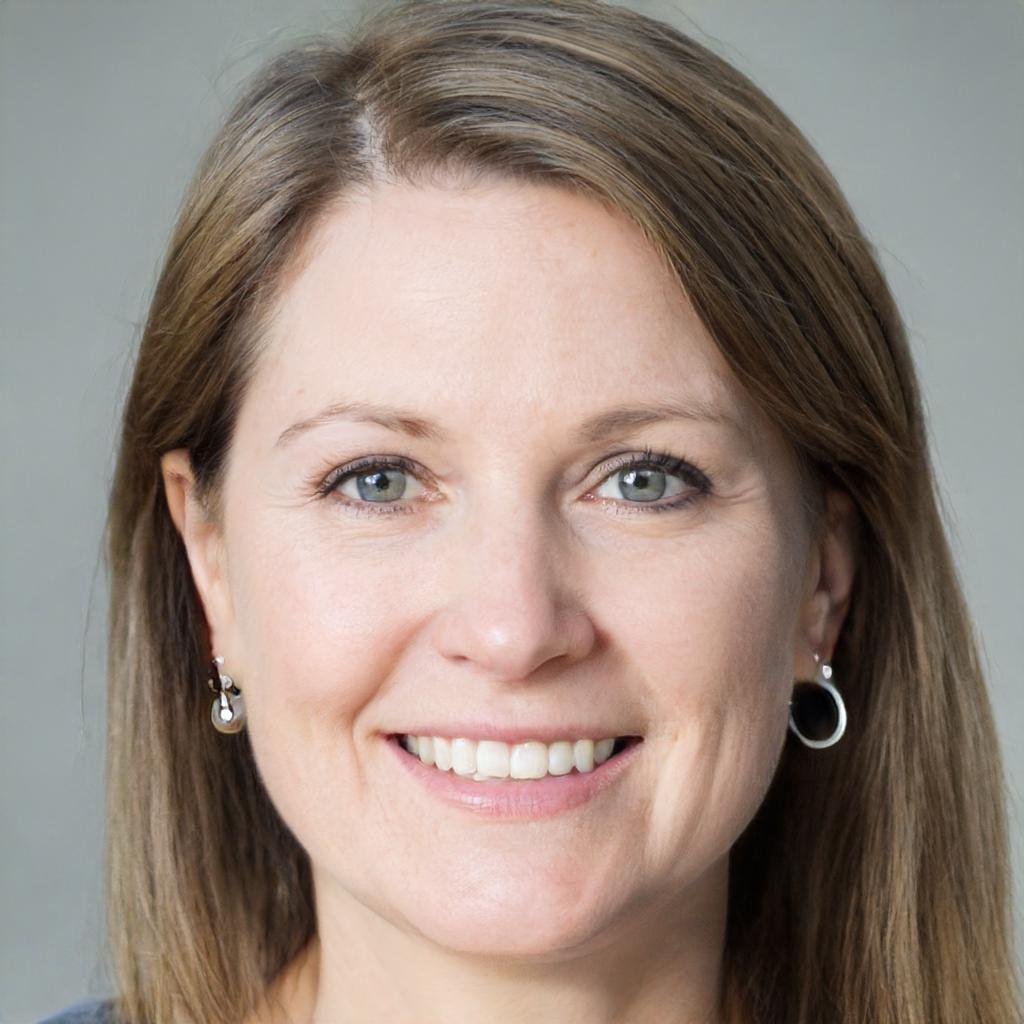}  &
				\includegraphics[width=0.18\textwidth]{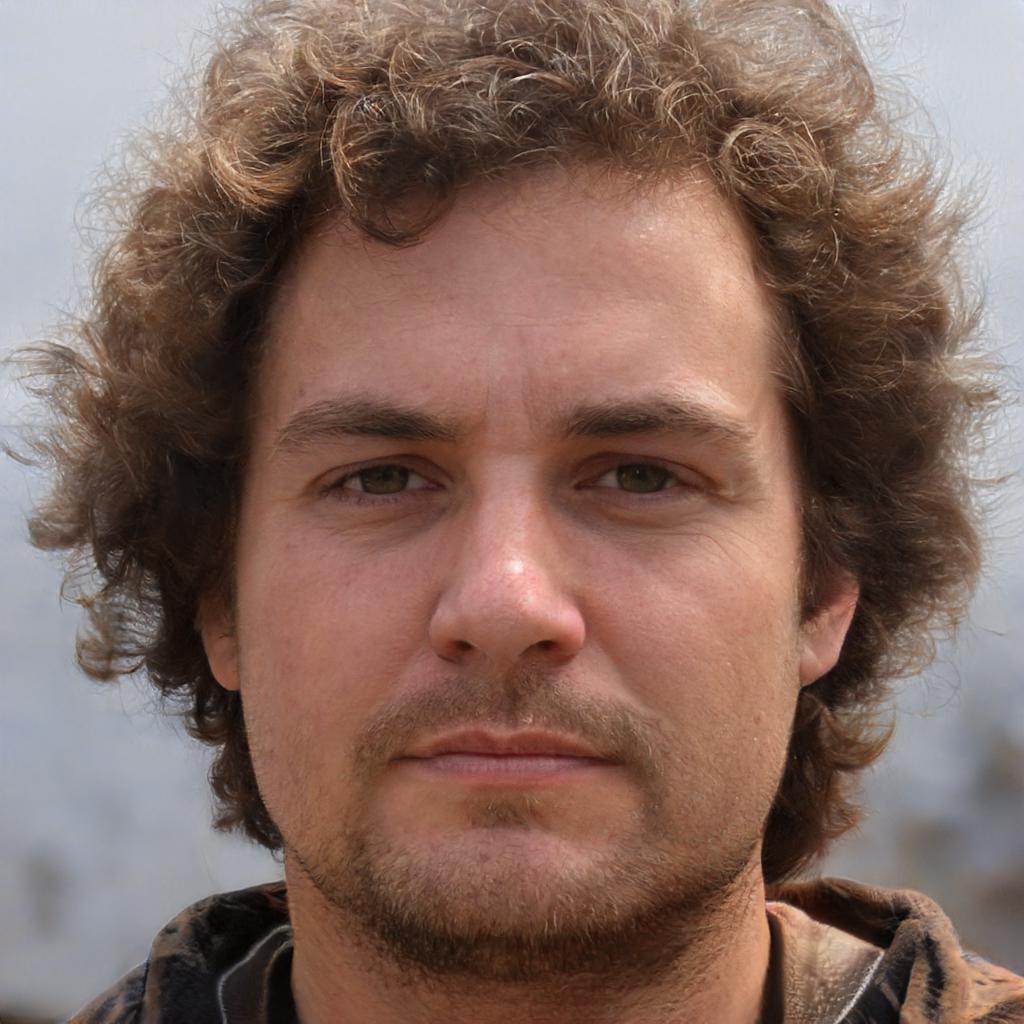} &
				\includegraphics[width=0.18\textwidth]{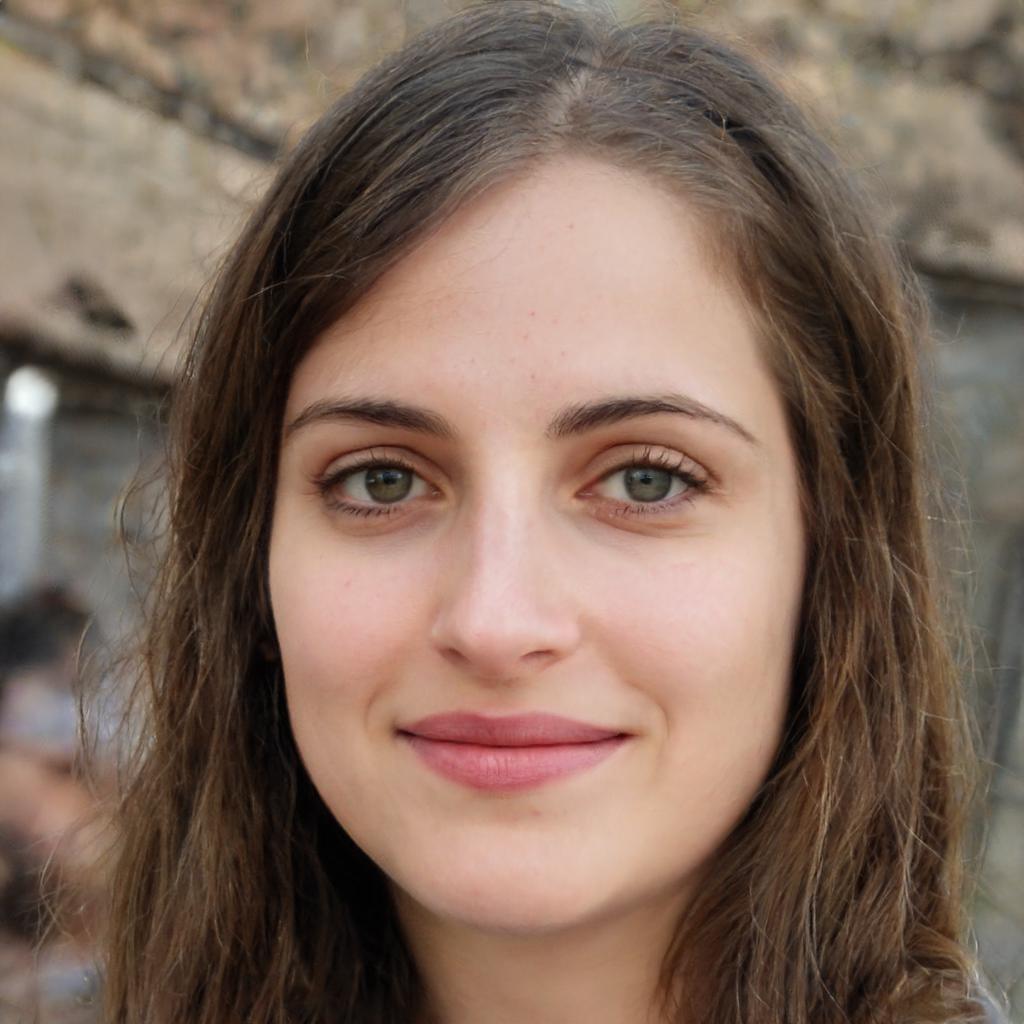} &
				\includegraphics[width=0.18\textwidth]{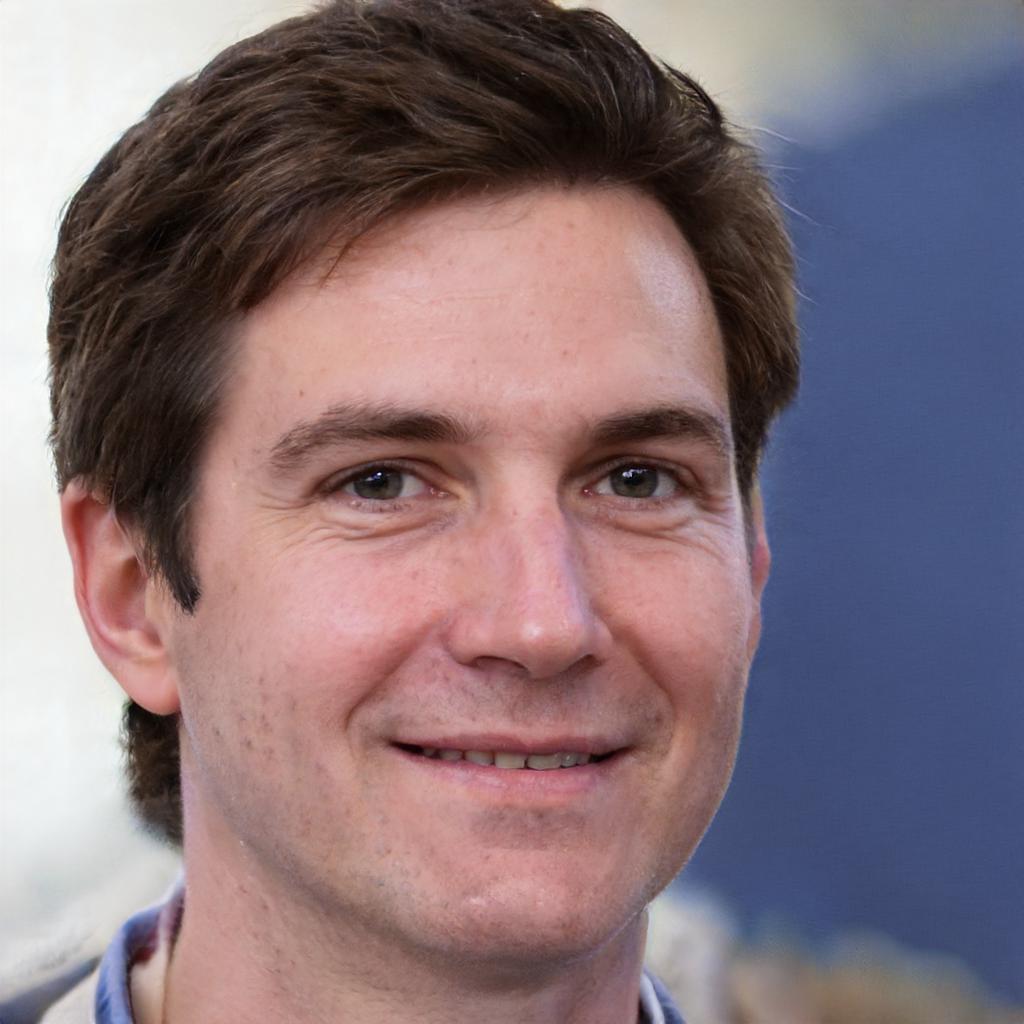} &
				\includegraphics[width=0.18\textwidth]{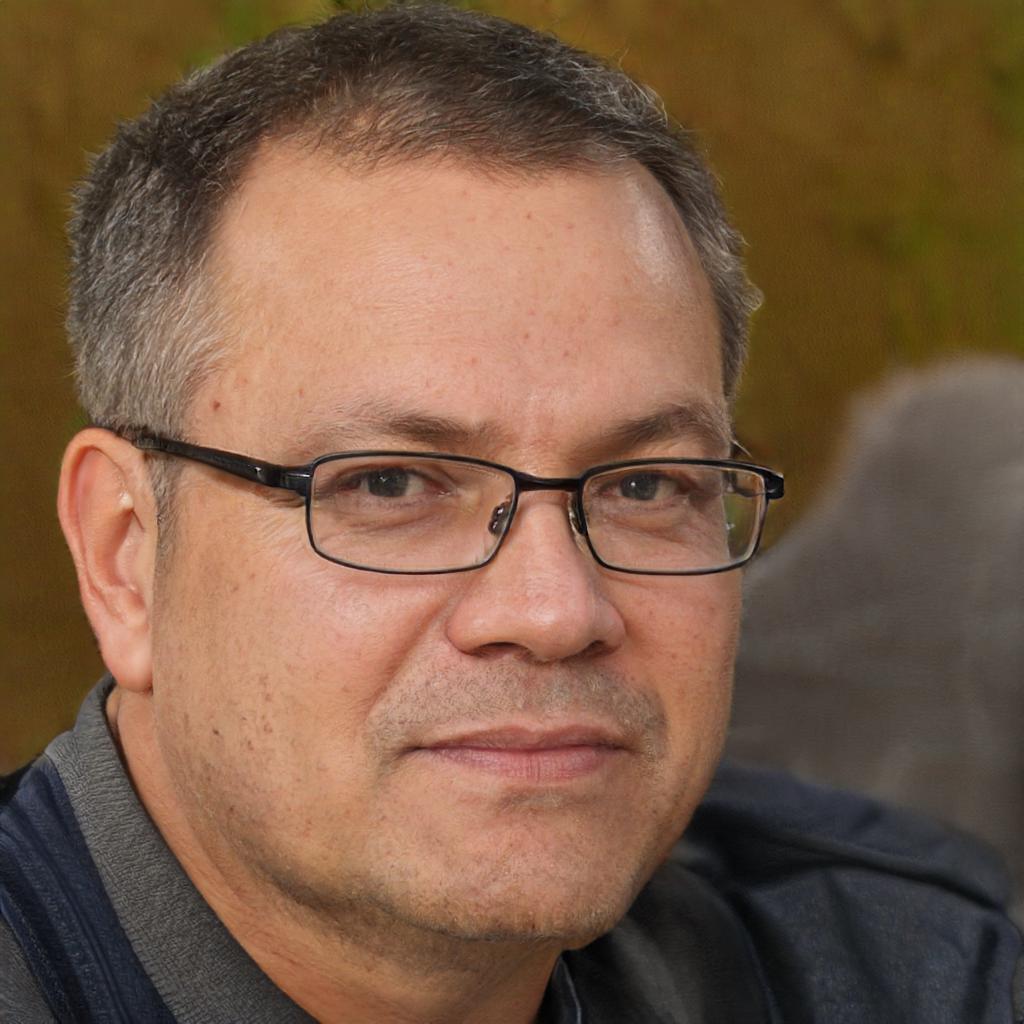}\\
				&&&& \\
				\textbf{\expandafter\MakeUppercase \partone:} &\includegraphics[width=0.18\textwidth]{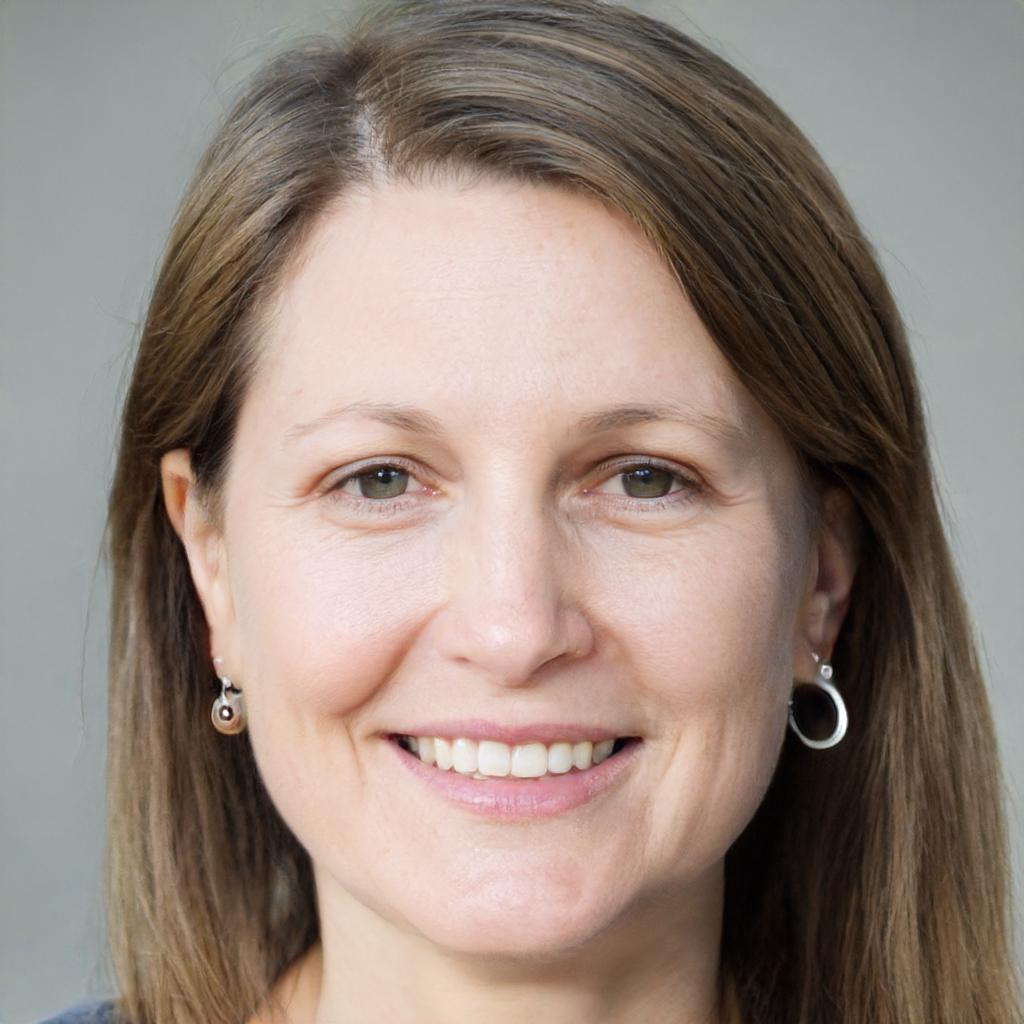} &
				\includegraphics[width=0.18\textwidth]{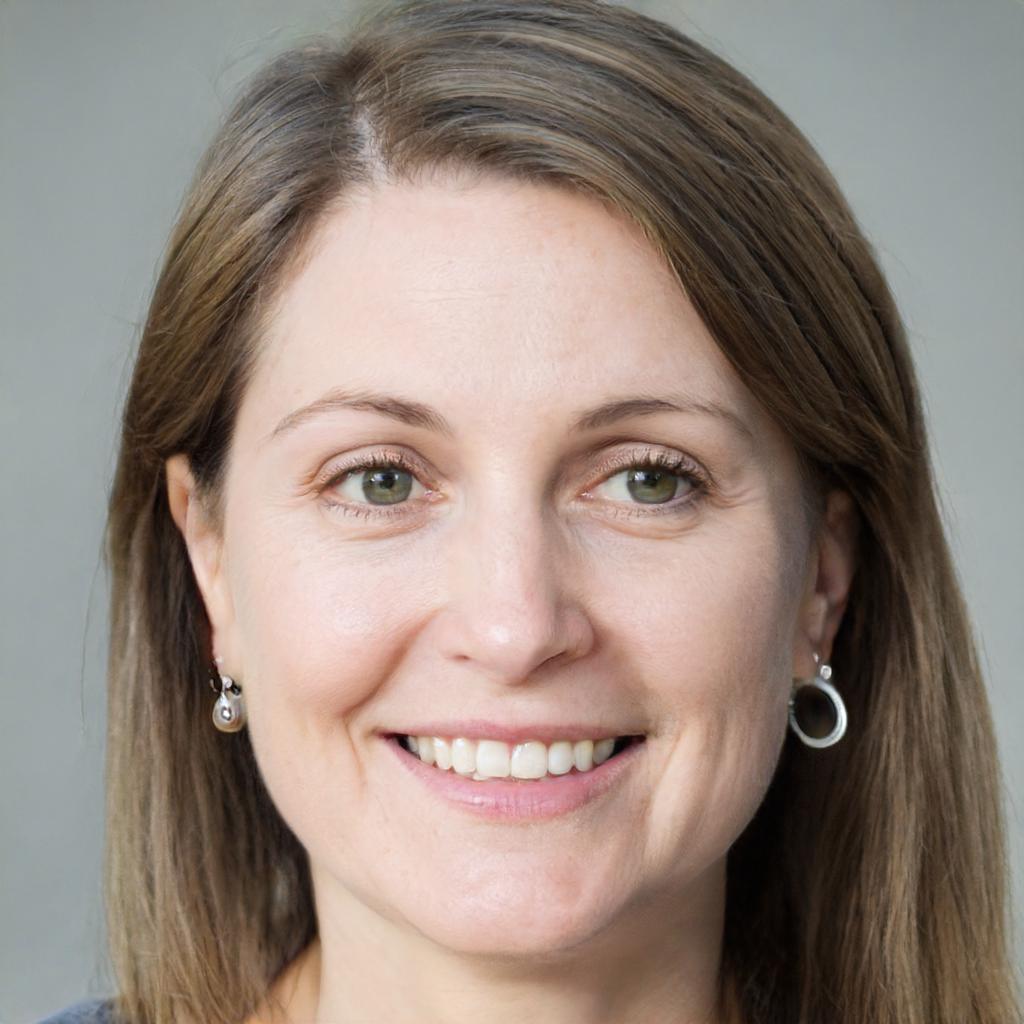} &
				\includegraphics[width=0.18\textwidth]{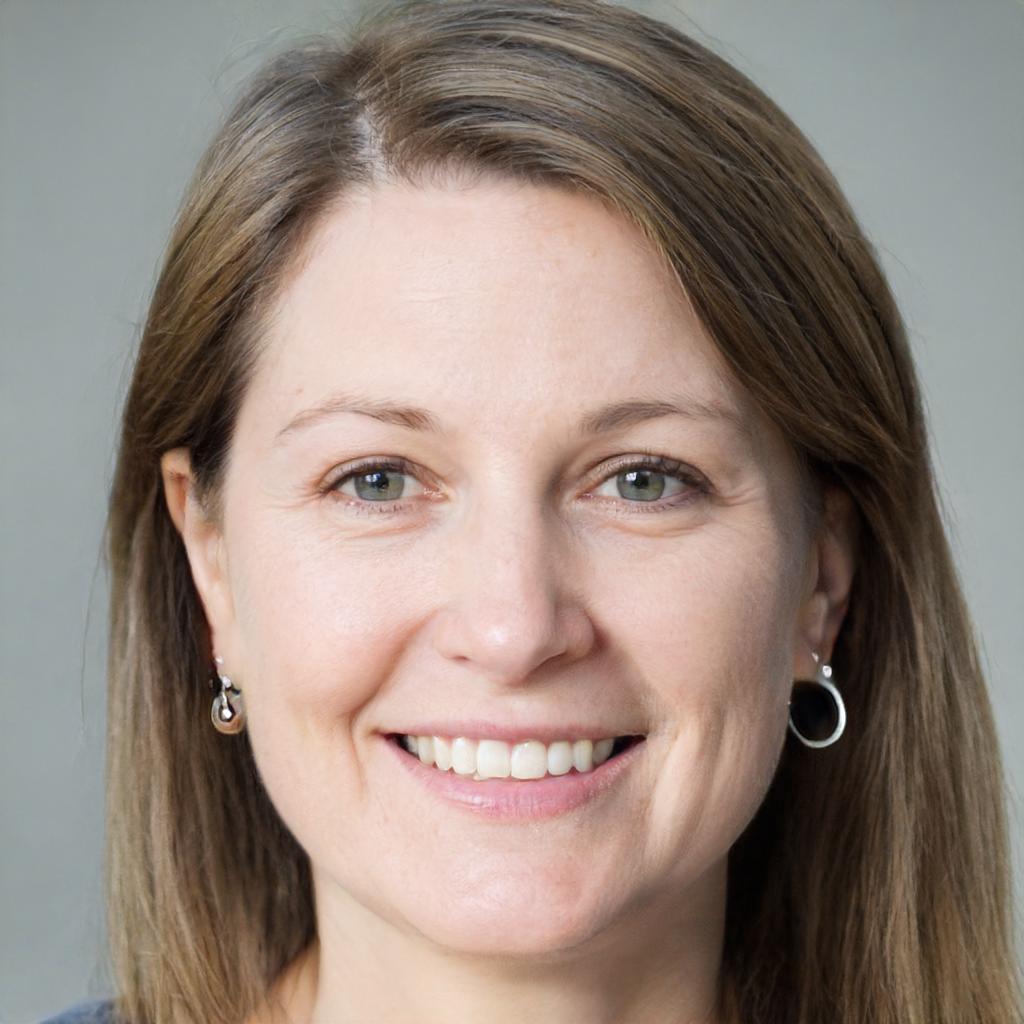} &
				\includegraphics[width=0.18\textwidth]{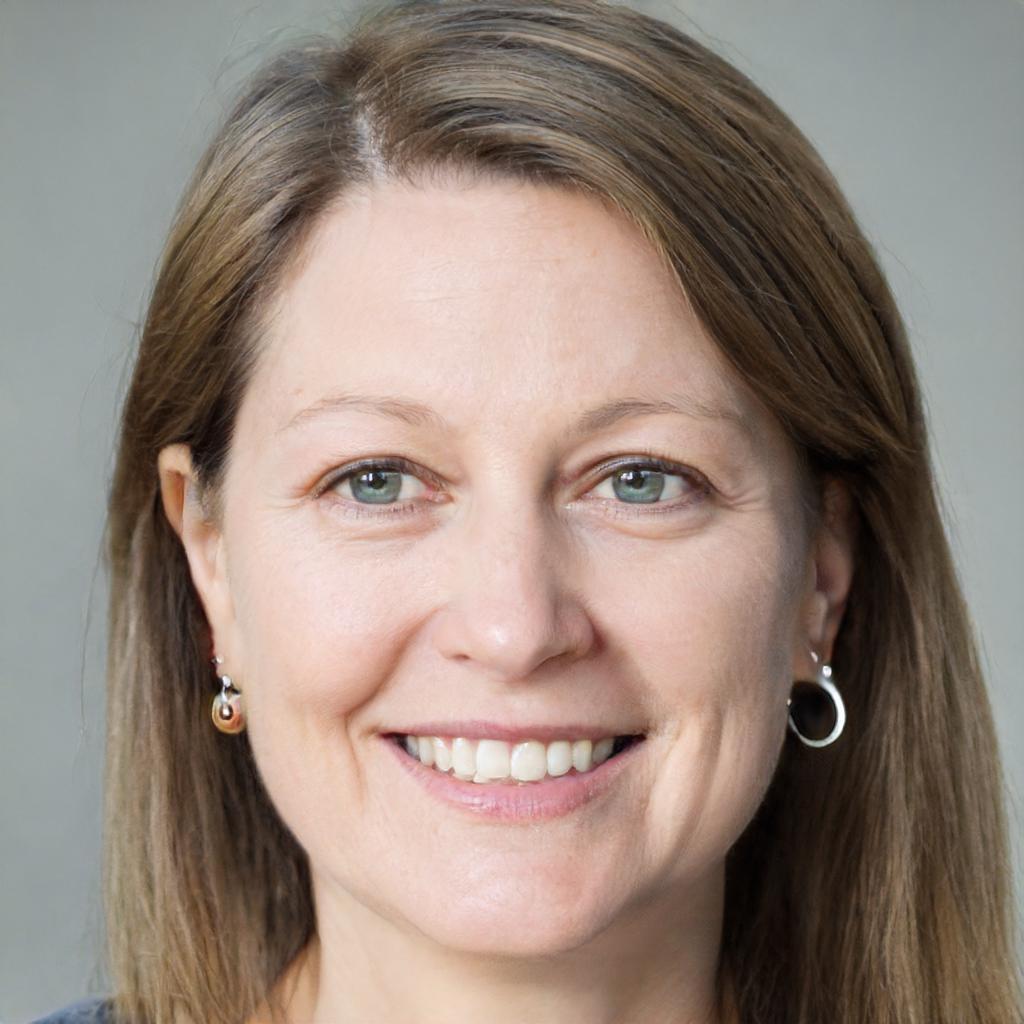}\\
				
				\textbf{\expandafter\MakeUppercase \parttwo:} &\includegraphics[width=0.18\textwidth]{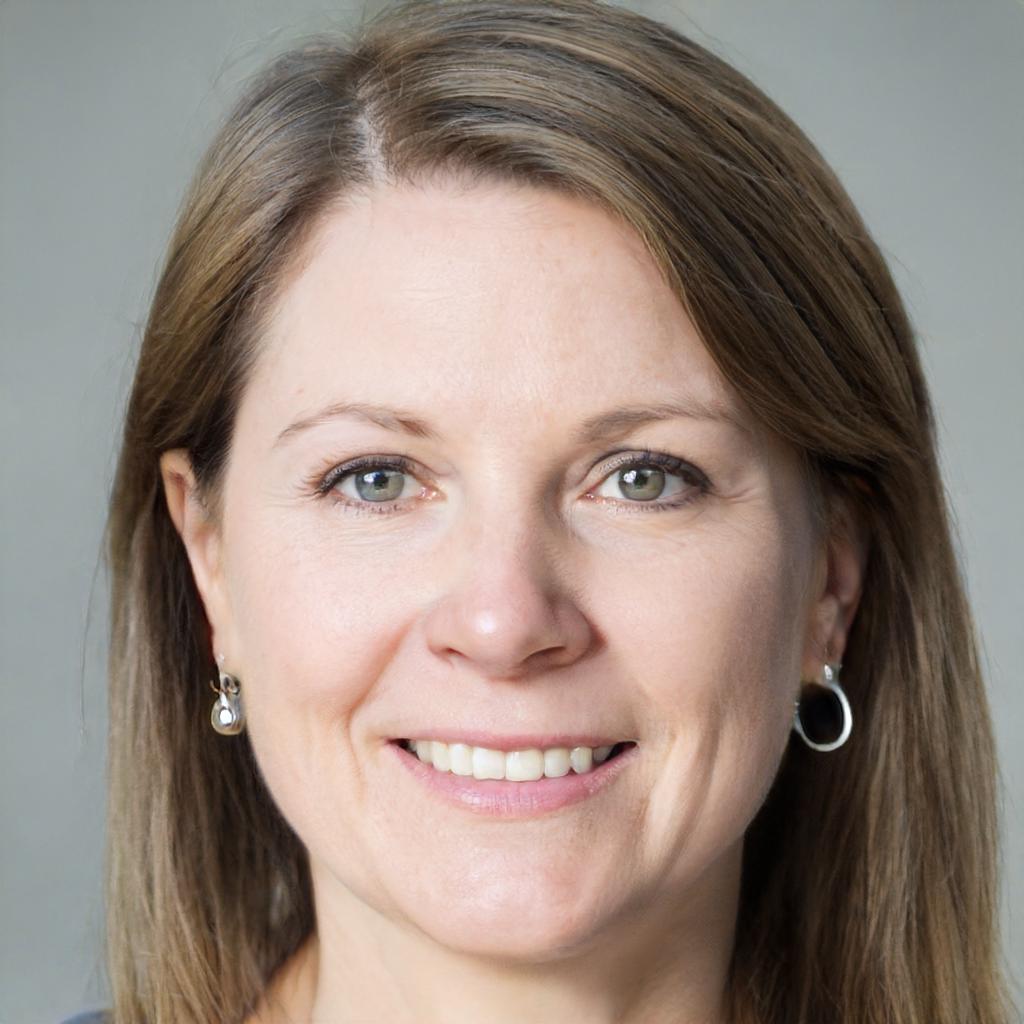} &
				\includegraphics[width=0.18\textwidth]{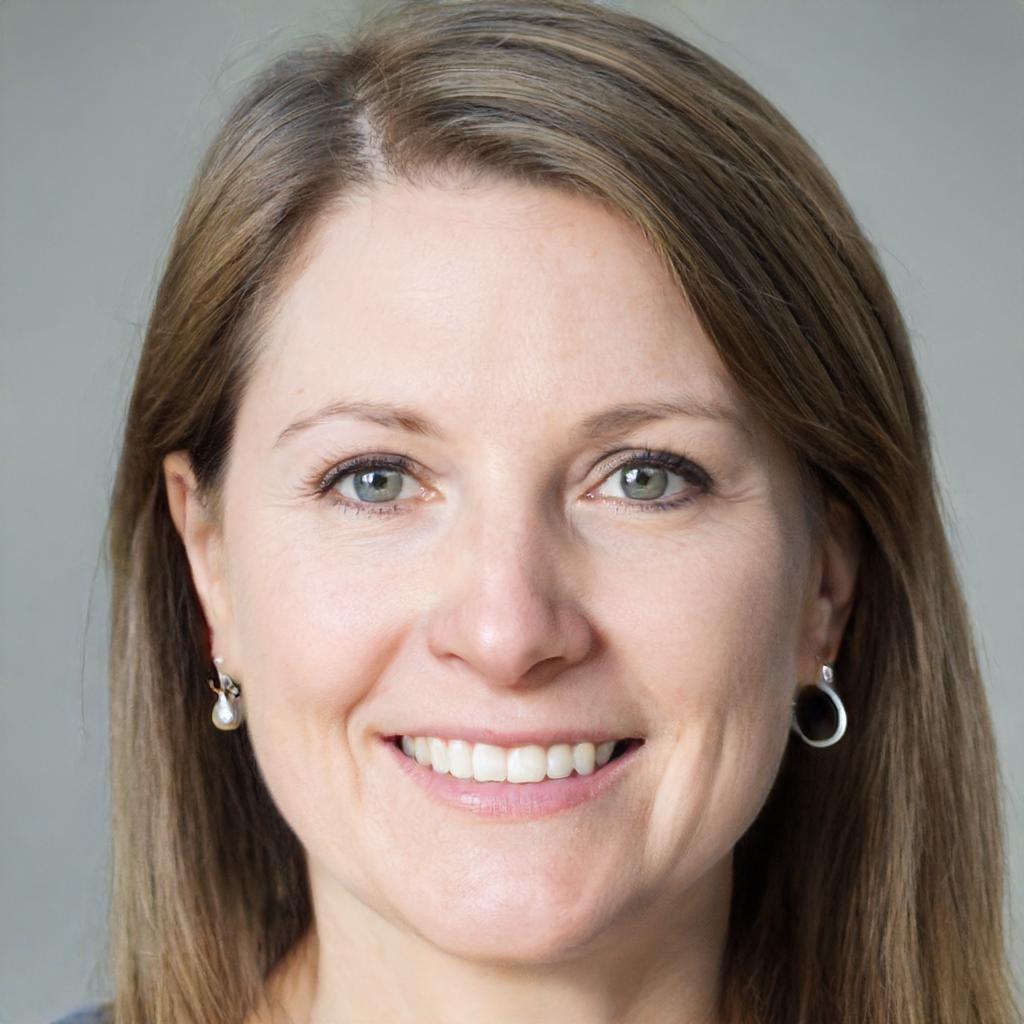} &
				\includegraphics[width=0.18\textwidth]{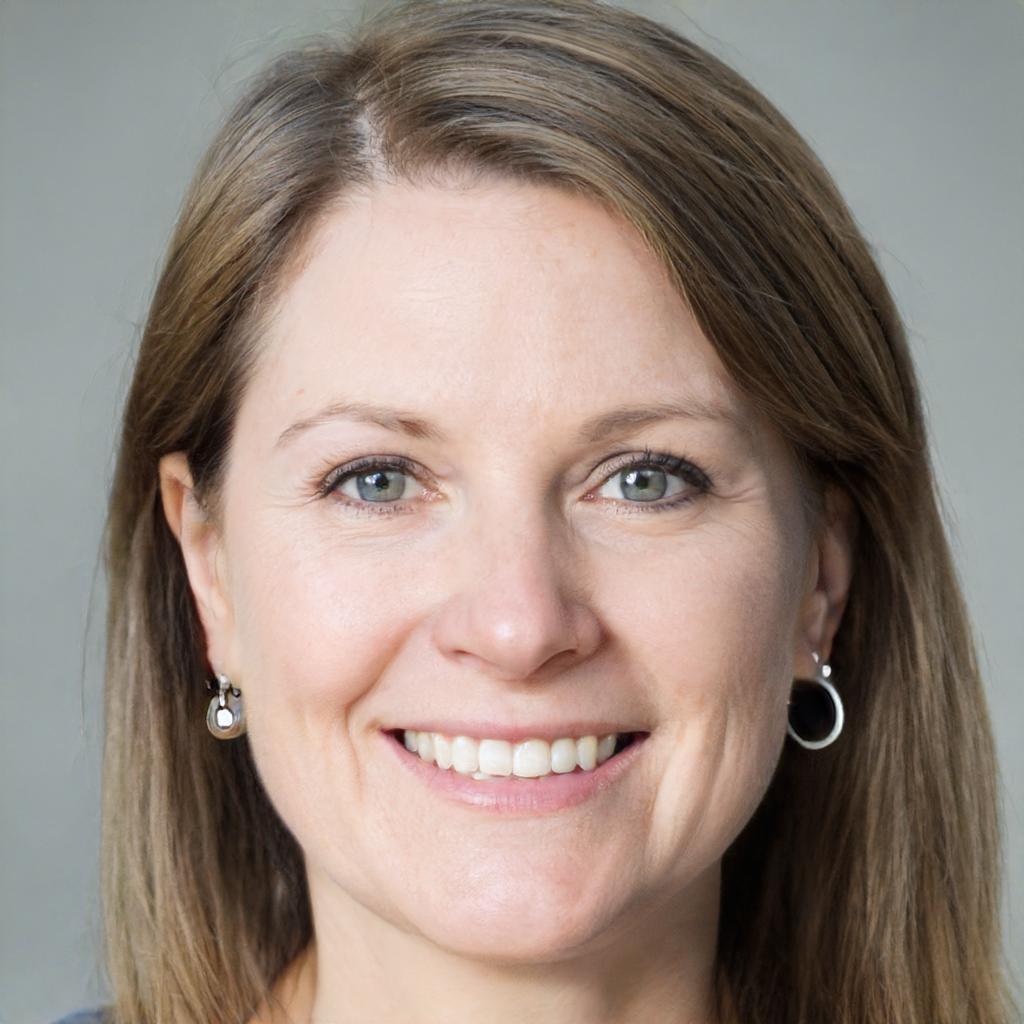} &
				\includegraphics[width=0.18\textwidth]{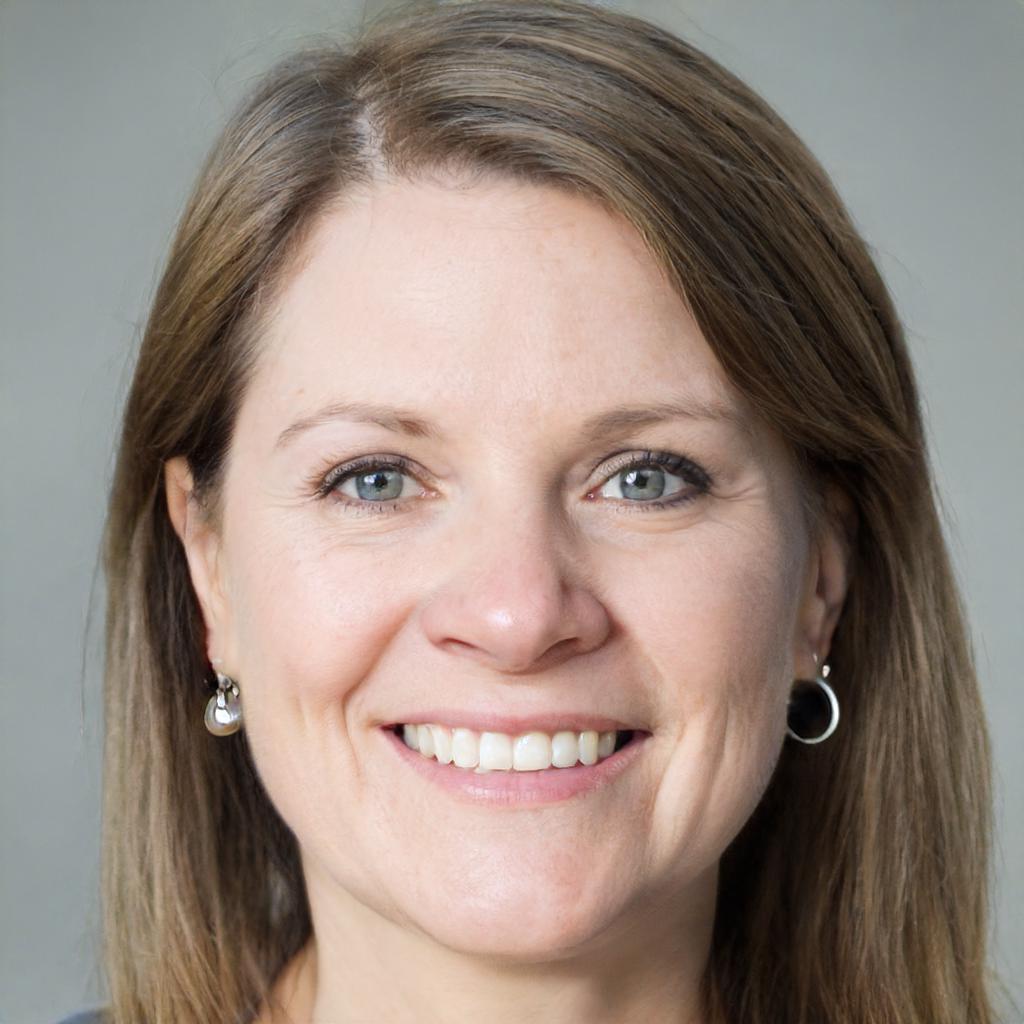}\\
				
				\textbf{\expandafter\MakeUppercase \partthree:} &\includegraphics[width=0.18\textwidth]{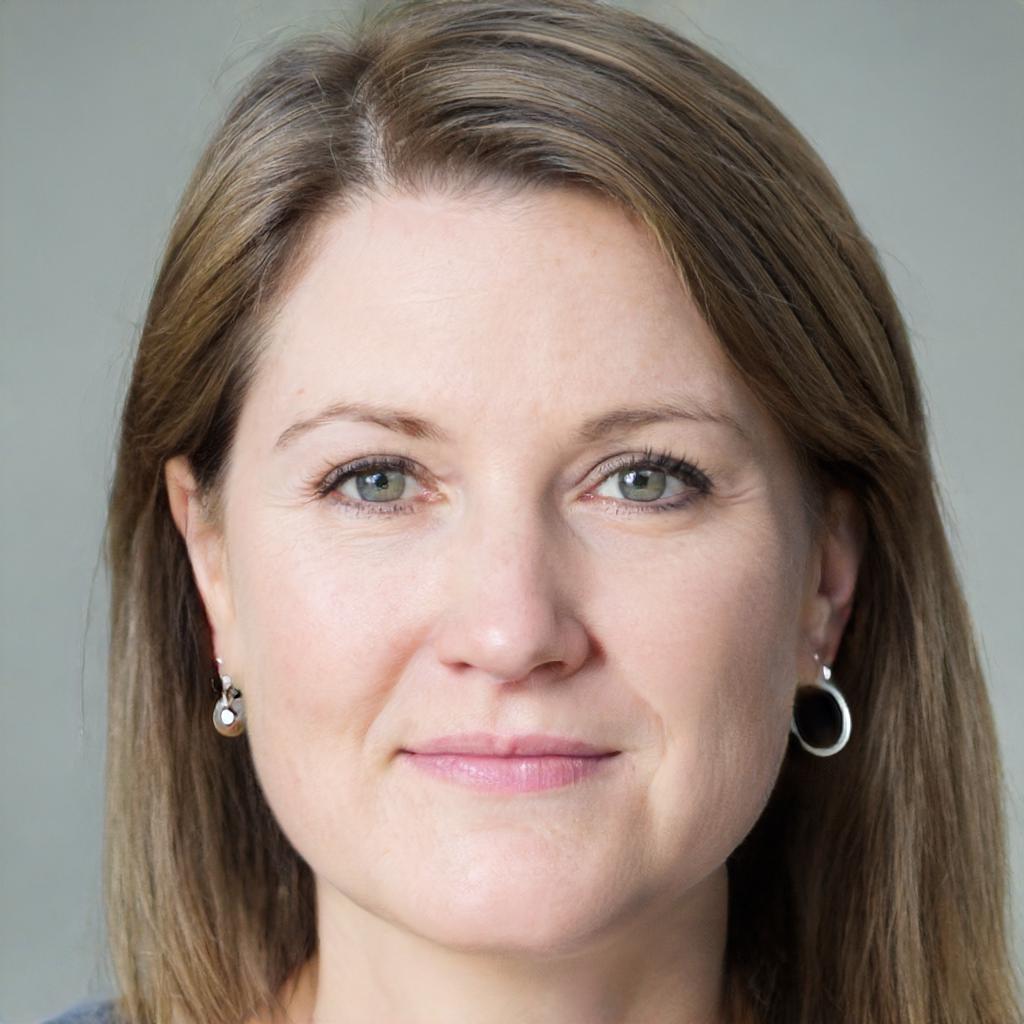} &
				\includegraphics[width=0.18\textwidth]{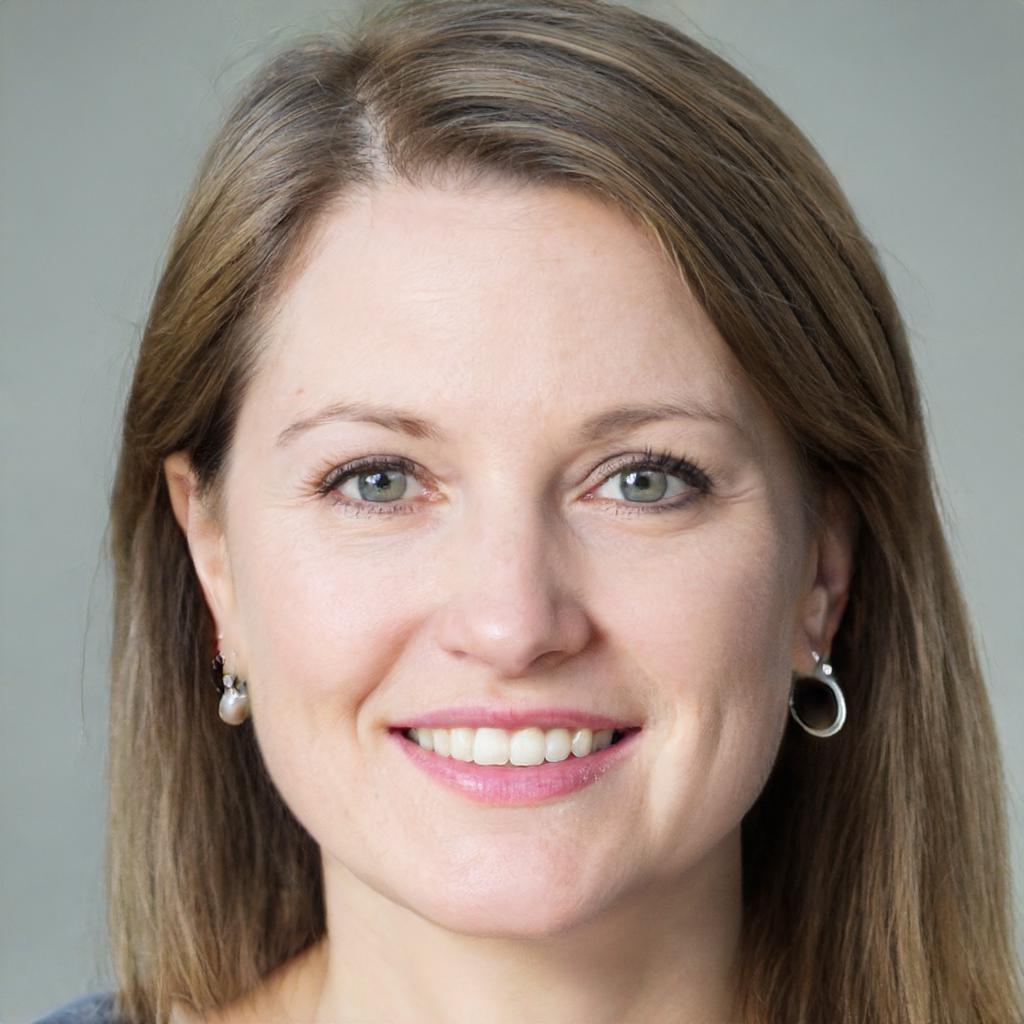} &
				\includegraphics[width=0.18\textwidth]{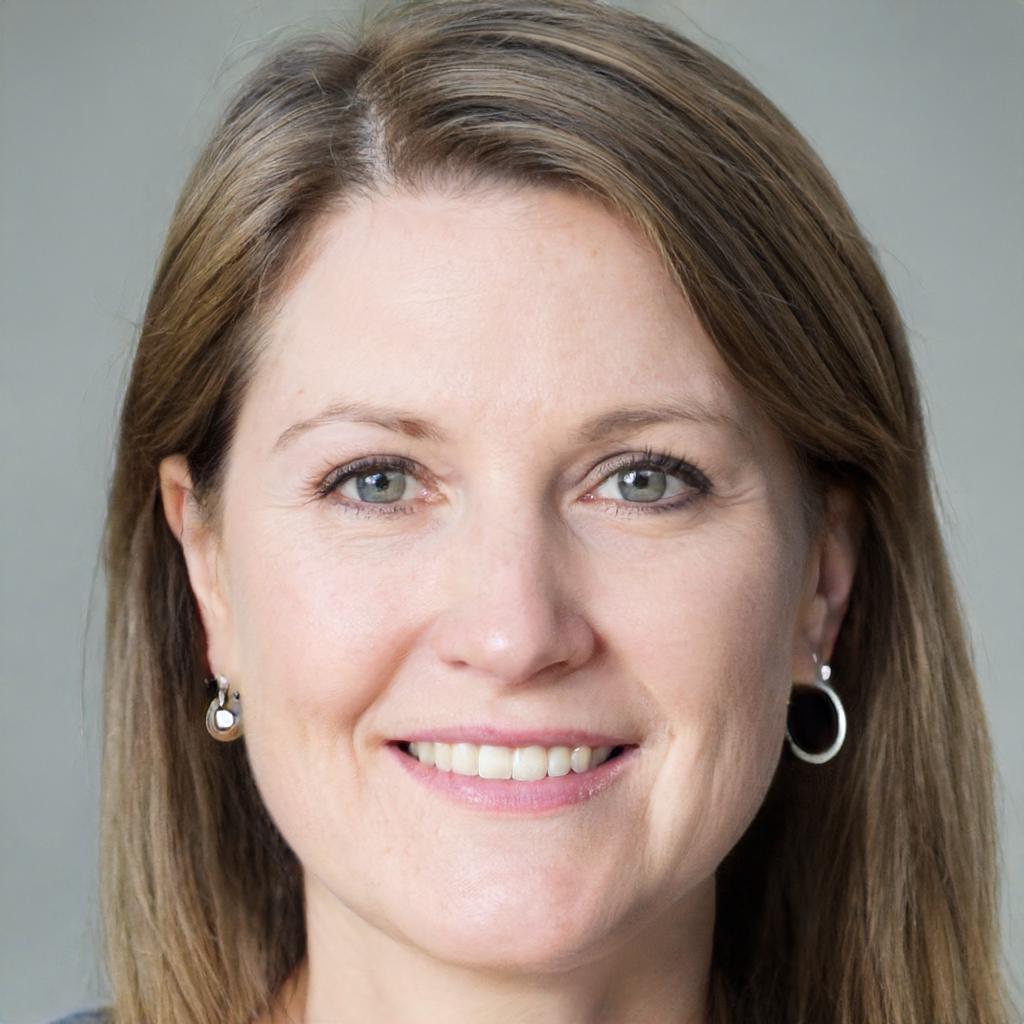} &
				\includegraphics[width=0.18\textwidth]{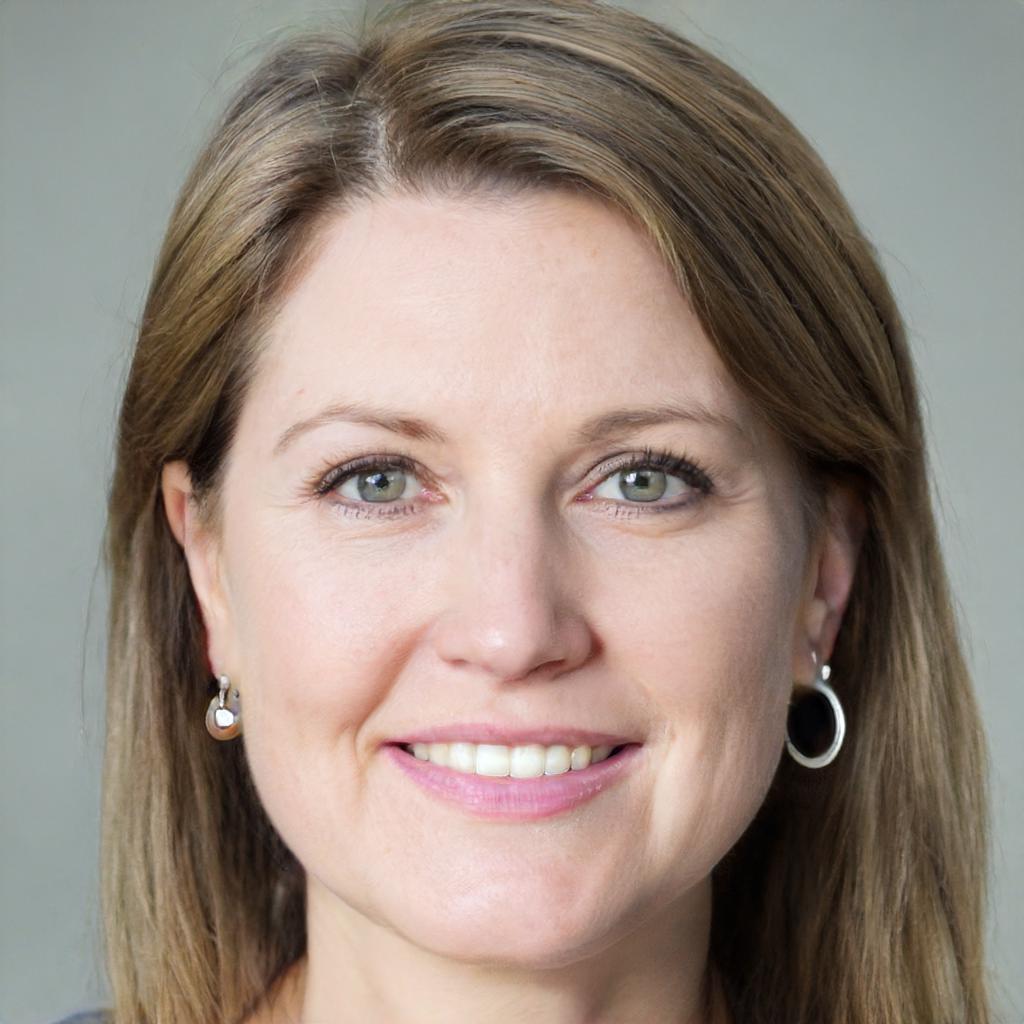}
		\end{tabular}}
		
		\caption{Our local editing method applied to StyleGAN2 trained on FFHQ.}
		\label{fig:main-stylegan2-ffhq}
	\end{figure*}

\begin{figure*}
		{\centering \setlength{\tabcolsep}{0pt} 
			\figurefont
			\renewcommand{\arraystretch}{0.1}
			\newcommand\figpathorig{figures/PartGrid/stylegan2_FFHQ_5858/}
			\newcommand\figpath{figures/PartGrid/stylegan2_FFHQ_5858_diff/}
			\newcommand\partone{eyes}
			\newcommand\parttwo{nose}
			\newcommand\partthree{mouth}
			\begin{tabular}{M{0.2\textwidth}M{0.18\textwidth}M{0.18\textwidth}M{0.18\textwidth}M{0.18\textwidth}}
				\textbf{Target} & \textbf{Reference 1} & \textbf{Reference 2} & \textbf{Reference 3}  & \textbf{Reference 4} \\
				\includegraphics[width=0.18\textwidth]{\figpathorig original.jpg}  &
				\includegraphics[width=0.18\textwidth]{\figpathorig style1.jpg} &
				\includegraphics[width=0.18\textwidth]{\figpathorig style2.jpg} &
				\includegraphics[width=0.18\textwidth]{\figpathorig style3.jpg} &
				\includegraphics[width=0.18\textwidth]{\figpathorig style4.jpg}\\
				&&&& \\
				\textbf{\expandafter\MakeUppercase \partone:} &\includegraphics[width=0.18\textwidth]{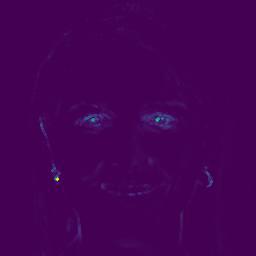} &
				\includegraphics[width=0.18\textwidth]{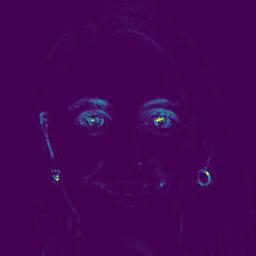} &
				\includegraphics[width=0.18\textwidth]{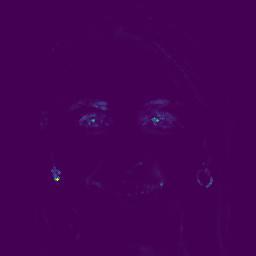} &
				\includegraphics[width=0.18\textwidth]{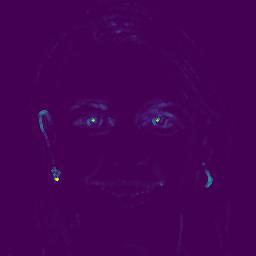}\\
				
				\textbf{\expandafter\MakeUppercase \parttwo:} &\includegraphics[width=0.18\textwidth]{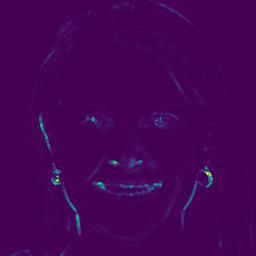} &
				\includegraphics[width=0.18\textwidth]{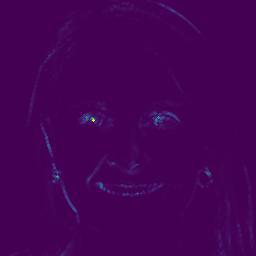} &
				\includegraphics[width=0.18\textwidth]{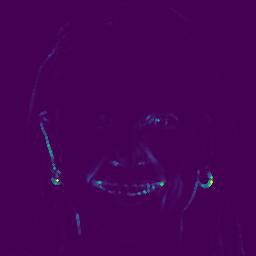} &
				\includegraphics[width=0.18\textwidth]{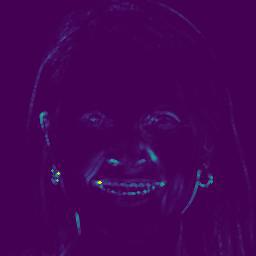}\\
				
				\textbf{\expandafter\MakeUppercase \partthree:} &\includegraphics[width=0.18\textwidth]{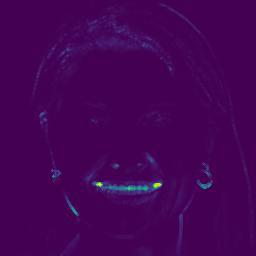} &
				\includegraphics[width=0.18\textwidth]{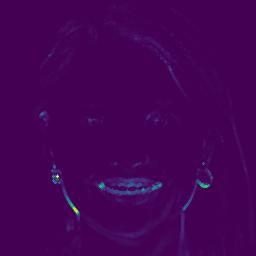} &
				\includegraphics[width=0.18\textwidth]{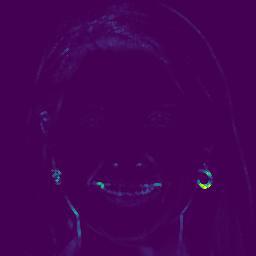} &
				\includegraphics[width=0.18\textwidth]{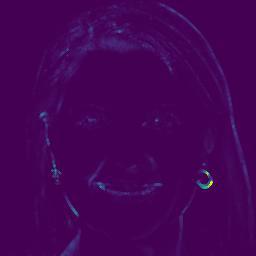}
		\end{tabular}}
		
		\caption{Diff maps corresponding to the results in Fig. \ref{fig:main-stylegan2-ffhq}, for StyleGAN2 trained on FFHQ.}
		\label{fig:main-stylegan2-ffhq-diff}
	\end{figure*}

	In this section we show additional results with StyleGAN2. Figs. \ref{fig:main-cats} and \ref{fig:main-cars} are extended versions of Fig. \ref{fig:main-stylegan2}.
	Figs. \ref{fig:main-cats-diff} and \ref{fig:main-cars-diff} show their diff maps.
	Figs. \ref{fig:main-stylegan2-ffhq} and \ref{fig:main-stylegan2-ffhq-diff} show results for StyleGAN2 trained of FFHQ.
	Additional examples can be found on the paper's GitHub page, linked above.

\end{document}

%% file: math_commands.tex
\usepackage{amsmath,amsfonts,bm}
\usepackage{amsthm}

\DeclareMathAlphabet{\mathcal}{OMS}{cmsy}{m}{n}

\def\eqref#1{equation~\ref{#1}}

\def\1{\bm{1}}

\def\va{{\bm{a}}}

\def\vq{{\bm{q}}}

\def\vw{{\bm{w}}}

\def\vz{{\bm{z}}}
\def\vsigma{{\bm{\sigma}}}

\def\mA{{\bm{A}}}

\def\mG{{\bm{G}}}

\def\mM{{\bm{M}}}

\def\mQ{{\bm{Q}}}
\def\mR{{\bm{R}}}
\def\mS{{\bm{S}}}

\def\mU{{\bm{U}}}
\def\mV{{\bm{V}}}

\DeclareMathAlphabet{\mathsfit}{\encodingdefault}{\sfdefault}{m}{sl}
\SetMathAlphabet{\mathsfit}{bold}{\encodingdefault}{\sfdefault}{bx}{n}
\newcommand{\tens}[1]{\bm{\mathsfit{#1}}}
\def\tA{{\tens{A}}}

\def\tU{{\tens{U}}}

\def\sR{{\mathbb{R}}}

\def\sW{{\mathbb{W}}}

\def\sZ{{\mathbb{Z}}}

\DeclareMathOperator*{\argmin}{arg\,min}